\documentclass{article}

\PassOptionsToPackage{numbers, sort, compress}{natbib}

\usepackage[main, final]{neurips_2026}

\usepackage[utf8]{inputenc} 
\usepackage[T1]{fontenc}    
\usepackage{hyperref}       
\usepackage{url}            
\usepackage{booktabs}       
\usepackage{amsfonts}       
\usepackage{nicefrac}       
\usepackage{microtype}      
\usepackage{xcolor}         
\usepackage{tabularx}
\usepackage{multirow} 
\usepackage{pifont}
\usepackage{makecell}
\usepackage{amsmath,amssymb} 
\usepackage{url}
\usepackage{hyperref}
\usepackage{svg}

\hypersetup{
	colorlinks=true,
	linkcolor=red,
	urlcolor=magenta,
	citecolor=blue,
}
\usepackage{colortbl}
\usepackage{graphicx}
\usepackage{listings}
\usepackage{setspace}
\usepackage{tablefootnote}

\usepackage{array}
\usepackage{booktabs} 
\usepackage{multirow} 
\usepackage{siunitx}  
\usepackage{wrapfig}    
\usepackage{threeparttable}
\usepackage{lipsum}     
\usepackage{minitoc}
\usepackage[toc,page,header]{appendix}
\usepackage{minitoc}
\usepackage{makecell}

\usepackage[most]{tcolorbox}
\usepackage{tikz}
\usetikzlibrary{arrows.meta}
\usepackage{xcolor}
\definecolor{navyblue}{RGB}{0, 0, 100} 

\definecolor{navyblue}{RGB}{20, 20, 132}

\usepackage[dvipsnames]{xcolor} 
\colorlet{jsonteal}{black}
\colorlet{reportblue}{black}
\definecolor{qwenpurple}{RGB}{112, 48, 160}

\newcommand{\cmark}{\textcolor{ForestGreen}{\ding{51}}} 
\newcommand{\xmark}{\textcolor{red}{\ding{55}}}       

\title{Anatomy-Activated Mixture-of-Experts\\ for 3D Medical Vision-Language Pre-training}

\author{
\bf Szymon Płotka\textsuperscript{1}\thanks{Correspondence to: Szymon Płotka (\href{mailto:szymon.plotka@uj.edu.pl}{\textsc{szymon.plotka@uj.edu.pl}})}\quad
\bf Gizem Mert\textsuperscript{2,3,4} \quad
\bf Pedro R. A. S. Bassi\textsuperscript{5,6,7} \quad 
\bf Wenxuan Li\textsuperscript{5,6,7} \quad
\\
\bf Zongwei Zhou\textsuperscript{5} \quad
\bf Anna Marcinkiewicz\textsuperscript{8} \quad
\bf Wiktoria Romańczyk\textsuperscript{9} \quad
\bf Jarosław B. Ćwikła\textsuperscript{10} \quad
\\
\bf Łukasz Struski\textsuperscript{1} \quad 
\bf Ewa Szczurek\textsuperscript{2,4,11,12} \quad
\bf Jacek Tabor\textsuperscript{1} \quad
\bf Arkadiusz Sitek\textsuperscript{6,7} \\
\textsuperscript{1} Jagiellonian University \quad
\textsuperscript{2} Helmholtz Munich \quad
\textsuperscript{3} Technical University of Munich \quad \\
\textsuperscript{4} Munich Center for Machine Learning \quad
\textsuperscript{5} Johns Hopkins University\\
\textsuperscript{6} Massachusetts General Hospital \quad
\textsuperscript{7} Harvard Medical School \quad \\
\textsuperscript{8} National Medical Institute of the Ministry of the Interior and Administration \\
\textsuperscript{9} Medical University of Białystok\quad
\textsuperscript{10} GAMMED\quad
\textsuperscript{11} University of Warsaw \quad
\textsuperscript{12} LMU Munich\\
}

\begin{document}

\maketitle
\doparttoc 
\faketableofcontents

\begin{abstract}
Medical Vision-Language Pre-training (VLP) leverages the semantic richness of radiology reports to learn 3D representations, yet current architectures remain fundamentally anatomy-agnostic. While clinical diagnostics rely on organ-specific context, standard encoders apply a uniform, dense parameterization to all volumetric patches, and existing Mixture-of-Experts (MoE) routers gate on isolated local features without explicit access to the organ composition of the scan or crop.
We propose Anatomy-Activated Mixture-of-Experts (A$^{2}$MoE), a framework that incorporates anatomical inductive biases within the model's computational path.  A$^{2}$MoE introduces two key components: (1) a Histogram Representation Router (HRR) that conditions expert selection on global visual context and a predicted anatomical
composition histogram, encouraging anatomy-dependent expert weighting; and (2) an Organ-Query Multi-Scale Projector (OQMSP) that utilizes learnable queries to extract organ-level embeddings for fine-grained alignment without requiring manual annotations at inference. By routing at the patch level rather than the token level, A$^{2}$MoE reduces routing complexity by orders of magnitude while satisfying load balancing by construction. Pre-trained on over 60,000 multi-phase CT scans, A$^{2}$MoE achieves state-of-the-art performance across segmentation, classification, and detection benchmarks. Furthermore, we demonstrate that our anatomically grounded inductive biases support robust generalization to MRI and ultrasound, establishing a scalable foundation for medical representation learning. The code is available at \href{https://github.com/gmum/A2MoE}{github.com/gmum/A2MoE}.
\end{abstract}

\section{Introduction}
\label{sec:intro}

Vision-Language Pre-training (VLP) has redefined multimodal representation learning by aligning visual features with semantic language cues in a shared embedding space~\cite{bao2022vlmo,yang2022vision,chen2023vlp,dou2022coarse,li2022blip,alayrac2022flamingo,moor2023med,li2023blip,su2024boost,zhang2025medunifier,chu2026medtri,liang2025medfilip}. 
In the medical domain, Medical Vision-Language Models (MedVLMs)~\cite{shuilarge,pmlr-v306-shi26b,cao2025boosting,hamamci2025better,zhao2026towards,claessens2025scaling,radford2021learning,zhang2022contrastive,moon2022multi,yan2025local,lei2026versatile,li2023lvit} leverage radiology reports to supervise 3D visual encoders, reducing dependence on scarce voxel-level annotations. 
Moreover, 3D medical images have an anatomical hierarchy that current encoders rarely exploit: local image patterns arise within organs, organs appear within broader scan-level anatomical compositions, and reports describe findings at corresponding organ-specific and global levels. 
Current MedVLM backbones largely ignore this structure, applying the same dense parameters to all volumetric patches regardless of their anatomical context. 
This is suboptimal because visually similar patterns can carry different clinical meanings in different organs or anatomical neighborhoods. 
We argue that medical VLP should therefore move beyond post-hoc image--text alignment and make anatomical context part of the model's computation path during feature extraction.

Recent advances in MedVLMs have recognized the importance of anatomical grounding, but have primarily addressed it through refined \textit{alignment objectives}. For instance, fVLM~\cite{shuilarge} matches CT regions to specific report snippets, while ViSD-Boost~\cite{cao2025boosting} enhances semantic density via organ-aware contrastive loss. However, these methods focus primarily on \textit{what} is aligned to text, but they do not fundamentally change \textit{how} visual features are computed inside the encoder. Current 3D backbones~\cite{xin2025med3dvlm, blankemeier2026merlin, lee2026read} treat every voxel with the same dense parameterization, regardless of whether a feature arises in the liver, kidney, spleen, bowel, or background. Explicit anatomical conditioning offers a complementary direction; in clinical practice, identical visual patterns often hold divergent meanings depending on their anatomical context. A "dark spot" (hypodensity), for example, carries distinct diagnostic weight in the liver versus the spleen or kidney, motivating an explicit anatomical input alongside learned visual features.

\begin{figure}
    \centering
    \includegraphics[width=0.8\linewidth]{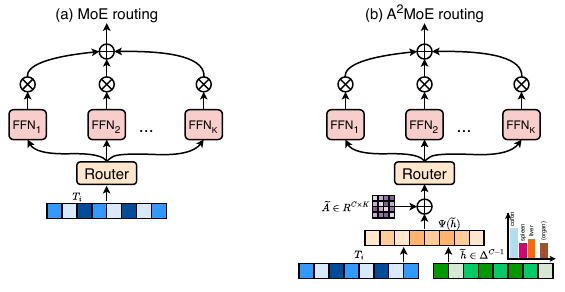}
    \caption{\textbf{Comparison of MoE routing strategies.} (a) MoE routing: The input feature map $x$ is pooled to form a global visual context, which is passed through an MLP router to produce logits over $K$ experts. Softmax-normalized weights are then used to compute a convex combination of expert outputs $E_k(x)$. Routing depends solely on visual features. (b) In addition to visual context, a foreground-normalized organ-composition histogram $\widetilde{\mathbf h}$ is embedded and fused with visual features to jointly condition the router. Each foreground entry represents the corresponding organ's share of the total foreground voxel mass in the input crop. A learnable class-to-expert affinity matrix $\mathbf A^{(s)}$ introduces a structural prior via a log-probability term $\log(\widetilde{\mathbf h}\widetilde{\mathbf A}^{(s)}+\epsilon)$, biasing expert selection toward anatomically consistent assignments. The resulting logits are softmax-normalized to obtain routing weights, and the final output is computed as a weighted sum of expert responses.}
    \label{fig:aamoe_routing}
\end{figure}

Mixture-of-Experts (MoE)~\cite{jacobs1991adaptive,shazeer2017,fedus2022switch} provides a principled mechanism for conditional computation, yet its standard formulation in computer vision remains "context-blind." Current MoE routers typically dispatch tokens based on isolated local patch embeddings, ignoring the global anatomical topology of the 3D volume. This leads to a suboptimal "semantic collision": patches with high textural similarity but distinct anatomical identities, such as a liver--kidney boundary versus a liver--stomach interface, are frequently mapped to the same expert parameters. Such interference dilutes the model’s capacity to learn specialized, organ-specific feature transforms. In addition to MoEs, existing cross-modal projection heads either rely on global pooling, which collapses local anatomical nuances, or depend on ground-truth masks to isolate organ-level evidence, a dependency that renders them impractical for real-world inference where such masks can be unavailable.

Motivated by these limitations in both expert routing and cross-modal projection, we propose \textbf{A}natomy-\textbf{A}ctivated \textbf{M}ixture-\textbf{o}f-\textbf{E}xperts (\textbf{A$^{2}$MoE}), a volumetric medical VLP framework that incorporates anatomical conditioning into the encoder and organ-indexed aggregation into the projector. Our central hypothesis is that \textit{explicitly conditioning model capacity on global anatomical composition allows the network to learn more discriminative, organ-specific representations than dense or context-blind backbones. By grounding the computation path in structural priors, the model can effectively decouple the complex, overlapping signals present in 3D medical scans}. 

To realize this, we introduce two core components. First, our Histogram Representation Router (HRR) jointly embeds global visual context with an anatomical composition histogram $\mathbf{h}$ and a learnable class-to-expert affinity prior (see Figure \ref{fig:aamoe_routing}). This conditions expert weighting on crop-level anatomical
context. Second, we address the challenge of mask-free inference with an Organ-Query Multi-Scale Projector (OQMSP). Instead of requiring explicit segmentation, OQMSP utilizes a set of learnable organ queries that cross-attend to MoE-refined features across multiple scales. This allows the model to "probe" for organ-specific evidence and extract per-organ embeddings based on learned attention weights, providing a scalable mechanism for fine-grained image–text alignment without masks at
inference. While pre-trained on CT, the structural inductive biases of A$^{2}$MoE are modality-agnostic, enabling seamless transfer to MRI and ultrasound where anatomical consistency remains a primary diagnostic pillar. Our main contributions are:

\begin{enumerate}
    \item \textbf{Anatomy-Activated MoE Routing.} We introduce the HRR, a novel routing mechanism that uses an anatomical composition histogram to guide conditional computation, enabling the model to allocate specialized parameters to specific anatomical structures.
    \item \textbf{Mask-Free Organ-Level Alignment.} We propose OQMSP, a projector that extracts per-organ visual embeddings through learned queries, allowing for fine-grained VLP alignment without the need for segmentation masks during inference.
    \item \textbf{Empirical Validation of Conditional VLP.} We show that anatomy-conditioned routing and organ-level alignment improve 3D medical representation learning and support robust transfer from CT pre-training to MRI and ultrasound.
\end{enumerate}

We demonstrate that A$^{2}$MoE serves as a robust foundation for multimodal representation learning. 
By pre-training on over 60,000 CT-report pairs, we achieve state-of-the-art performance across segmentation, classification, and detection tasks. Notably, A$^{2}$MoE exhibits scalability across diverse structural granularities, from fine-grained anomalies to large-scale anatomical systems, and maintains high generalization performance on out-of-distribution benchmarks. Ultimately, A$^{2}$MoE provides a scalable mechanism for aligning unstructured semantic priors with dense 3D representations, facilitating effective cross-modal knowledge transfer.

\section{Related Work}
\label{sec:related_work}

\noindent
\textbf{Mixture-of-Experts.} MoE enables conditional computation by routing inputs to specialized sub-networks, scaling model capacity without proportionally increasing computational cost~\cite{shazeer2017,fedus2022switch,riquelme2021scaling,Kada_2026_CVPR}. In vision, V-MoE~\cite{riquelme2021scaling} and Soft
MoE~\cite{puigcerversparse} route image tokens to experts based on
local patch embeddings, while LIMoE~\cite{mustafa2022multimodal}
extends sparse MoE to contrastive language-image pre-training.
In medical imaging, HoME~\cite{plotka2026mamba} combines Mamba
state-space blocks with soft expert routing for linear-time 3D
segmentation, and Swin~SMT~\cite{plotka2024swin} integrates Soft~MoE
into Swin~UNETR for whole-body CT segmentation.
PAMoE~\cite{wu2025learning} routes tissue patches to specialized experts for gigapixel whole-slide image analysis, adapting capacity to heterogeneous histological regions, while
ERMoE~\cite{cheng2026ermoe} reparameterizes experts in a learned
eigenbasis and shows that the resulting routing yields anatomically
interpretable specializations in 3D brain MRI. In the multimodal domain, MoHAVE~\cite{kim2025mohave} and
MoME~\cite{cappellazzo2025mome} apply hierarchical and Matryoshka-style MoE to audio-visual speech recognition, while
M3-JEPA~\cite{lei2025mjepa} uses a multi-gate MoE predictor for
general multimodal alignment. While current MoE models perform routing based on low-level feature heuristics, they remain agnostic to the global anatomical landscape. Our work addresses this fundamental gap: the Histogram Representation Router is the first to condition expert selection on a high-level, structured representation of anatomical distribution, ensuring that routing is biologically informed rather than purely data-driven.

\noindent
\textbf{Anatomical conditioning and organ-level alignment.} Anatomical priors have been incorporated into both expert routing
and medical vision--language representation learning.
OAMoE~\cite{mi2026organ} organizes segmentation experts into
anatomically guided pools with adaptive expert selection. HAR-MoE~\cite{mu2026har} combines
task-conditioned expert screening with organ-conditioned refinement
and learnable affinity priors for unified ultrasound segmentation
and classification, sharing image-level routing decisions across
spatial tokens. For cross-modal learning, ASG~\cite{li2024anatomical} aligns
anatomical regions with corresponding report sentences, while
Anatomy-VLM~\cite{gu2026anatomy} combines anatomical localization,
structured clinical knowledge, and multi-scale alignment for
chest radiograph interpretation.
In volumetric CT, fVLM~\cite{shuilarge} performs anatomy-level
contrastive pre-training with disease-aware pairing;
ViSD-Boost~\cite{cao2025boosting} strengthens anatomical
representations through normal--abnormal discrimination and
anatomy normality modeling; and RADAR~\cite{zhang2026expert}
scales anatomy-wise image--text learning for generalist abdominal
CT diagnosis.
Building on these directions, A$^{2}$MoE couples
crop-composition-conditioned encoder routing with multi-scale
organ-level alignment: an anatomy histogram and learned
class-to-expert affinity prior condition expert mixtures, while
shared organ queries aggregate features across encoder stages
without masks at inference.

\noindent
\textbf{Medical Vision--Language Pre-training.}
Early methods such as GLoRIA~\cite{huang2021gloria} and
LoVT~\cite{muller2022joint} combine global image--report alignment
with local cross-modal objectives to learn transferable
representations from chest radiographs.
Subsequent work extends language-supervised representation
learning to volumetric imaging through complementary strategies.
Merlin~\cite{blankemeier2026merlin} combines CT volumes,
radiology reports, and structured electronic health records
to support diverse downstream tasks.
HLIP~\cite{zhao2026towards} enables pre-training on uncurated
clinical studies through hierarchical attention across slices,
scans, and studies, while SPECTRE~\cite{claessens2025scaling}
combines self-supervised and cross-modal pre-training for
volumetric CT transformers.
MG-3D~\cite{ni2026mg} incorporates multi-grained report knowledge
through intra-patient cross-modal learning and inter-patient
semantic alignment, while ExGra-Med~\cite{mh2026exgra}
explores graph-based cross-modal alignment for semantic grounding.
Beyond representation alignment,
R-Super~\cite{bassi2025learning,bassi2025scaling} converts
radiology reports into voxel-wise supervision for tumor
segmentation, extending report-derived knowledge to dense prediction.
Addressing volumetric representation efficiency,
BTB3D~\cite{hamamci2025better} learns compact, frequency-aware
3D tokens for report generation and text-conditioned CT synthesis.
Complementing these methodological advances,
CARES~\cite{xia2024cares} benchmarks the trustworthiness of
medical vision--language models.
A$^{2}$MoE builds on this progress by coupling
anatomy-conditioned expert mixtures within the visual encoder
with shared organ queries for multi-scale image--text alignment.

\section{Preliminaries}

\paragraph{Selective State-Space Models (Mamba).} Our architecture builds on the Mamba layer~\cite{mamba}, a structured state-space model designed for long-sequence modeling with linear computational complexity. In contrast to Transformer architectures, whose attention mechanism incurs a quadratic cost in sequence length, Mamba achieves linear-time processing via a discrete-time recurrence derived from continuous-time state-space dynamics. Given a feature sequence \(z = \{z_t\}_{t=1}^N \in \mathbb{R}^{B \times N \times d}\), the hidden state \(h_t\) and output \(y_t\) evolve according to:
\begin{equation}
h_t = A(z_t)\, h_{t-1} + B(z_t)\, z_t, 
\qquad 
y_t = C(z_t)\, h_t,
\end{equation}
where \(A(\cdot), B(\cdot), C(\cdot)\) denote learnable, input-dependent linear maps. This parameterization enables the model to capture both short- and long-range dependencies with computational cost that scales
linearly with sequence length for fixed feature and state dimensions \(\mathcal{O}(Nd)\). Within our network, the Mamba layer acts as the backbone for volumetric sequence modeling, providing an efficient mechanism for capturing spatial dependencies in 3D imaging data.

\paragraph{Dense Mixtures-of-Experts with Soft Gating.} A Mixture-of-Experts (MoE)~\cite{jacobs1991adaptive} processes inputs
using a bank of $E$ independent experts. Unlike sparse MoE
\cite{riquelme2021scaling,mustafa2022multimodal}, which uses
top-$k$ gating~\cite{zhou2022mixture,lapsum} to activate a subset
of experts, dense soft gating computes a differentiable weighted
combination of all $E$ expert outputs:
\begin{equation}
y_{b,n} = \sum_{e=1}^{E} p_{b,n,e}\,
\mathrm{FFN}_e(x_{b,n}),
\end{equation}
where $\mathrm{FFN}_e$ denotes the $e$-th expert and
$p_{b,n,e}\geq 0$, with $\sum_e p_{b,n,e}=1$, denotes its
mixture weight. This dense output mixture differs from the
slot-based token dispatch used in Soft MoE~\cite{puigcerversparse}.

Token-level gating requires routing-logit evaluations proportional
to the number of tokens. Our anatomy-activated MoE instead predicts
crop-specific mixture weights that are shared across spatial
positions within each encoder stage and conditioned on anatomical
composition. All experts still execute; spatially shared gating
reduces routing-logit evaluations rather than sparsifying expert
computation. Here, anatomy-activated refers to anatomical
conditioning of the mixture weights, without requiring equal
expert utilization or a one-to-one correspondence between experts
and organs.

\section{Methodology}

\subsection{Anatomy-aware image--text pairs}
\label{sec:data}

We denote a multimodal dataset as $\{(\mathbf{X}_n^{I},\,\mathbf{X}_n^{T})\}_{n=1}^{N}$, where $\mathbf{X}_n^{I}$ is the $n$-th CT volume, $\mathbf{X}_n^{T}$ its radiology report, and $N$ the number of pairs. Following \cite{shuilarge}, we decompose each pair into anatomy-level components:
\begin{equation}
(\mathbf{X}_n^{I},\,\mathbf{X}_n^{T}) \;\longmapsto\;
\bigl\{(\mathbf{X}_{n,c}^{I},\,\mathbf{X}_{n,c}^{T})\bigr\}_{c\in\mathcal{C}_n},
\end{equation}
where $\mathbf{X}_{n,c}^{I}$ and $\mathbf{X}_{n,c}^{T}$ are the volumetric region and free-text findings of organ class $c$, and $\mathcal{C}_n\subseteq\{1,\dots,C-1\}$ is the set of anatomical classes present and described for the $n$-th study. Anatomical structures are extracted via TotalSegmentator~\cite{wasserthal2023totalsegmentator}. Diagnostic reports are decomposed into organ-level findings with Qwen3-30B-A3B-Instruct~\cite{yang2025qwen3}. Subsequently, the masks and text snippets are reviewed by experienced radiologists to improve anatomical precision (see Appendix~\ref{sec:appendixdatapreprocessing}). These masks and organ-level snippets are used to construct pre-training supervision only; at inference, A$^2$MoE requires only the raw CT volume as input.

\subsection{Anatomy-Activated Mixture-of-Experts (A$^2$MoE) layer}
\label{sec:a2moe}

Dense 3D encoders route every spatial position through identical feed-forward parameters, implicitly assuming homogeneous feature statistics across the volume \cite{lu2026segmote,plotka2026mamba}. This is suboptimal for volumetric medical images, where anatomical structures vary in scale, shape, and appearance within a single crop \cite{szczepanski2024let,zhang2025foundation}. Standard MoE layers activate different experts per input but condition gating solely on hidden visual features, which does not explicitly represent the anatomical composition of the crop. We propose \textbf{A$^2$MoE}, which conditions expert selection on a compact \emph{anatomy representation} of the crop together with global visual context. A$^2$MoE executes four steps at each encoder stage $s$: (i)~build a batch of routing vectors $\mathbf{v}_s$ from the feature map $\mathbf{X}_s$; (ii)~form a normalized histogram representation $\widetilde{\mathbf{h}}$; (iii)~map $(\mathbf{v}_s,\widetilde{\mathbf{h}})$ to expert logits and softmax weights $\mathbf{W}_s\in\mathbb{R}^{B\times K_s}$; (iv)~apply the same mixture $\mathbf{W}_s$ to every spatial location of $\mathbf{X}_s$ and add a residual. 

\noindent
\textbf{Routing context.} Let $\mathbf{X}_s\in\mathbb{R}^{B\times C_s\times D_s\times H_s\times W_s}$ denote the feature volume at hierarchical encoder stage $s\in\{1,2,3,4\}$. We aggregate complementary global spatial statistics using global average and max pooling (GAP, GMP) to capture both tissue composition and salient structural boundaries, then concatenate a learnable stage embedding $\mathbf{e}_s\in\mathbb{R}^{d_s}$:
\begin{equation}
\mathbf{v}_s \;=\;
\bigl[\operatorname{GAP}(\mathbf{X}_s)\,\|\,\operatorname{GMP}(\mathbf{X}_s)\,\|\,\mathbf{e}_s\bigr]
\;\in\;\mathbb{R}^{B\times (2C_s+d_s)}.
\end{equation}
The stage embedding $\mathbf e_s$ is broadcast across the $B$ samples. GAP captures overall tissue composition, while GMP preserves small or boundary structures that averaging would suppress.

\noindent
\textbf{Anatomical composition.} Let $\Omega=\{1,\ldots,DHW\}$ denote the
voxel domain of the input crop, and let class $0$ denote background. From the
crop's pseudo-label mask $L$, we compute:
\begin{equation}
 h_{b,c}=\frac{1}{|\Omega|}\sum_{j\in\Omega}\mathbf 1[L_{b,j}=c],
 \qquad c\in\{0,\ldots,C-1\}.
\end{equation}
Each stage predicts raw composition using a lightweight MLP,
$\hat{\mathbf h}_{s,b}=\operatorname{softmax}(\Psi_{\rm pres}^{(s)}(\mathbf v_{s,b}))$.
The head predicts crop-level voxel fractions; foreground normalization is applied only when constructing the routing representation.
Let $\mathcal T$ denote the foreground normalization defined below, and write $\widetilde{\mathbf h}=\mathcal T(\mathbf h)$. We supervise the raw and foreground-normalized composition using:
\begin{equation}
 \mathcal L_{\rm comp}^{(s)}=\frac1B\sum_b
 \left[\operatorname{KL}(\mathbf h_b\Vert\hat{\mathbf h}_{s,b})+
 \operatorname{KL}(\mathcal T(\mathbf h_b)\Vert \mathcal T(\hat{\mathbf h}_{s,b}))\right].
\end{equation}

\noindent
\textbf{Anatomical routing representation.} For a composition vector
$\mathbf h$ over background and foreground classes, let
$m_{\rm fg}(\mathbf h)=\sum_{j=1}^{C-1}h_j$. The routing representation is defined as follows:
\begin{equation}
[\mathcal T(\mathbf h)]_c =
\begin{cases}
h_c/m_{\rm fg}(\mathbf h),
    & c\geq1,\quad m_{\rm fg}(\mathbf h)>0,\\
0,
    & c=0,\quad m_{\rm fg}(\mathbf h)>0,\\
h_c,
    & m_{\rm fg}(\mathbf h)=0.
\end{cases}
\end{equation}
The transformation removes background and normalizes the remaining organ fractions. When the crop contains no foreground, the raw distribution is retained; the implementation uses a small positive tolerance for this fallback. The same normalization is used for label-derived and predicted composition. During pre-training, routing uses label-derived or detached predicted composition according to the schedule in Appendix~\ref{sec:objective_details}; inference uses predicted composition only. The resulting representation preserves the relative extent of visible organs and supplies anatomical context to the router. Expert weights also depend on visual features and learned affinities; normalization alone does not impose equal expert usage or a one-to-one organ--expert correspondence.

\noindent
\noindent
\textbf{Representation-aware gating.} Each stage $s$ has $K_s$ experts
and a learned class-to-expert affinity matrix
$\mathbf A^{(s)}\in\mathbb R^{C\times K_s}$.
Its row-wise softmax, $\widetilde{\mathbf A}^{(s)}$, assigns a
distribution over experts to each anatomical class. Weighting these
distributions by the normalized anatomical histogram
$\widetilde{\mathbf h}$ yields the structural expert prior:
\begin{equation}
\mathbf p_s
=
\widetilde{\mathbf h}\,\widetilde{\mathbf A}^{(s)}
\;\in\;(\Delta^{K_s-1})^B.
\end{equation}
The gating logits combine a joint visual--anatomical route,
a direct anatomical route, and the structural prior:
\begin{equation}
\boldsymbol\ell_s =
\underbrace{
\Psi_{\rm gate}^{(s)}
\!\left([\mathbf v_s\Vert\psi^{(s)}(\widetilde{\mathbf h})]\right)
}_{\text{visual--anatomical route}}
+
\underbrace{
\Psi_{\rm anat}^{(s)}(\widetilde{\mathbf h})
}_{\text{direct anatomical route}}
+
\underbrace{
\beta_s\log(\mathbf p_s+\epsilon)
}_{\text{structural affinity prior}}.
\end{equation}
Here $\psi^{(s)}$ embeds the anatomical routing representation,
$\Psi_{\rm gate}^{(s)}$ maps the concatenated context to $K_s$ logits,
and $\Psi_{\rm anat}^{(s)}$ provides a direct anatomy-to-logit path.
The learned positive coefficient $\beta_s$ controls the strength of the anatomical prior, while $\epsilon$ is a small constant
ensuring numerical stability.
The normalized expert mixture weights are:
\begin{equation}
\mathbf W_s
=
\operatorname{softmax}
\!\left(\boldsymbol\ell_s/T_{\rm route}\right),
\qquad T_{\rm route}=0.7.
\end{equation}
These affinities provide anatomical conditioning without imposing
a one-to-one correspondence between organs and experts.

\begin{wrapfigure}{r}{0.5\textwidth}
\vskip 0.14in
\begin{minipage}{0.5\textwidth}
\begin{center}
\includegraphics[width=\textwidth]{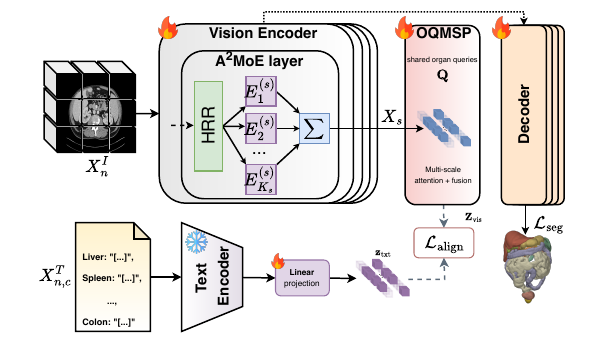}
\end{center}
\caption{\textbf{Overview of the A$^{2}$MoE architecture.}
Given a 3D CT input $\mathbf X_n^{I}$, the Histogram Representation
Router (HRR) combines visual context and anatomical composition
to predict stage-specific expert weights shared across spatial
positions. These weights combine the outputs of all experts
$\{E_k^{(s)}\}_{k=1}^{K_s}$.
OQMSP uses shared organ queries $\mathbf Q$ to aggregate multi-scale
encoder features $\mathbf X_s$ into organ-level visual embeddings
$\mathbf Z_{\rm vis}$. These are aligned with corresponding text
embeddings $\mathbf z_{\rm txt}$, obtained through a frozen text
encoder and a trainable linear projection, using
$\mathcal L_{\rm align}$.
A parallel decoder combines encoder features with spatial skip
connections and is supervised by $\mathcal L_{\rm seg}$.
At inference, anatomical composition is predicted from image features.}
\label{fig:overview}
\vspace{-0.5in}
\end{minipage}
\end{wrapfigure}

\noindent
\textbf{Expert aggregation.} Each stage holds $K_s$ bottleneck MLP experts $\{E_k^{(s)}\}_{k=1}^{K_s}$ implemented as $1\!{\times}\!1\!{\times}\!1$ convolutions over channels. With $\hat{\mathbf{X}}_s = \operatorname{InstanceNorm}(\mathbf{X}_s)$,
\begin{equation}
\mathbf{Y}_{s,b} \;=\; \mathbf{X}_{s,b} \;+\;
\sum_{k=1}^{K_s} W_{s,b,k}\,E_k^{(s)}(\hat{\mathbf{X}}_{s,b}).
\end{equation}

\noindent
\textbf{Computational cost.} Let $M_s=D_sH_sW_s$ be the number of spatial positions at stage $s$, and let $d_s^{\rm gate}$ be the width entering a linear expert-logit projection. Per sample, evaluating this projection at every token costs $\mathcal O(M_s d_s^{\rm gate}K_s)$, whereas one crop-level evaluation costs $\mathcal O(d_s^{\rm gate}K_s)$. These bounds concern logit evaluation after context aggregation: pooling still reads the spatial feature map, and anatomy embedding and affinity-prior computations contribute additional per-crop work. All experts are applied to the feature map, so the savings concern repeated routing computations rather than a sparse reduction in expert FLOPs.

\subsection{The A$^2$MoE VLM architecture}
\label{sec:architecture}

A$^2$MoE-VLM is based on a 3D CNN-Mamba vision encoder, a frozen biomedical text encoder, and OQMSP that aligns per-organ visual features with the corresponding organ-specific report findings at every encoder stage (see Figure~\ref{fig:overview}).

\noindent
\textbf{Vision encoder.} The A$^{2}$MoE Block (A$^{2}$MoEB) extends the SegMamba~\cite{xing2024segmamba} layer with a Gated Spatial Convolution (GSC) \cite{xing2024segmamba} module and the A$^{2}$MoE layer (Sec.~\ref{sec:a2moe}).  Given a volume $\mathbf{X}\in\mathbb{R}^{B\times 1\times D\times H\times W}$, a stem produces the initial feature map in $\mathbb{R}^{B\times 48\times D/2\times H/2\times W/2}$, which is then processed by four A$^{2}$MoEBs with strided downsampling. For the $\ell$-th layer of stage $s$,
\begin{equation}
\mathbf{X}^{\prime\,\ell}_s = f_{\mathrm{GSC}}(\mathbf{X}^{\,\ell}_s),
\quad
\bar{\mathbf{X}}^{\,\ell}_s = f_{\mathrm{Mamba}}\!\bigl(f_{\mathrm{Norm}}(\mathbf{X}^{\prime\,\ell}_s)\bigr)+\mathbf{X}^{\prime\,\ell}_s,
\quad
\mathbf{X}^{\,\ell+1}_s = {f_{\mathrm{A^2MoE}}}\!\bigl(f_{\mathrm{Norm}}(\bar{\mathbf{X}}^{\,\ell}_s)\bigr)+\bar{\mathbf{X}}^{\,\ell}_s,
\end{equation}
where $f_{\text{Norm}}$ denotes instance normalization. Features undergo spatial linearization to form a 1D sequence prior to $f_{\text{Mamba}}$ and are subsequently restored to their volumetric dimensions before $f_{\text{A$^{2}$MoE}}$. We write $\mathbf{X}_s$ for the MoE-refined output of stage $s$.

\noindent
\textbf{Text encoder.} Each per-organ finding $t$ is encoded by a frozen PubMedBERT~\cite{gu2021domain} backbone. Token hidden states $\{\mathbf{u}_j\}$ are averaged with the attention mask $m_j\in\{0,1\}$ and passed through a trainable linear projection $\phi_{\text{txt}}$:
\begin{equation}
\mathbf{z}_{\text{txt}} \;=\;
\operatorname{L2}\!\left(
\phi_{\text{txt}}\!\Bigl(
\tfrac{\sum_j m_j\mathbf{u}_j}{\sum_j m_j}\Bigr)
\right)
\;\in\;\mathbb{S}^{d_e-1},
\end{equation}
where $\operatorname{L2}$ denotes $\ell_2$-normalization. Mean pooling aggregates information across content tokens, providing a pooled representation of each finding~\cite{reimers2019sentence}.

\begin{wrapfigure}{r}{0.35\textwidth}
\vskip -0.15in
\begin{minipage}{0.35\textwidth}
\begin{center}
\includegraphics[width=\textwidth]{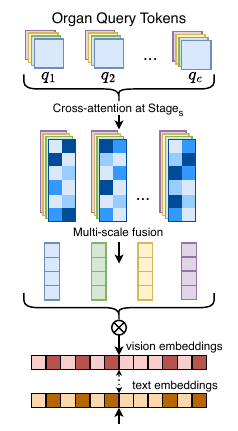}
\end{center}
\vskip -0.2in
\caption{\textbf{An overview of OQMSP.} Learnable organ queries $\{\mathbf q_c\}_{c=1}^{C-1}$ cross-attend to MoE-refined feature volumes at each pyramid stage, producing stage-specific organ embeddings that are fused with learned scale weights. The query attention acts as a mask-free soft-pooling operator, allowing organ-centric visual embeddings to be aligned with text embeddings without requiring segmentation masks at inference.}
\label{fig:oqmsp}
\vspace{-0.5in}
\end{minipage}
\end{wrapfigure}

\noindent
\textbf{Organ-Query Multi-Scale Projector (OQMSP).} Aligning volumetric features with per-organ text requires a per-organ embedding at every scale. Prior multi-scale CLIP-style projectors either pool globally, losing anatomical specificity~\cite{park2026moeclip}, or rely on a ground-truth mask for organ-wise pooling, which can be unavailable at inference or computationally expensive. We instead use a lightweight DETR-style slot projector~\cite{carion2020end} whose forward pass does not use segmentation masks. Masks are used only to define organ-level supervision during pre-training.

We learn $C-1$ organ queries $\mathbf{Q}\in\mathbb{R}^{(C-1)\times d_e}$ (one per non-background class), shared across all four stages, together with per-stage embeddings $\mathbf r_s\in\mathbb R^{d_e}$\footnote{At stage $s$ the key/value $3{\times}3{\times}3$ convolution can use a stride $>1$ to keep attention tractable on the finest-resolution volume.}. At stage $s$, a stage-specific 3D convolution produces keys and values from $\mathbf{X}_s$; $\operatorname{LN}$ denotes layer normalization:
\begin{equation}
[\mathbf{K}_s,\mathbf{V}_s] \;=\;
\operatorname{LN}\!\bigl(\operatorname{Flatten}(\operatorname{Conv3D}_s(\mathbf{X}_s))\bigr)
\;\in\;\mathbb{R}^{B\times N_s\times 2d_e},
\end{equation}
where $N_s$ is the number of key/value tokens after the stage-specific convolution, including any spatial downsampling. A multi-head cross-attention (MHA) block with a residual feed-forward network (FFN) then yields per-stage organ embeddings:
\begin{align}
    &\tilde{\mathbf{Q}}_s = \operatorname{LN}\!\bigl(\mathbf{Q}+\mathbf r_s\bigr), \quad \mathbf{E}_s = \mathbf{Q}+\operatorname{MHA}\!\bigl(\tilde{\mathbf{Q}}_s,\mathbf{K}_s,\mathbf{V}_s\bigr), \\
    &\hat{\mathbf{E}}_s = \operatorname{L2}\!\bigl(\mathbf{E}_s+\operatorname{FFN}_s(\mathbf{E}_s)\bigr) \in \mathbb{R}^{B\times(C-1)\times d_e}.
\end{align}

Sharing $\mathbf Q$ across stages encourages consistent organ representations across resolutions, so the consistency regularizer below becomes an explicit prior on a scale-invariant organ latent. The cross-attention weights additionally provide
an organ-indexed, mask-free soft localization map. Scales are combined with learnable weights normalized by softmax $\boldsymbol{\alpha}=\operatorname{softmax}(a_1,\dots,a_4)$:
\begin{equation}
\mathbf{Z}_{\text{vis}} \;=\;
\operatorname{L2}\!\Bigl(\operatorname{LN}\!\Bigl(\textstyle\sum_{s=1}^{4}\alpha_s\hat{\mathbf{E}}_s\Bigr)\Bigr)
\;\in\;\mathbb{R}^{B\times(C-1)\times d_e}.
\end{equation}
The row $\mathbf{Z}_{\text{vis}}[b,c]$ is the visual embedding of organ $c$ in sample $b$ and is produced by OQMSP without an explicit segmentation-mask or histogram input. At inference, the encoder features remain conditioned on predicted anatomical composition.

\noindent
\textbf{Pre-training objective.} We jointly train with a weighted objective that combines spatial segmentation, anatomy-conditioned routing, organ--text alignment, cross-stage consistency, and query grounding:

\begin{equation}
\mathcal L =
\lambda_{\rm seg}\,\mathcal L_{\rm seg}
+\lambda_{\rm MoE}\,\mathcal L_{\rm MoE}
+\lambda_{\rm align}\,\mathcal L_{\rm align}
+\lambda_{\rm stage}\,\mathcal L_{\rm stage}
+\lambda_{\rm ground}\,\mathcal L_{\rm ground}.
\end{equation}

Here, $\mathcal L_{\rm seg}$ is Dice--cross-entropy supervision for the parallel decoder, and $\mathcal L_{\rm MoE}$ combines crop-composition supervision with regularization of the anatomy-conditioned router. $\mathcal L_{\rm align}$ is symmetric organ-level InfoNCE between OQMSP visual embeddings and their corresponding report findings. $\mathcal L_{\rm stage}$ penalizes cosine disagreement between same-organ embeddings from different encoder stages, while $\mathcal L_{\rm ground}$ uses a KL loss to align normalized query attention with downsampled organ pseudo-masks. Pseudo-masks supervise the decoder and query attention during pre-training; the router and OQMSP do not require masks at inference.
Full objective definition details appear in Appendix~\ref{sec:objective_details}.

\section{Experiments}
\subsection{Experimental setup}

\noindent
\textbf{Datasets.} We pre-train A$^2$MoE on a curated dataset of 67,331 multi-center, multi-phase CT scans from 37,989 patients. These data are sourced from the public Merlin \cite{blankemeier2026merlin,bassi2026merlin} and CancerVerse \cite{li2026cancerverse} datasets and one proprietary dataset. All scans are paired with corresponding radiology reports. We provide comprehensive dataset specifications in Appendix \ref{sec:appendixdatasets}.

\noindent
\textbf{Evaluation metrics.} We compare the performance of state-of-the-art pre-training methods for various downstream tasks, including multi-disease classification, detection, and segmentation. For classification and detection, following \cite{hamamci_generalist_2026,shuilarge}, we evaluate using specificity, sensitivity, and ROC-AUC. For segmentation, we use the Dice similarity coefficient (DSC) and the 95th-percentile Hausdorff distance (HD95).

\noindent
\textbf{Implementation details.} Following \cite{blankemeier2026merlin}, we resample each CT scan to $1.5\! \times\! 1.5\! \times\! 3.0$ $\text{mm}^{3}$ anisotropic resolution. For intensity normalization, all CT scans are clipped to the Hounsfield unit (HU) window of $[-1000, 1000]$ and scaled linearly to $[0, 1]$. During pre-training, $192\!\times\! 192 \!\times\! 96$ foreground-biased crops are used as model inputs. Anatomy histograms are computed from the visible label voxels inside each crop; organ-text contrastive pairs are formed only for organs with sufficient crop support, while the presence head is trained on the crop-level histogram. Detailed pre-training and fine-tuning settings are provided in Appendix~\ref{sec:experimentalsetup}.

\begin{table}[t]
\centering
\caption{\textbf{Segmentation performance results for state-of-the-art methods and ablations of key components.} We evaluate across six datasets, spanning three modalities: CT, MRI, and US. We evaluate performance using DSC ($\uparrow$, \%) and HD95 ($\downarrow$, mm), with the best results shown in \textbf{bold}. The table compares the SegMamba baseline (\#1) against our proposed components, A$^2$MoE (\#2) and OQMSP (\#3). $^{\dagger}$ indicates pre-trained models. $^{\ddagger}$ Measured using $128\times128\times128$-voxel patches.}
\label{tab:segmentationbenchmark}
\resizebox{\textwidth}{!}{%
\scriptsize
\begin{tabular}{l@{\hspace{6pt}}c@{\hspace{4pt}}c cc cc cc cc cc cc}
\toprule
\multirow{2}{*}{Method} & & & \multicolumn{2}{c}{AMOS-CT \cite{ji2022amos}} & \multicolumn{2}{c}{PANORAMA \cite{alves2026artificial}} & \multicolumn{2}{c}{AMOS-MRI \cite{ji2022amos}} & \multicolumn{2}{c}{GEPAR3D \cite{szczepanski2025gepar3d}} & \multicolumn{2}{c}{MRISegmenter \cite{zhuang2025mrisegmenter}} & \multicolumn{2}{c}{MVSeg \cite{carnahan2021deepmitral}} \\
\cmidrule(lr){4-5} \cmidrule(lr){6-7} \cmidrule(lr){8-9} \cmidrule(lr){10-11} \cmidrule(lr){12-13} \cmidrule(lr){14-15}
& Params & VRAM$^{\ddagger}$ & DSC & HD95 & DSC & HD95 & DSC & HD95 & DSC & HD95 & DSC & HD95 & DSC & HD95 \\
\midrule
VoCo$^{\dagger}$\cite{wu2024voco,wu2025large} & 53.0 & 4.4 & 87.1 & 16.4 & 71.6 & 15.9 & 80.1 & 10.5 & 83.5 & 6.8 & 73.0 & 7.7 & 84.3 & 5.1\\
uC 3DU-Net \cite{huang2024upping} & 21.7 & 3.1 & 83.1 & 16.0 & 76.6 & 8.4 & 74.4 & 16.9 & 83.8 & 12.5 & 76.7 & 5.9 & 83.9 & 4.7\\
SuPreM$^{\dagger}$ \cite{liwell} & 62.2 & 4.4 & 87.1 & 7.7 & 76.6 & \textbf{4.7} & 75.8 & 19.5 & 80.6 & 11.0 & 70.8 & 8.4 & 84.3 & 4.9\\
Mamba-HoME \cite{plotka2026mamba} & 170.1 & 3.7 & 86.3 & 18.7 & 77.5 & 4.8 & 77.7 & 9.9 & 86.1 & 6.5 & 75.2 & 6.5 & 84.5 & 4.8\\ \midrule
Merlin$^{\dagger}$ \cite{blankemeier2026merlin} & 252.3 & 3.0 & 85.6 & 29.8 & 77.5 & 9.4 & 79.1 & 10.4 & 86.5 & \textbf{2.7} & 75.8 & 6.0 & 84.1 & \textbf{3.9}\\
fVLM$^{\dagger}$  \cite{shuilarge} & 110.0 & 0.6 & 85.2 & 16.4 & 76.8 & 9.8 & 78.1 & 10.9 & 84.1 & 10.1 & 72.0 & 8.1 & 80.2 & 5.0\\
BTB3D$^{\dagger}$  \cite{hamamci2025better} & 101.1 & 1.4 & 86.1 & 18.3 & 77.3 & 9.2 & 77.8 & 11.1 & 84.0 & 9.8 & 76.9 & 6.2 & 84.6 & 4.0\\ \midrule
\rowcolor{blue!10} \#1: SegMamba \cite{xing2024segmamba} & 66.8 & 2.9 & 85.5 & 17.3 & 76.0 & 8.6 & 76.7 & 9.3 & 85.3 & 10.6 & 74.5 & 6.8 & 83.8 & 5.8 \\
\rowcolor{blue!10} \#2: \hspace{0.5em} + A$^{2}$MoE & 93.3 & 3.0 & 87.5 & 7.6 & 78.2 & 7.1 & 78.5 & 12.2 & 86.1 & 7.9 & 77.6 & 5.9 & 84.5 & 4.4\\
\rowcolor{blue!10} \#3: \hspace{1em} \textbf{+ OQMSP}$^{\dagger}$ & 104.9 & 3.1 & \textbf{88.7} & \textbf{6.2} & \textbf{79.1} & 5.8 & \textbf{82.3} & \textbf{8.9} & \textbf{87.7} & 8.1 & \textbf{79.8} & \textbf{5.5} & \textbf{85.4} & 4.0\\
\bottomrule
\end{tabular}%
}
\end{table}

\subsection{Comparison with state-of-the-art methods}

We compare our method with several state-of-the-art segmentation models, including VoCo \cite{wu2024voco,wu2025large}, SegMamba \cite{xing2024segmamba}, Mamba-HoME \cite{plotka2026mamba}, SuPreM \cite{liwell}, and uC 3DU-Net \cite{huang2024upping}. We also evaluate against MedVLM-based methods, specifically Merlin \cite{blankemeier2026merlin}, fVLM \cite{shuilarge}, BTB3D \cite{hamamci2025better}, RadGPT \cite{bassi2025radgpt}, and CT-CLIP \cite{hamamci_generalist_2026}. For classification and detection tasks, we compare against CT-CHAT \cite{hamamci_generalist_2026}, CT2Rep \cite{hamamci2024ct2rep}, M3D \cite{bai2024m3d}, and RadFM \cite{wu2025towards}. To ensure a fair comparison, we trained each model using consistent hardware settings and data augmentations \cite{eisenmann2023winner}. All hyperparameters were set according to their original papers.

\subsection{Downstream tasks} 

To demonstrate the robustness and capability of our pre-trained encoder, we conduct additional experiments on several downstream tasks, including segmentation, classification, and detection.

\noindent
\textbf{Segmentation.} We integrate the pre-trained encoder with a UNETR-based decoder \cite{hatamizadeh2022unetr}. These experiments utilize both in-distribution and out-of-distribution datasets across multiple modalities, including AMOS-CT/MRI \cite{ji2022amos}, PANORAMA \cite{alves2026artificial}, GEPAR3D \cite{szczepanski2025gepar3d}, MRISegmenter \cite{zhuang2025mrisegmenter}, and MVSeg \cite{carnahan2021deepmitral}. Table \ref{tab:segmentationbenchmark} shows quantitative segmentation performance results. A$^{2}$MoE demonstrates superior performance across diverse imaging modalities, consistently outperforming established state-of-the-art models. Quantitative evaluation indicates that the architecture is uniquely capable of segmenting both fine-grained anatomical details and large-scale structures with high precision.

\begin{figure}[t]
    \centering
    \includegraphics[width=\linewidth]{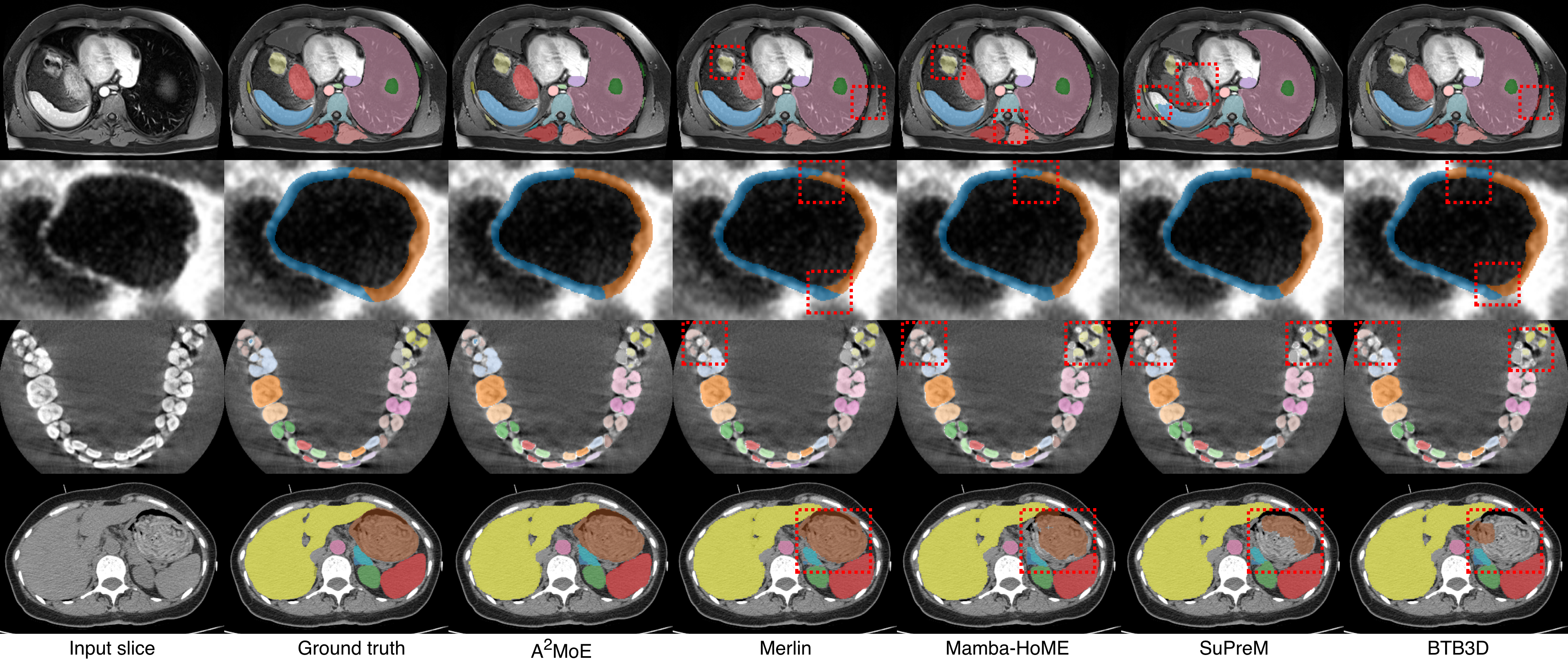}
    \caption{\textbf{Qualitative segmentation results.} Comparison of A$^{2}$MoE against the top four baseline methods. Key differences are highlighted in red bounding boxes (see Appendix \ref{appendix:qualitative_results} for more samples).}
    \label{fig:qualitative_main}
\end{figure}

To assess the robustness of the observed performance gains,  we use critical-difference diagrams (see Figure \ref{fig:cd_combined_statistical_tests}). We consider differences statistically significant at $p < 0.05$. Qualitatively, A$^{2}$MoE produces anatomically consistent results that are largely free from the artifacts typically found in competing methods (see Figure~\ref{fig:qualitative_main}). Furthermore, the architecture maintains high reliability across both in-distribution and out-of-distribution datasets, demonstrating a robust ability to generalize across varying object scales and data acquisition protocols.

\noindent
\textbf{Classification.} Figure~\ref{fig:classification_mean} and Table~\ref{tab:ratic_per_grade} evaluate abdominal trauma classification on RATIC~\cite{rudie2024rsna}. A$^2$MoE achieves the best mean ROC-AUC among the evaluated encoders and improves most strongly on bowel and liver injury, where organ-localized context is clinically important. These results support the use of anatomy-conditioned representations beyond segmentation, while kidney and spleen results show that the gains are not uniform across every organ and metric.

\noindent
\textbf{Detection.} Figure~\ref{fig:detection_plot} and Table~\ref{tab:detection_results} evaluate tumor detection on AbdomenAtlas~3.0 \cite{bassi2025radgpt}. A$^{2}$MoE is most competitive for small lesions ($\le$2 cm), improving sensitivity for kidney and liver tumors and matching the best pancreatic sensitivity among the evaluated models. The specificity--sensitivity trade-off differs by organ, especially for kidney tumors, so we interpret these results as evidence that anatomy-conditioned features improve small-lesion sensitivity rather than as a uniformly dominant detector.

\subsection{Ablation studies}

We investigate the impact of several factors on the performance of A$^2$MoE, focusing on: (1) key architectural components; (2) histogram-aware routing; and (3) the number of experts per encoder stage. For a fair comparison, we maintain the same training dataset and hyperparameters across all experimental configurations to isolate the impact of each component. Detailed ablation studies are provided in Appendix \ref{sec:appendixablationstudies}.

\noindent
\textbf{Effect of the key components.} We evaluate the contribution of the two main components of our segmentation model: A$^{2}$MoE routing and contrastive alignment. We compare three configurations: (i) the SegMamba baseline~\cite{xing2024segmamba} (\#1);
(ii) SegMamba with A$^{2}$MoE (\#2); and (iii) the full model combining A$^{2}$MoE and contrastive alignment (OQMSP).
As shown in Table~\ref{tab:segmentationbenchmark}, A$^{2}$MoE improves over the baseline across datasets, while the full model gives the strongest overall
performance. These results indicate that anatomy-activated routing and contrastive alignment provide complementary benefits for segmentation tasks.

\begin{wraptable}{r}{0.7\textwidth}
\vskip -0.26in
\begin{minipage}{0.7\textwidth}
\scriptsize
\caption{\textbf{Ablation of histogram-aware MoE routing.} We report DSC
(\%) with HD95 (mm) in parentheses, with the best \textbf{bolded}.}
\resizebox{\linewidth}{!}{
\begin{threeparttable}
\begin{tabular}{lcccc}
\toprule[1pt]
Config & AMOS-CT & AMOS-MRI & GEPAR3D & MVSeg \\
\midrule[0.5pt]
MoE (Baseline) & 86.0 (14.4) & 77.4 (14.5) & 85.6 (9.8) & 84.0 (5.0) \\
$\hspace{0.5em} + \psi^{(s)}(\widetilde{\mathbf h})$ & 87.0 (12.0) & 77.9 (14.0) & 85.8 (8.5) & 84.3 (4.8) \\
\rowcolor{blue!10} $\hspace{1em} + \widetilde{\mathbf h}\widetilde{\mathbf A}^{(s)} (\mathbf{A}^2\text{MoE})$ &\textbf{87.5 (7.6)} & \textbf{78.5 (12.2)} & \textbf{86.1 (7.9)} & \textbf{84.5 (4.4)} \\
\bottomrule[1pt]
\end{tabular}
\end{threeparttable}
}
\label{tab:moerouting_ablations}
\end{minipage}
\vskip -0.15in
\end{wraptable}

\noindent
\textbf{Effect of histogram-aware routing.} We ablate the two anatomy-conditioned components of the router: the histogram embedding $\psi^{(s)}(\widetilde{\mathbf h})$ and the structural affinity prior $\beta_s\log(\widetilde{\mathbf h}\widetilde{\mathbf A}^{(s)}+\epsilon)$. Table~\ref{tab:moerouting_ablations} shows that both components improve segmentation consistently across datasets. The gains in DSC are moderate but
stable, while the reductions in HD95 are larger, indicating that histogram-aware routing primarily improves boundary robustness and reduces distant false
positives rather than only increasing volumetric overlap. Adding the affinity prior gives the strongest improvement, suggesting that explicit class-to-expert
structure provides useful anatomical regularization for MoE routing.

\begin{wraptable}{r}{0.7\textwidth}
\vskip -0.26in
\begin{minipage}{0.7\textwidth}
\scriptsize
\caption{\textbf{Ablation study results with various numbers of experts per stage.} We report mDSC (\%). The best values are \textbf{bolded}.}
\label{tab:experts_ablations}
\resizebox{\linewidth}{!}{
\begin{threeparttable}
\setlength{\tabcolsep}{4pt}
\begin{tabular}{l c c c c c c c c} 
\toprule[1pt]
\multirow{2}{*}{\#} & \multirow{2}{*}{Experts} & \multirow{2}{*}{Params} & \multicolumn{6}{c}{mDSC (\%)} \\
\cmidrule(l){4-9}
 &  &  & A-CT & PANO & A-MRI & GEP & MRISeg & MVSeg \\
\midrule[0.5pt]
1 & [4, 4, 4, 4] & 83.8 & \textbf{87.5} & 77.9 & 78.0 & 85.9 & \textbf{77.7} & 84.8 \\
2 & [8, 8, 8, 8] & 87.0 & 87.3 & 75.1 & 77.9 & 85.0 & 77.5 & \textbf{85.0} \\
3 & [4, 8, 16, 32] & 102.3 & \textbf{87.5} & 77.4 & 77.4 & 86.1 & 77.6 & 84.5 \\
\rowcolor{blue!10} 4 & [16, 16, 16, 16] & 93.3 & 86.9 & \textbf{78.2} & \textbf{78.5} & \textbf{86.2} & 77.4 & 84.9 \\
\bottomrule[1pt]
\end{tabular}
\end{threeparttable}}
\end{minipage}
\vskip -0.15in
\end{wraptable}

\noindent
\textbf{Effect of the number of experts.} We evaluate how expert count and stage-wise allocation impact performance. As shown in Table~\ref{tab:experts_ablations}, the uniform $[16,16,16,16]$ configuration offers the best balance, achieving state-of-the-art results on three datasets while remaining competitive across others. Smaller uniform setups lack sufficient capacity, and pyramid allocations exhibit instability across modalities. Consequently, we adopt the $[16,16,16,16]$ configuration as our default setup.

\section{Conclusions}

We introduced A$^{2}$MoE, a 3D MedVLM that incorporates explicit anatomical priors into both conditional computation and cross-modal alignment. The proposed HRR grounds expert selection in crop-level class histograms combined with a learnable class-to-expert affinity matrix, producing a gating distribution per crop and stage. Furthermore, the OQMSP aligns MoE-refined visual features with organ-specific textual representations through scale-shared queries. At inference, anatomical composition is predicted from image features, allowing the visual branch to operate without segmentation masks or report text. Pre-trained on a large-scale cohort of over $60{,}000$ CT scans, A$^{2}$MoE achieves state-of-the-art performance on external benchmarks for segmentation, classification, and detection. Moreover, it demonstrates strong transfer to MRI and ultrasound, suggesting that anatomical grounding serves as a modality-agnostic prior.

\noindent
\textbf{Limitations and future directions.} Despite its performance, A$^{2}$MoE has certain limitations. First, the current framework relies on a fixed set of anatomical classes for the composition histogram; while this covers major organ systems, it may require adaptation for highly granular or rare pathological structures not represented in the pre-training prior. Second, while the model exhibits strong cross-modality transfer, its performance is inherently tied to the quality and diversity of the initial CT-report pairs. Finally, A$^{2}$MoE focuses on static 3D volumes; extending this anatomical conditioning to longitudinal data remains an open challenge for future research.

\section*{Acknowledgments}

This research was funded in part by National Science Center, Poland 2025/57/B/ST6/03100. We gratefully acknowledge Polish high-performance computing infrastructure PLGrid (HPC Center: ACK Cyfronet AGH) for providing computer facilities and support within computational grant no. PLG/2025/018792. 

%
%
\bibliographystyle{plain}
\bibliography{main}

\newpage
\appendix

\renewcommand \thepart{}
\renewcommand \partname{}

\part{Appendix} 
\hypersetup{linkcolor=black, urlcolor=black, citecolor=black}
\setcounter{secnumdepth}{4}
\setcounter{tocdepth}{4}
\parttoc 
\hypersetup{
	colorlinks=true,
	linkcolor=red,
	urlcolor=magenta,
	citecolor=blue,
}

\clearpage

\section{Data pre-processing}
\label{sec:appendixdatapreprocessing}

\subsection{Anatomy parsing}

\begin{figure}[t]
    \centering
    \includegraphics[width=\linewidth]{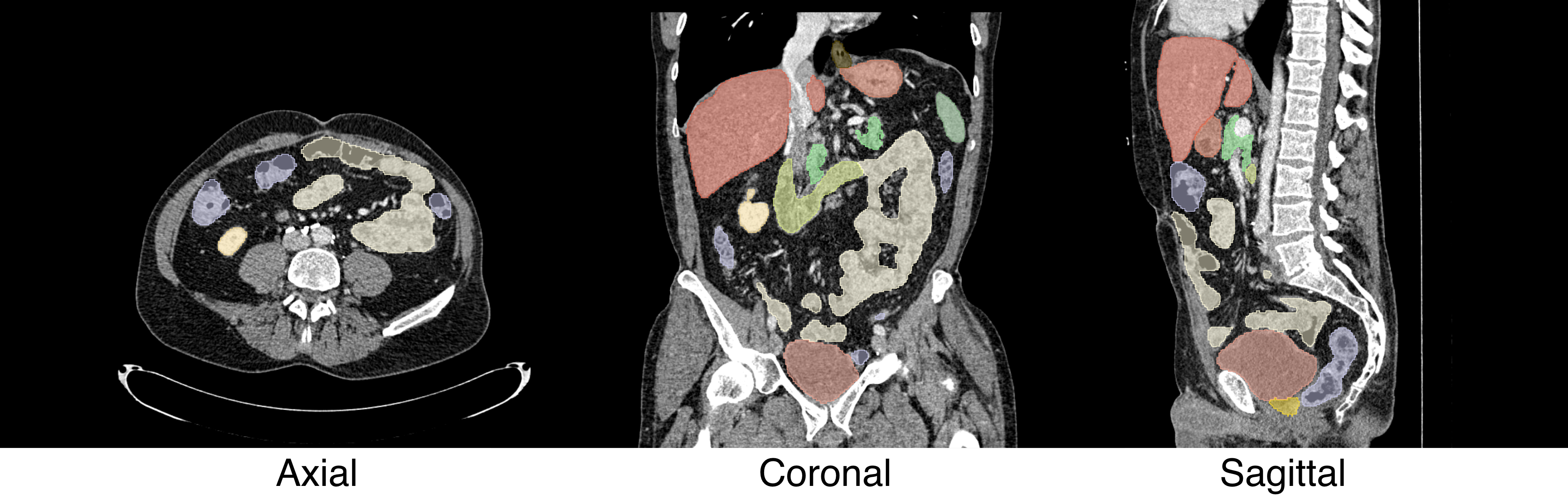}
    \caption{Anatomical segmentation results obtained with TotalSegmentator \cite{wasserthal2023totalsegmentator} across axial, coronal, and sagittal planes for 14 diverse abdominal and pelvic structures. The visualization demonstrates the model’s spatial consistency across orthogonal views for organs of varying scales, ranging from large structures (e.g., liver, colon) to small, high-variance targets (e.g., adrenal glands, duodenum). These high-fidelity masks serve as a structured anatomical prior, providing localized supervision for image-text alignment and ensuring that the joint embedding space preserves the underlying biological topology.}
    \label{fig:ts_segmentation_results}
\end{figure}

\begin{wraptable}{r}{0.4\textwidth} 
    \centering
    \vspace{-0.26 in}
    \caption{\textbf{Anatomical pseudo-labels.} Overview of the 14 anatomical classes extracted through TotalSegmentator \cite{wasserthal2023totalsegmentator} to serve as auxiliary supervision. These classes encompass a diverse range of abdominal and pelvic structures with varying morphologies. These masks spatially ground the model, aligning radiological tokens with anatomical regions to preserve biological topology in the latent space.}
    \label{tab:selectedclasses}
    \small 
    \begin{tabular}{lc}
        \toprule
        \textbf{Class} & \textbf{Anatomy} \\
        \midrule
        0 & Background \\
        1  & Spleen \\
        2  & Left kidney \\
        3  & Right kidney \\
        4  & Gallbladder \\
        5  & Liver \\
        6  & Stomach \\
        7  & Pancreas \\
        8  & Left adrenal gland \\
        9  & Right adrenal gland \\
        10 & Duodenum \\
        11 & Small bowel \\
        12 & Colon \\
        13 & Urinary bladder \\
        14 & Prostate \\
        \bottomrule
    \end{tabular}
\end{wraptable}

\begin{figure}[h]
    \centering
    \includegraphics[width=\linewidth]{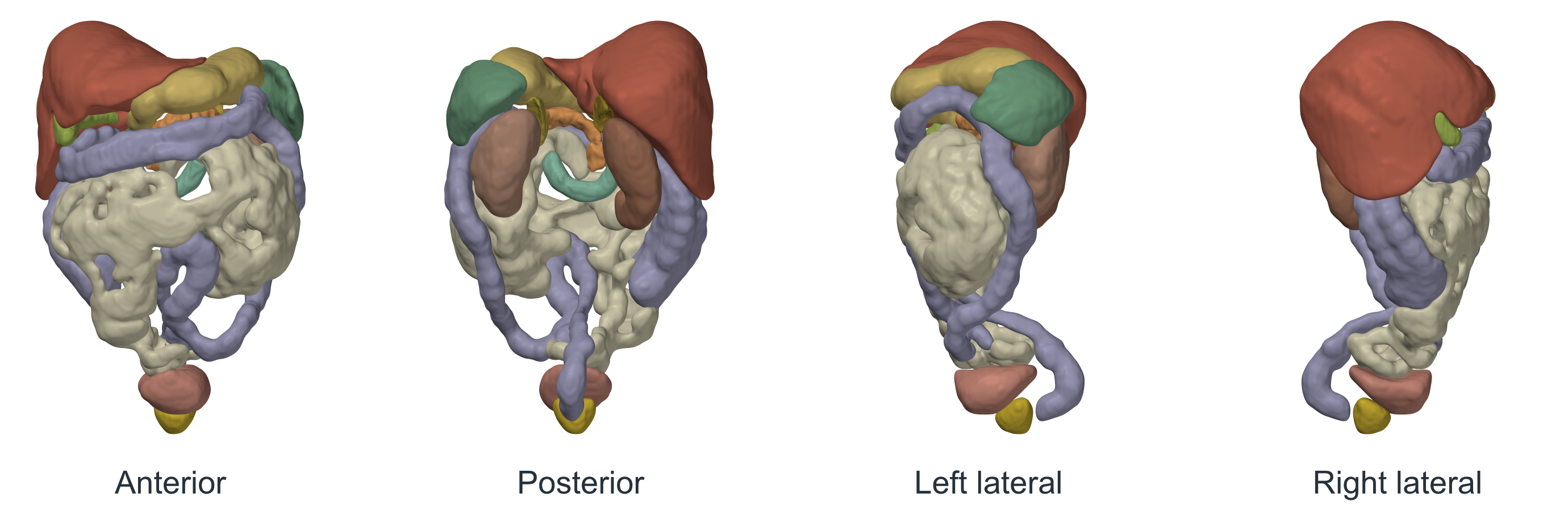}
    \caption{Visualization of the 3D projections of the pseudo-labels generated with TotalSegmentator \cite{wasserthal2023totalsegmentator}, shown from the anterior, posterior, left lateral, and right lateral views. These projections highlight the global structural consistency and anatomical coherence of the pseudo-labels (see Table~\ref{tab:selectedclasses}). The multi-view renderings provide a qualitative assessment of the overall segmentation quality across the full 3D volume.}
    \label{fig:3dvisualizations}
\end{figure}

We leverage the pre-trained TotalSegmentator~\cite{wasserthal2023totalsegmentator} to generate voxel-wise pseudo-labels (see Figure~\ref{fig:ts_segmentation_results}). Note that the pseudo-labels and radiology organ snippets were verified by experienced radiologists; see Sec.~\ref{sec:verificationsnippets} for details. Although the original model encompasses 117 anatomical structures, we directly extract 14 key abdominal organs, including \textit{spleen, left kidney, right kidney, gallbladder, liver, stomach, pancreas, left adrenal gland, right adrenal gland, duodenum, small bowel, colon, urinary bladder, and prostate} (see Table~\ref{tab:selectedclasses}). This pruning process serves two primary purposes: first, it filters out structures irrelevant to our domain-specific radiology reports (e.g., thoracic organs); and second, it removes low-prevalence classes (e.g., skeletal and vascular structures) that introduce unnecessary noise. This semantic alignment bridges the gap between high-resolution volumetric segmentations and the coarser, descriptive terminology utilized in clinical practice. By consolidating these structures, we reduce computational complexity and ensure the segmentation output is clinically actionable, facilitating better integration with standardized reporting frameworks (see Figure~\ref{fig:3dvisualizations}).

The adoption of these masks also addresses a fundamental challenge in medical vision-language modeling: the inherent scarcity and spatial ambiguity of lesion localization in clinical text. Pathological findings in specific anatomical regions often follow a long-tailed distribution, leading to an acute class imbalance during training. Without such anatomical grounding, many training iterations would consist of "normal" samples, potentially leading to representation collapse or a bias toward the majority class. By incorporating these 14 structures as an auxiliary spatial prior, we provide the model with a robust framework to learn meaningful diagnostic features even in the presence of sparse or imbalanced clinical labels.

\subsubsection{Report decomposition}
\label{appendix:report_decomposition}

Raw radiology reports are cleaned and decomposed at the organ level based on the selected anatomical structures. We clean these reports by removing unnecessary information, such as dates, details regarding modalities other than CT, and findings related to previous examinations. Following \cite{shuilarge}, the decomposition is performed as follows:

\begin{enumerate} 
    \item We develop domain-knowledge prompts with guidance from experienced radiologists to ensure comprehensive terminology coverage, preventing missed organ structures arising from synonym variation in clinical reports (see Section \ref{appendix:prompts}); 
    \item We employ Qwen3-30B-A3B-Instruct \cite{yang2025qwen3} for image-text alignment at the organ level using the prompts presented in Section \ref{appendix:prompts}. To accelerate inference, we employ vLLM \cite{kwon2023efficient}; 
    \item We then align the descriptions of each anatomical structure from the \textit{findings} and \textit{impression} sections,
    \item Lastly, for an organ described as \textit{normal}, we use a template describing the absence of pathology (see Section  \ref{appendix:normaltemplates}).
\end{enumerate}

\subsubsection{Diagnosis extraction}

To automate the labeling process, we utilized an LLM (Qwen3-30B-A3B-Instruct), prompted with a domain-specific heuristic to distinguish between clinical pathology and benign anatomical descriptions for each organ (see Table \ref{tab:selectedclasses}). The core challenge in radiological labeling is preventing "false positives" triggered by descriptive text that does not indicate active disease.

We established a strict hierarchy for the \textit{Pathological (1)} label: it is reserved for intrinsic tissue abnormalities, including mass effects, inflammatory processes (e.g., dilatation, stones), or persistent lesions (including stable cysts). Conversely, the \textit{Normal/Unremarkable (0)} label accounts for three specific clinical scenarios:

\begin{enumerate}
    \item Anatomical and Vascular Variants: Non-pathological deviations such as "replaced" arterial supplies,
    \item Stable Post-Surgical States: The absence of an organ or expected post-operative changes without secondary complications,
    \item External Proximity: Displacement or compression of an organ by an adjacent mass (e.g., "modeling") is excluded from the pathological count unless the report explicitly notes tissue infiltration or parenchymal damage.
\end{enumerate}

\subsubsection{Normal templates}
\label{appendix:normaltemplates}

To mitigate the representation collapse often observed in medical domain embeddings, we move beyond the use of monolithic labels like "Normal." Such simplistic categorizations fail to capture the multimodal distribution of healthy anatomical states, leading to an underspecified latent space that lacks the granularity required for robust contrastive learning.

Instead, we synthesize structured normal templates to serve as high-fidelity prototypical anchors. By embedding rich, organ-specific descriptive text rather than a binary state, we force the model to align its representations with detailed physiological features (e.g., organ axis, attenuation, and morphology). This approach ensures that during the contrastive optimization, specifically when minimizing the distance between positive pairs, the model learns a structured embedding space where "normality" is defined by a dense cluster of anatomical constraints. Consequently, pathological deviations are not just "not normal," but are explicitly represented as semantic residuals relative to a highly defined physiological baseline, enhancing the model’s ability to generalize across heterogeneous clinical manifestations.

\begin{tcolorbox}[
  breakable,
  colframe=jsonteal!80,
  colback=jsonteal!4,
  coltitle=white,
  fonttitle=\bfseries\footnotesize,
  title=Normal Templates,
  colbacktitle=jsonteal!80,
  fontupper=\small,
  before skip=6pt,
  after skip=3pt
]

\begin{verbatim}
{
    "Spleen": "The spleen is normal in size and homogeneous, with no 
    splenomegaly or focal lesions identified.",
    "Kidney left": "The kidney left is normal in size and axis. 
    No hydronephrosis, renal calculi, or solid masses.",
    "Kidney right": "The kidney right is normal in size and axis. 
    No hydronephrosis, renal calculi, or solid masses.",
    "Gallbladder": "The gallbladder has a normal caliber. No wall 
    thickening, pericholecystic fluid, or gallstones.",
    "Liver": "The liver is normal in size with smooth contours. No focal 
    hepatic lesions, biliary dilatation, or steatosis.",
    "Stomach": "The stomach shows normal wall thickness with no 
    distension or masses.",
    "Pancreas": "The pancreas demonstrates normal attenuation. 
    No pancreatic duct dilatation or focal masses.",
    "Adrenal gland left": "The adrenal gland left have normal morphology 
    with no nodules or masses.",
    "Adrenal gland right": "The adrenal gland right have normal morphology 
    with no nodules or masses.",
    "Small bowel": "The small bowel loops are normal in caliber 
    without wall thickening or obstruction.",
    "Colon": "The colon is normal in caliber with no wall thickening, 
    obstruction, or free air.",
    "Urinary bladder": "The urinary bladder is well-distended with 
    no wall thickening or masses.",
    "Prostate": "The prostate is normal in size with no enlargement 
    or focal nodules."
}
\end{verbatim}
\end{tcolorbox}

\clearpage

\begin{tcolorbox}[
  breakable,
  colframe=reportblue!80,
  colback=reportblue!4,
  coltitle=white,
  fonttitle=\bfseries\footnotesize,
  title=Original radiology report,
  colbacktitle=reportblue!80,
  fontupper=\small,
  before skip=3pt,
  after skip=6pt
]

\textit{FINDINGS:} Lower thorax: Bibasilar dependent opacities, greater on the right  likely reflect dependent atelectasis.   Liver and biliary tree: Numerous subcentimeter hypodensities are too  small to definitively characterize but are likely to represent benign  entity such as hamartomas or small cysts. Mild to moderate intraand  extrahepatic biliary ductal dilation is likely related to  postcholecystectomy state.   Gallbladder: Surgically absent.   Spleen: Normal.   Pancreas: Normal.   Adrenal glands: Normal.   Kidneys and ureters: Bilateral subcentimeter hypodensities are  indeterminant, likely representing small cysts. Symmetric enhancement without hydronephrosis.   Gastrointestinal tract: Multiple dilated upstream small bowel loops  are present to the level of the left upper quadrant where 2 adjacent  transition points are noted. Bowel loops between these areas  of transition are thickened and have hazy margins. There is marked  infiltration of the mesentery suggesting venous compromise and  ischemia. Distal small bowel loops are collapsed. No pneumatosis.  Colonic diverticulosis without evidence of diverticulitis.   Peritoneal cavity: Small moderate volume of ascites in the right lower quadrant and pelvis. No free air.    Bladder: Normal.   Prostate and seminal vesicles: Normal.   Vasculature: Calcified atherosclerotic disease without evidence of aneurysm.   Lymph nodes: Normal.   Abdominal wall: Normal.   Musculoskeletal: Mild degenerative spine change.
\end{tcolorbox}
\begin{center}
\begin{tikzpicture}[>=Latex]
  \draw[very thick, -{Latex[length=4mm]}, qwenpurple]
    (0,0) -- (0,-1.4);
  \node[right=0.6cm] at (0,-0.7) {%
    \includegraphics[height=5em]{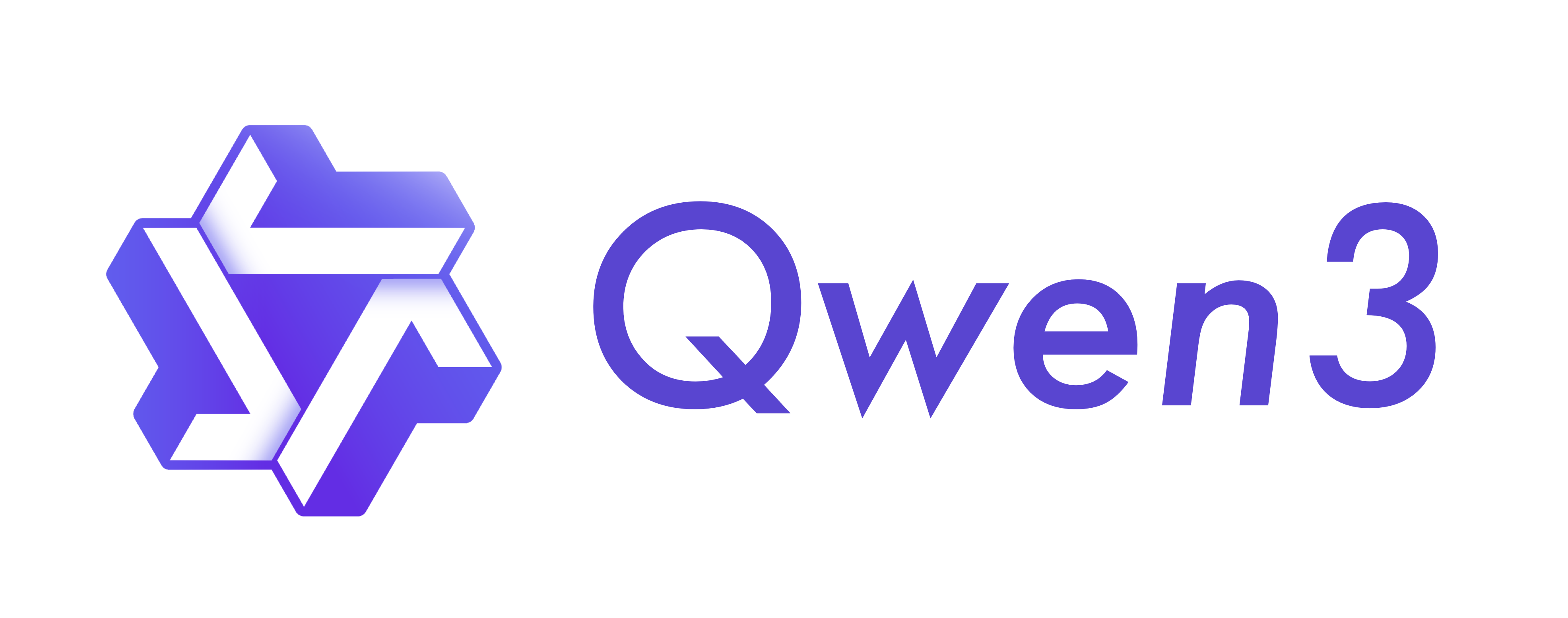}%
    \hspace{0.3em}\scriptsize 
  };
\end{tikzpicture}
\end{center}

\begin{tcolorbox}[
  breakable,
  colframe=jsonteal!80,
  colback=jsonteal!4,
  coltitle=white,
  fonttitle=\bfseries\footnotesize,
  title=Structured radiology report,
  colbacktitle=jsonteal!80,
  fontupper=\small,
  before skip=6pt,
  after skip=3pt
]

\begin{verbatim}
{
  "study_id": "AC423ce5b",
  "cleaned_report": "[... The full cleaned report text is here ...]",
  "organs": {
    "Spleen": "The spleen is normal in size and homogeneous, with no 
    splenomegaly or focal lesions identified.",
    "Kidney left": "The kidney left is normal in size and axis. 
    No hydronephrosis, renal calculi, or solid masses.",
    "Kidney right": "The kidney right is normal in size and axis. 
    No hydronephrosis, renal calculi, or solid masses.",
    "Gallbladder": "The gallbladder is surgically absent.",
    "Liver": "The liver is normal in size with smooth contours. No focal 
    hepatic lesions, biliary dilatation, or steatosis.",
    "Stomach": "Unremarkable",
    "Pancreas": "The pancreas demonstrates normal attenuation. 
    No pancreatic duct dilatation or focal masses.",
    "Adrenal gland left": "The adrenal gland left have normal morphology 
    with no nodules or masses.",
    "Adrenal gland right": "The adrenal gland right have normal morphology 
    with no nodules or masses.",
    "Small bowel": "Multiple dilated upstream small bowel loops present 
    to the level of the left upper quadrant where two adjacent transition points
    are noted. Bowel loops between these areas of transition are thickened
    and have hazy margins. Marked infiltration of the mesentery suggesting 
    venous compromise and ischemia. Distal small bowel loops are collapsed.",
    "Colon": "The colon is normal in caliber with no wall thickening, 
    obstruction, or free air.",
    "Urinary bladder": "The urinary bladder is well-distended with no wall 
    thickening or masses.",
    "Prostate": "The prostate is normal in size with no enlargement 
    or focal nodules."
  }
}
\end{verbatim}
\end{tcolorbox}

\clearpage

\subsubsection{Radiology report translation}

We translate our proprietary radiology reports from Polish to English using Bielik-11B-v3.0-Instruct~\cite{ociepa2025bielik}.

\subsubsection{Verification of anatomy-aware snippets}
\label{sec:verificationsnippets}

\textbf{Pseudo-label correction.} To maximize segmentation reliability, we implement a multi-layered verification framework that integrates clinical ground truth with automated quality-control heuristics. The first layer involves cross-referencing pseudo-masks with the original radiology report to ensure anatomical plausibility. For instance, when a report explicitly documents a "status post-nephrectomy" or "surgically absent" organ, any voxels erroneously detected by TotalSegmentator \cite{wasserthal2023totalsegmentator} for that structure are programmatically reclassified as background. This report-guided correction directly addresses the most common systematic failure mode of deep-learning segmenters: the tendency to misidentify scar tissue, surgical clips, or adjacent structures as an organ that is no longer present.

Building upon this clinical validation, we adopt the methodology of "Label Critic" \cite{bassi2025label} as a formal quality-control mechanism to manage label uncertainty. This framework identifies likely incorrect labels in single-annotation settings by flagging outliers that deviate from learned anatomical priors. Furthermore, in scenarios where multiple segmenters or iterations produce competing outputs, the mechanism acts as an arbiter to select the superior annotation based on spatial consistency and boundary fidelity. By combining NLP-driven anatomical pruning with these formal QC metrics, the pipeline ensures that downstream quantitative analysis, such as radiomics or volumetry, is performed on masks that are verified against the patient’s specific surgical history and radiological reality.

\textbf{Radiology report snippets.} To ensure the clinical accuracy of the dataset, each anatomy-aware snippet underwent manual review by a senior radiologist with over 30 years of clinical experience. The verification process focused on confirming that the extracted segments maintained anatomical integrity and clinical relevance relative to the original imaging reports.

\clearpage

\subsection{Prompts}
\label{appendix:prompts}

\subsubsection{Raw report cleaning}

This cleaning pseudo-prompt serves as a specialized preprocessing heuristic designed to isolate "point-in-time" clinical observations from the longitudinal noise characteristic of radiology narratives. It enforces an aggressive filtration logic that programmatically de-identifies temporal data and excises comparative metadata, such as cross-modality references (e.g., prior MRI or PET/CT) and stability descriptors, which frequently induce contextual bias in downstream extraction models.

By distilling the narrative into a pure anatomical state-of-health summary, the prompt ensures that the model’s attention is localized strictly to current physiological findings and objective measurements. This standardized text-normalization protocol, including its complete rule-set and edge-case handling, is available in our code repository.

\begin{tcolorbox}[
  breakable,
  colframe=jsonteal!80,
  colback=jsonteal!4,
  coltitle=white,
  fonttitle=\bfseries\footnotesize,
  title=Protocol 1: Structured Report Normalization Pipeline,
  colbacktitle=jsonteal!80,
  fontupper=\small,
  before skip=6pt,
  after skip=3pt
]

\begin{verbatim}
ALGORITHM: Radiology Text Normalization
INPUT: Raw CT findings (text)
OUTPUT: Cleaned CT findings (text)

INSTRUCTION SET:
1.  STRIP_TEMPORAL: 
    - Remove <DATE>, YEAR, [0-9]{4}, 'now', 'interval', 'current'.
2.  STRIP_LONGITUDINAL_REFERENCES:
    - Remove all clauses containing: 'prior', 'stable', 'unchanged', 
      'comparison', 'compared to', 'no change'.
3.  STRIP_CROSS_MODAL_NOISE:
    - Remove references to: 'PET/CT', 'MRI', 'ultrasound', 'X-ray', 
    'scintigraphy'.
4.  STRIP_NON_SEMANTIC_SYNTAX:
    - Remove all content within ( ), [ ], < >, { }.
    - Remove billing codes and meta-annotations.
5.  CORE_FINDING_RETAIN:
    - Extract current anatomical findings.
    - Preserve clinical measurements (e.g., '4 mm').
    - Maintain medical terminology and morphology descriptions.

EXECUTION:
- Apply aggressive filtering to prune phrases containing comparative language.
- Ensure natural linguistic flow of remaining medical tokens.
- Output high-density clinical text only.
\end{verbatim}
\end{tcolorbox}

\clearpage

\subsubsection{Organ-aware report decomposition}

\begin{tcolorbox}[
  breakable,
  colframe=jsonteal!80,
  colback=jsonteal!4,
  coltitle=white,
  fonttitle=\bfseries\footnotesize,
  title=Protocol 2: Hierarchical Organ-Specific Feature Extraction,
  colbacktitle=jsonteal!80,
  fontupper=\small,
  before skip=6pt,
  after skip=3pt
]

\begin{verbatim}
ALGORITHM: Segment-then-Map Anatomical Extraction
INPUT: Cleaned CT findings (text), Target Organ List
OUTPUT: Structured JSON Map (Organ: Finding)

INSTRUCTION SET:
1.  LOCALIZE_CONTEXT:
    - Identify sentences specific to each Target Organ.
    - Validate finding ownership (e.g., exclude "peri-pancreatic" from Pancreas).
2.  LATERALIZATION_RESOLVER:
    - Map plural descriptors (e.g., "Kidneys") to both lateral child nodes.
    - Isolate unilateral findings strictly to the specific side (Right/Left).
3.  EVIDENCE_HIERARCHY_DECISION:
    - IF PATHOLOGY_DETECTED:
        - Identify: Calcification, cyst, nodule, stone, surgical change.
        - ACTION: Extract verbatim clinical snippet.
    - ELSE IF EXPLICIT_NORMAL:
        - Identify: "normal", "unremarkable", "no findings".
        - ACTION: Inject EXACT string from Standardized Template Pool.
    - ELSE:
        - ACTION: Assign "Unremarkable" (Indicates omission/absence).
4.  PATHOLOGY_GUARDRAIL:
    - Explicitly inhibit Template Injection if any secondary findings 
      (e.g., fat infiltration, resections, clips) are identified.

EXECUTION:
- Iterate through Target Organ List.
- Perform high-fidelity mapping of sub-segments (e.g., Rectum -> Colon).
- Output strictly validated JSON schema.
\end{verbatim}
\end{tcolorbox}

\clearpage

\subsubsection{Diagnosis extraction}

\begin{tcolorbox}[
  breakable,
  colframe=jsonteal!80,
  colback=jsonteal!4,
  coltitle=white,
  fonttitle=\bfseries\footnotesize,
  title=Protocol 3: Binary Pathological Classification and Scoring,
  colbacktitle=jsonteal!80,
  fontupper=\small,
  before skip=6pt,
  after skip=3pt
]

\begin{verbatim}
ALGORITHM: Binary Pathology Labeling (0/1)
INPUT: Structured JSON Findings Map
OUTPUT: Binary JSON Map (Organ: Int)

SCORING CRITERIA (STRICT):
1.  CLASS_0 (STRICTLY_NORMAL):
    - Match must be 100% identical to the 'Normal Template'.
    - Apply if finding is explicitly 'Unremarkable'.
2.  CLASS_1 (ANY_FINDING):
    - Apply for any deviation from physiological baseline.
    - INCIDENTIALOMAS: Tiny cysts, granulomas, nodules.
    - CALCIFICATIONS: Vascular or parenchymal calcifications.
    - SURGICAL_STATES: Resections, cholecystectomy, clips, stents.
    - MORPHOLOGY: Atrophy, organ absence, wall thickening.

EXECUTION:
- Evaluate each organ entry in the findings map.
- Trigger CLASS_1 for any text not matching the predefined Normal Pool.
- Return a condensed JSON map mapping Organ names to binary integers.
\end{verbatim}
\end{tcolorbox}

\clearpage

\section{Datasets}
\label{sec:appendixdatasets}

\begin{figure}[t]
    \centering
    \includegraphics[width=\linewidth]{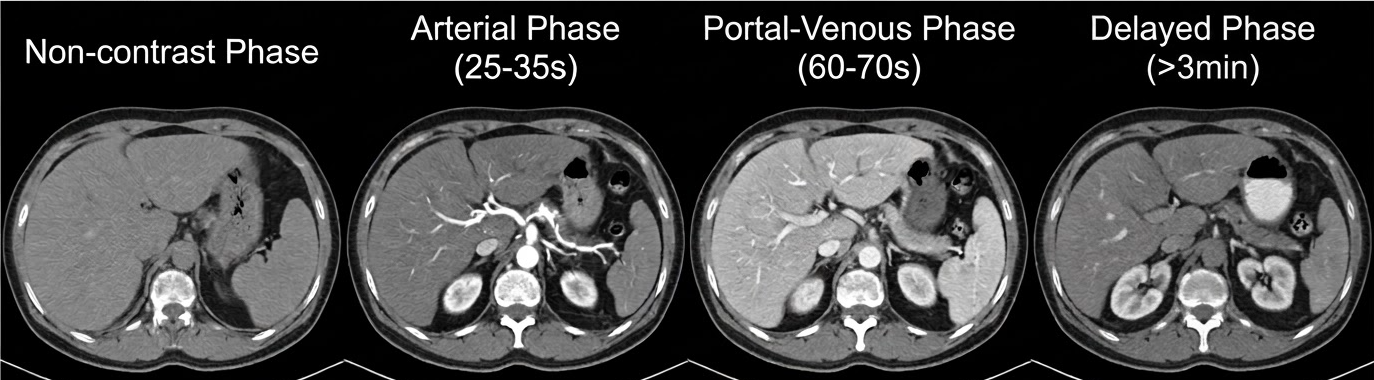}
    \caption{\textbf{Multi-phase contrast-enhanced CT scans.} In our pre-training and evaluation, we utilize multi-phase CT volumes encompassing non-contrast, arterial (25--35~s), portal-venous (60--70~s), and delayed ($>$3 min) phases. We demonstrate the diversity of hemodynamic enhancement patterns across these temporal windows; as illustrated, each phase provides unique voxel intensities for the same anatomical structures. This multi-temporal alignment enables the model to learn robust anatomy-normality representations and capture temporal-spectral dependencies that are typically absent in single-phase medical vision-language datasets.}
    \label{fig:phasesoverview}
\end{figure}

In this section, we provide a detailed description of the datasets used in the pre-training, testing, and fine-tuning stages.

\subsection{Training}

To train our model, we curate a large-scale, multimodal, multi-center dataset comprising both public and proprietary sources. This collection includes 67,331 multi-phase CT scans across 53,643 exams from 37,989 patients, each paired with its corresponding radiology report. Detailed descriptions of the individual cohorts are provided below, and Table \ref{tab:appendix_training_datasets} summarizes the dataset distributions.

\begin{table}[ht]
    \centering
    \caption{\textbf{Distribution of training CT datasets.} We summarize the number of scans, exams, and unique patients across the three cohorts.}
    \label{tab:appendix_training_datasets}
    \begin{tabular}{lccc} 
        \toprule
        \textbf{Dataset} & \textbf{Scans} & \textbf{Exams} & \textbf{Patients} \\ 
        \midrule
        Merlin \cite{blankemeier2026merlin} & 25,494 & 25,494 & 18,317 \\
        CancerVerse \cite{li2026cancerverse} & 24,422 & 22,344 & 14,599 \\
        Proprietary set & 17,415 & 5,805 & 4,004 \\
        \midrule
        \textbf{Total} & 67,331 & 53,643 & 37,989 \\
        \bottomrule
    \end{tabular}
\end{table}

\subsubsection{Merlin} 

Merlin \cite{blankemeier2026merlin} consists of 25,494 publicly available\footnote{Merlin is publicly available at \url{https://stanfordaimi.azurewebsites.net/datasets/60b9c7ff-877b-48ce-96c3-0194c8205c40}} contrast-enhanced computed tomography (CECT) scans from 18,317 patients. Each CT exam is paired with the corresponding radiology report. The dataset includes abdominal and pelvic CT exams conducted between 2012 and 2018 at the Stanford Hospital Emergency Departments. 

\subsubsection{CancerVerse} 

CancerVerse \cite{li2026cancerverse} is a publicly available longitudinal and multimodal CT dataset for multicancer screening\footnote{CancerVerse is publicly available at \url{https://huggingface.co/datasets/BodyMaps/CancerVerse}}. The current release contains 24,422 CT scans from 14,599 patients, acquired at hospitals in Switzerland and Turkey between 2010 and 2025. It covers the chest, abdomen, and pelvis and pairs CT imaging with radiology reports, complemented by pathology results and clinical variables. The dataset provides expert voxel-wise tumor annotations for 13 cancer types and includes a verified healthy cohort for evaluating false-positive detections. Patient histories span up to 14.4 years, with up to 26 examination dates per patient, supporting longitudinal analysis of imaging findings and their associated reports.

\subsubsection{Proprietary set}

We utilized a nationwide cohort from Poland comprising 17,415 multi-phase CT scans across 5,805 examinations from 4,004 patients. The dataset was collected from over 300 hospitals across the country, with each scan paired with its corresponding radiology report. Ethical approval for this study was granted by the Bioethics Committee at the Faculty of Medicine, Collegium Medicum, University of Warmia and Mazury in Olsztyn (Approval No. 9/2023, March 30, 2023).

\subsection{Fine-tuning}

\subsubsection{AMOS-CT}

The AMOS-CT dataset\footnote{AMOS-CT dataset is publicly available at \url{https://zenodo.org/records/7262581}} \cite{ji2022amos} consists of a total of 500 CT scans collected from diverse clinical centers, ensuring significant variability in scanner types and patient demographics. However, the ground-truth labels are publicly accessible only for the training and validation cohorts, comprising 200 and 100 scans, respectively. Each volume provides high-quality, voxel-level manual annotations for 15 distinct abdominal organs: spleen, right kidney, left kidney, gallbladder, esophagus, liver, stomach, aorta, inferior vena cava, pancreas, right adrenal gland, left adrenal gland, duodenum, bladder, and prostate/uterus. For our experimental setup, we utilize the 200 training scans for model optimization and the 100 validation scans as our test set. The dataset is particularly challenging due to its large intra-class variation; for instance, the number of axial slices per scan ranges from 68 to 353, and the voxel spacing varies significantly from $[0.45,1.07]\times[0.45,1.07]\times[1.25,5.0]$ $\text{mm}^3$. The mean spacing is $0.70 \times 0.70 \times 4.20$ $\text{mm}^3$. Beyond its scale, AMOS-CT is recognized for including both healthy and pathological cases, requiring models to be robust against anatomical distortions and varied fields of view typical in real-world clinical environments.

\subsubsection{AMOS-MRI}

AMOS-MRI is the MRI subset of the AMOS dataset \cite{ji2022amos}, comprising 100 3D abdominal MRI scans with voxel-wise annotations of 15 anatomical structures\footnote{AMOS-MRI is publicly available at \url{https://zenodo.org/records/7262581}}. The annotations cover major abdominal organs and vessels, including the liver, spleen, kidneys, pancreas, stomach, aorta, and inferior vena cava. Although the training annotations include 15 foreground anatomical classes, the AMOS-MRI test-set annotations do not include the urinary bladder or the prostate/uterus class. We therefore evaluate segmentation performance on the remaining 13 foreground classes. The benchmark includes 40 scans for training, 20 for validation, and 40 for testing, with the test set further divided into 11 in-distribution and 29 out-of-distribution scans. Its heterogeneous acquisition settings support the evaluation of segmentation accuracy and generalization across scanner domains.

\subsubsection{PANORAMA} The PANORAMA dataset \cite{alves2024panorama} comprises 2,238 multi-center contrast-enhanced computed tomography (CECT) scans acquired in a single portal-venous phase. It includes 1,964 newly acquired scans from five European centers, as well as publicly available data from two medical datasets in the United States, namely NIH \cite{roth2015deeporgan} and MSD Pancreas \cite{antonelli2022medical}. The dataset consists of 676 PDAC (pancreatic ductal adenocarcinoma) cases and 1,562 non-PDAC cases. While the dataset contains six voxel-wise labels (PDAC lesion, veins, arteries, pancreatic parenchyma, pancreatic duct, and common bile duct), we use only the PDAC lesion and merge the pancreatic parenchyma and pancreatic duct classes into a single \textit{pancreas} label. Each CT image consists of 37--1,572 slices, with a resolution ranging from 512$\times$512 to 1024$\times$1024 pixels. The voxel spacing ranges over $[0.31,1.03]\times[0.31,1.03]\times[0.45,5.0]$ mm$^3$ with a mean of 0.75 $\times$ 0.75 $\times$ 2.0 mm$^3$.

\subsubsection{GEPAR3D}

We utilize the GEPAR3D dataset\footnote{GEPAR3D dataset is publicly available at \url{https://zenodo.org/records/15739014}} \cite{szczepanski2025gepar3d}, which is derived from 98 publicly available cone-beam computed tomography (CBCT) scans originally released by \citep{cui2022fully}. While the initial iteration of these data provided only binary masks (distinguishing tooth structure from background), it presented significant limitations for tasks requiring individual tooth identification or pathological analysis. To address this, the GEPAR3D authors performed an extensive manual reannotation process~\citep{szczepanski2025gepar3d}. Experienced clinicians labeled the dataset into 32 distinct anatomical classes following the Universal Numbering System, effectively transforming it into a benchmark for fine-grained multi-class dental segmentation. The dataset is characterized by its high anatomical complexity, featuring diverse dental conditions including restorations and implants that introduce metallic artifacts, varying stages of tooth decay and bone loss, and impacted teeth and irregular dental alignments. The GEPAR3D test set consists of 46 publicly available CBCT scans from four external centers as part of the GEPAR3D dataset \citep{szczepanski2025gepar3d}. This cohort includes two subsets sourced from public repositories: 15 scans from \citep{cui2022fully}, which also constitute a subset of the ToothFairy2 dataset \citep{bolelli2024segmenting, bolelli2025segmenting, lumetti2024enhancing}, and two newly released centers provided to the research community. The remaining scans originate from two independent clinical centers, ensuring a diverse evaluation of model generalization across different scanning protocols and anatomical variations.

\subsubsection{MRISegmenter} We utilize the MRISegmenter abdominal MRI cohort\footnote{The MRISegmenter segmentation tool, model weights, and accompanying dataset are publicly available at \url{https://github.com/rsummers11/MRISegmenter} under the data-use terms described in the original publication.}~\cite{zhuang2025mrisegmenter}, introduced as part of a retrospective study of patients acquired at the U.S.\ National Institutes of Health (NIH) Clinical Center. The resource comprises 195 patients with 780 3D T1-weighted abdominal MRI series in total, corresponding to four acquisitions per patient: axial pre-contrast T1 and contrast-enhanced T1 in arterial, portal venous, and delayed phases. Each series is accompanied by dense, voxel-level reference segmentations of abdominal organs and anatomic structures, providing a fine-grained multi-class benchmark for volumetric abdominal MRI segmentation beyond coarse organ-only labels. The cohort is clinically heterogeneous across adult patients (training split: 135 subjects; internal test split: 60 subjects) and spans the full multi-phase dynamic-liver protocol, which introduces realistic variability in parenchymal enhancement, vascular conspicuity, and partial-volume effects across phases. 

\subsubsection{MVSeg}

We utilize the MVSeg2023 benchmark\footnote{The MVSeg2023 challenge data are catalogued on Hugging Face at \url{https://huggingface.co/datasets/pcarnahan/MVSeg2023}; the official challenge portal is \url{https://www.synapse.org/MVSeg2023}. Access may require accepting the host platform's data-use conditions.}~\cite{carnahan2021deepmitral}, released in conjunction with the MICCAI~2023 workshop on \emph{Mitral Valve Segmentation from 3D Transesophageal Echocardiography}. The goal is automatic voxel-wise delineation of the mitral valve leaflets from \emph{single-frame} three-dimensional transesophageal echocardiography (3D TEE) volumes acquired at end-diastole in a mid-esophageal, en-face mitral view. Volumes were collected on a Philips Epiq cardiac ultrasound system according to clinical standard of care at King's College Hospital (London, UK), converted from vendor DICOM to Cartesian coordinates with Philips QLab, and distributed as NIfTI files with spatial metadata limited to image size and spacing. Each case consists of one patient-specific 3D TEE volume with dense annotations of the anterior and posterior mitral leaflets (three labels: background, posterior leaflet, anterior leaflet), yielding a fine-grained yet low-cardinality segmentation task that is representative of intra-operative imaging noise, acoustic shadowing, and thin-structure ambiguity. The public release is partitioned into 105 training, 30 validation, and 40 test volumes for standardized comparison.

\subsubsection{AbdomenAtlas 3.0}

We utilize AbdomenAtlas~3.0\footnote{AbdomenAtlas~3.0 and the RadGPT
release are publicly available at \url{https://github.com/MrGiovanni/RadGPT}.}~\cite{bassi2025radgpt},
a large-scale abdominal CT resource as part of the RadGPT pipeline. The cohort aggregates
9{,}262 isotropic three-dimensional CT volumes in NIfTI format
(2{,}789{,}975 axial slices) drawn from 17 previously
released public abdominal CT datasets that originally shipped without
paired radiology reports. Each study is distributed as a \emph{triplet}
of (i)~the CT volume, (ii)~dense voxel-wise annotations of liver, kidney,
and pancreatic tumors together with ancillary structures used for staging
and localization---including Couinaud liver sub-segments (1--8),
pancreatic sub-regions (head, body, tail), and peripancreatic vasculature
relevant to pancreatic ductal
adenocarcinom (PDAC) \mbox{T1--T4}
staging, and (iii)~expert-reviewed radiology text. In total
3{,}955 scans contain at least one annotated tumor; the
curators report a $4.2\times$ expansion of publicly available
tumor masks relative to the union of the source releases, with
3{,}011 additional tumors identified after revision of automatic segmentations by board-certified radiologists. The textual modality
comprises structured, narrative, and (where applicable) enhanced reports
generated and verified under radiologist supervision with RadGPT,
documenting per-lesion size, attenuation (HU), volume, anatomic location
within sub-segments, and resectability-related attributes, for a total of
1{,}843{,}262 report tokens. AbdomenAtlas~3.0 is the first public
distribution in this line that couples every scan with both high-resolution tumor (and staging-relevant) segmentations and
paired, tumor-focused radiology text, making it a natural benchmark for 3D abdominal vision--language pre-training and
segmentation-assisted report generation.

\subsubsection{RATIC}

We utilize the RSNA Abdominal Traumatic Injury CT (RATIC)
cohort\footnote{RATIC is distributed for non-commercial research through
the RSNA AI Challenge platform at \url{https://imaging.rsna.org/dataset/5}
and the companion Kaggle release at
\url{https://www.kaggle.com/c/rsna-2023-abdominal-trauma-detection}.}~\cite{rudie2024rsna},
curated by the Radiological Society of North America (RSNA) in collaboration
with the American Society of Emergency Radiology and the Society of
Abdominal Radiology. The resource comprises 4{,}274 consecutive
adult ($\geq$18~years) abdominal--pelvic CT studies, totaling
6{,}481 image series contributed from 23 institutions
across 14 countries on six continents---at the time of its
release, the largest publicly available multi-center collection
explicitly labeled for blunt abdominal traumatic injury on CT. Each
study ships in DICOM format together with comma-separated tabular
metadata: patient demographics, series-level acquisition descriptors
(imaging phase and anatomic coverage), and study-level injury labels for
the liver, kidneys, spleen, bowel/mesentery, and active contrast
extravasation. Solid-organ injuries are graded on the American
Association for the Surgery of Trauma (AAST) organ-injury scale,
with low-grade (grades~I--III) versus high-grade (grades~IV--V)
injuries dichotomized to reduce inter-rater noise; bowel/mesenteric
injury and extravasation receive complementary image-level annotations
because of their heterogeneous appearance across phases. In
addition, voxel-wise segmentations of the liver, spleen, left and right
kidneys, and a composite gastrointestinal label (esophagus through
colon) are publicly released in NIfTI format for a 206-series
pilot subset of the training portion, enabling joint segmentation and
injury-classification benchmarks. The dataset was used for the RSNA
2023 Abdominal Trauma Detection challenge and is intended to stress-test
algorithms under realistic protocol heterogeneity (e.g.\ mono- versus
multi-phase acquisition, variable slice thickness), reflecting the
clinical variability encountered in emergency CT workflows.

\clearpage

\section{Experimental setup}
\label{sec:experimentalsetup}

We provide a detailed description of our experimental setup, covering data preprocessing, model implementation specifics, and evaluation metrics.

\subsection{Hardware and software setup}
\label{appendix:hardware_and_software_setup}

All experiments are conducted on a high-performance computing cluster utilizing 4 $\times$ NVIDIA GH200 GPUs, each with 96 GB of memory. All networks are implemented using Python 3.11, PyTorch 2.7 \cite{paszke2019pytorch}, and the medical imaging framework MONAI 1.3.0. Training stability and efficiency are ensured by employing the PyTorch Distributed Data-Parallel (DDP) framework. For DDP training, we use PyTorch Lightning Fabric.

\subsection{Pre-training}

We utilize TotalSegmentator \cite{wasserthal2023totalsegmentator} to segment anatomical structures from the 3D CT scans. For downstream analysis, we focus on the primary anatomical structures within the abdomen and pelvis, resulting in $C-1=14$ foreground classes and $C=15$ total classes (see Table \ref{tab:selectedclasses} for the full list).

We resample all CT scans to 1.5 $\times$ 1.5 $\times$ 3.0 mm$^{3}$ anisotropic resolution.  All CT scans are normalized to a range of $[0, 1]$. We first apply a clipping operation to the raw Hounsfield Units (HU) to focus on clinically relevant intensities and remove artifacts. The normalized intensity $I_{\text{norm}}$ is calculated as:

\begin{equation}
I_{\text{norm}} = \frac{\text{clamp}(I, HU_{\min}, HU_{\max}) - HU_{\min}}{HU_{\max} - HU_{\min}}
\label{eq:normalization}
\end{equation}

where the clipping bounds are set to $[-1000, 1000]$.

\subsection{Pre-training objective and routing details}
\label{sec:objective_details}
The implementation described here uses the anatomical routing representation defined in the main text. Each stage predicts raw crop composition. During training, label-derived or detached predicted composition is selected independently for each sample; for zero-based epoch $e$, the probability of predicted routing is $p_e=0.5+0.5\min(e/3,1)$. At inference, routing uses predicted composition only. No segmentation mask or report is required.

\noindent
\textbf{Routing supervision.} The routing objective sums the composition loss and router penalties over stages:
\begin{equation}
\mathcal L_{\rm MoE}=\sum_{s=1}^{4}\left[
\mathcal L_{\rm comp}^{(s)}+
0.5\,\mathbb E_b\operatorname{KL}\!\left(
\mathbf W_{s,b}\,\Vert\,
\operatorname{sg}\!\left[\mathcal T(\mathbf h_b)\widetilde{\mathbf A}^{(s)}\right]\right)
+\mathcal R_s\right].
\end{equation}
Here $\operatorname{sg}$ stops gradients through the alignment target. The usage regularizer $\mathcal R_s$ combines rolling anatomical-prior matching, a normalized per-sample entropy-band penalty with bounds $[0.25,0.8]$, and foreground affinity coverage. Its outer weight is $0.1$ and the relative coverage weight is $0.1$. These terms encourage controlled mixture usage, without guaranteeing equal weights or organ-exclusive experts. The positive prior strength is parameterized as $\beta_s=\operatorname{softplus}(b_s)$, and the log-prior uses $\epsilon=10^{-6}$.

\noindent
\textbf{Organ--text supervision.} The symmetric within-organ InfoNCE objective uses structured report findings. Exact duplicates after case folding and whitespace normalization form multiple positives. Negatives are different-case, different-text findings with opposite known normal/abnormal report labels; other pairs are masked. Each direction uses current-batch anchors and current plus queued candidates. Detached per-organ FIFO queues of length 64 are cleared at the start of each training epoch and are not used for validation. An anchor contributes only when it has both a positive and an eligible negative. Missing findings are excluded. Organ--text pairs require nonempty findings and visible organ support of at least $0.001$ of the crop; no additional whole-organ coverage filter is applied in this recipe. Report labels are weak supervision and do not establish clinical incompatibility between every negative pair.

\noindent
\textbf{Cross-stage consistency.} Organ-indexed embeddings are encouraged to agree across the four scales:
\begin{equation}
\mathcal L_{\rm stage}=\frac16\sum_{1\leq s<s'\leq4}
\mathbb E\!\left[1-\cos(\hat{\mathbf E}_s,\hat{\mathbf E}_{s'})\right].
\end{equation}

\noindent
\textbf{Query grounding.} For each valid organ--stage target, its binary pseudo-mask is average-pooled to the attention grid and normalized to a distribution $\mathbf m_{s,b,c}$. With normalized query attention $\mathbf a_{s,b,c}$, we use
\begin{equation}
\mathcal L_{\rm ground}=\frac1{|\mathcal G|}
\sum_{(s,b,c)\in\mathcal G}
\operatorname{KL}(\mathbf m_{s,b,c}\Vert\mathbf a_{s,b,c}),
\end{equation}
where $\mathcal G$ contains eligible targets with nonzero mask mass. The term is zero when no target is valid. Masks supervise attention during training but are not inputs to OQMSP's forward computation. This encourages organ-level attention rather than establishing lesion localization.

\subsection{Downstream tasks}

In this section, we provide a detailed description of our experimental setup for downstream tasks, including segmentation, classification, and detection.



\subsubsection{Segmentation}

Each segmentation model architecture is trained with an input patch size of $P \times P \times P$, where $P = 128$, to accommodate GPU memory constraints while preserving sufficient spatial context. We employ the sum of Dice and cross-entropy losses as our final loss function.

We utilize the AdamW optimizer with an initial learning rate $\eta_0$ set to 1e-4. The learning rate schedule follows cosine annealing with a minimum learning rate of $\eta_{\text{min}}$, which is set to 1e-6 and a warm-up period of $W$ epochs. We train the model for a total of $T$ epochs with a batch size of $B=2$ per GPU, totaling an effective batch size of $B_{\text{eff}} = 8$. Weight decay is applied with a factor of $\lambda=1e-5$.

\noindent
\textbf{AMOS-CT.} All CT scans are resampled to 1.5 $\times$ 1.5 $\times$ 1.5 mm$^{3}$ isotropic resolution. Then, we set the clipping window to $[-1000, 1000]$ and linearly scale the intensities to $[0,1]$. We train each model for 800 epochs. 

\noindent
\textbf{PANORAMA.} All CT scans are resampled to 0.8 $\times$ 0.8 $\times$ 3.0 mm$^{3}$ anisotropic resolution. Then, we set the clipping window to $[-175,250]$ and linearly scale the intensities to $[0,1]$. We train each model for 800 epochs.

\noindent
\textbf{GEPAR3D.} All CT scans are resampled to 0.4 $\times$ 0.4 $\times$ 0.4 mm$^{3}$ isotropic resolution. Then, we set the clipping window to $[0, 5000]$ and linearly scale the intensities to $[0,1]$. We train each model for 1,000 epochs.

\noindent
\textbf{AMOS-MRI.} All MRI scans are resampled to 1.5 $\times$ 1.5 $\times$ 1.5 mm$^{3}$ isotropic resolution. Then, we set the clipping window to $[0, 1000]$ and linearly scale the intensities to $[0,1]$. We train each model for 1,000 epochs.

\noindent
\textbf{MRISegmenter.} All MRI scans are resampled to 1.5 $\times$ 1.5 $\times$ 3.0 mm$^{3}$ anisotropic resolution. Then, we set the clipping window to $[0, 1000]$ and linearly scale the intensities to $[0,1]$. We train each model for 1,000 epochs.

\noindent
\textbf{MVSeg.} All US scans are resampled to 0.5 $\times$ 0.5 $\times$ 0.5 mm$^{3}$ isotropic resolution. Then, we set the clipping window to $[0, 255]$ and linearly scale the intensities to $[0,1]$. We train each model for 1,000 epochs.

\subsubsection{Classification}

\begin{figure}[h]
    \centering
    \includegraphics[width=0.8\linewidth]{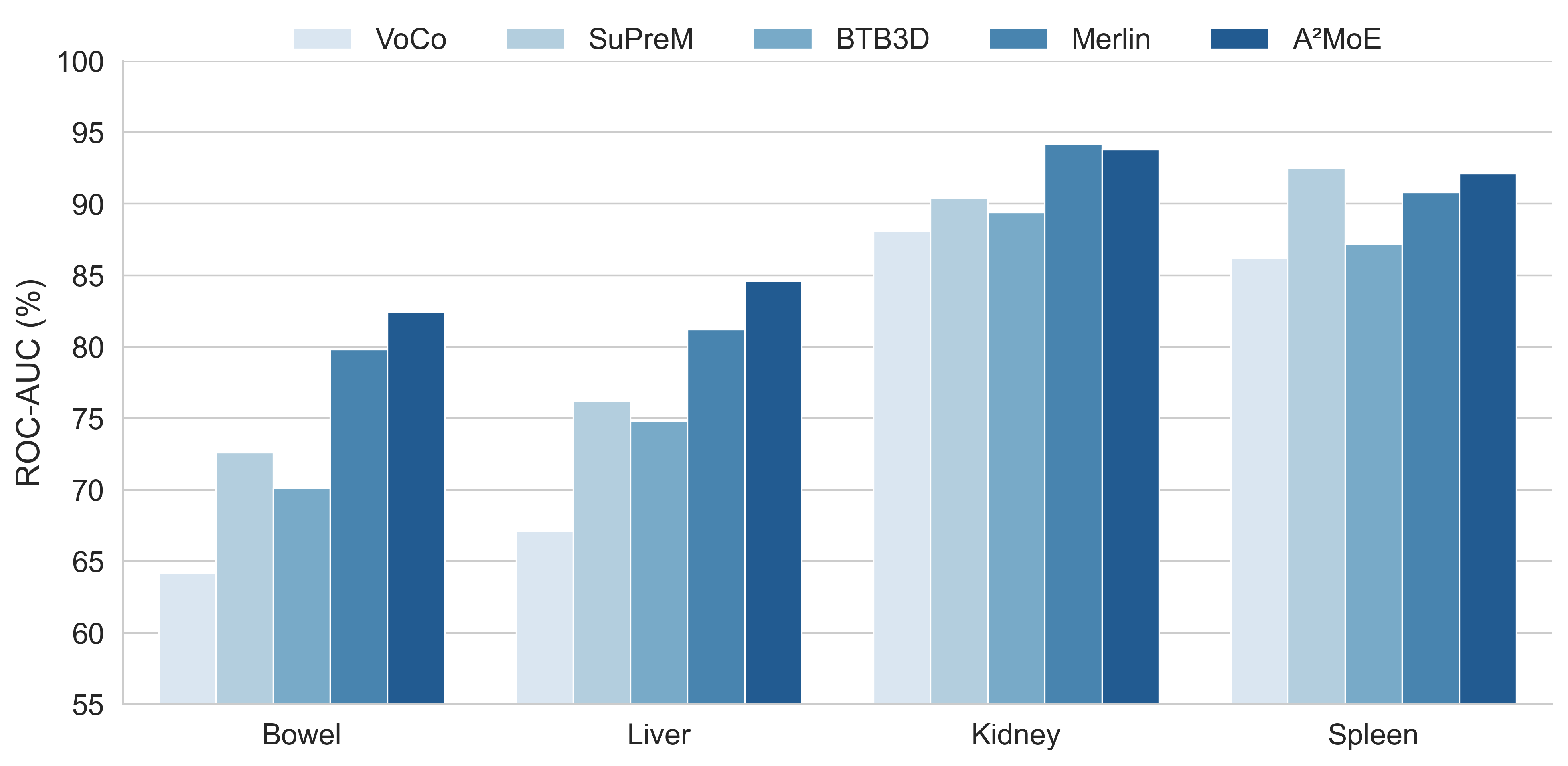}
    \caption{\textbf{ROC-AUC scores for the classification tasks} across bowel, liver, kidney, and spleen abdominal injuries on RATIC \cite{rudie2024rsna}.}
        \label{fig:classification_mean}
\end{figure}

We evaluate abdominal trauma \emph{classification} on RATIC~\cite{rudie2024rsna}. Each CT scan is resampled to $1.5 \times 1.5 \times 3.0$~mm$^{3}$ anisotropic spacing. For every compared model we keep only the pre-trained 3D \emph{encoder} (the segmentation decoder is discarded), attach a case-level head, and fine-tune the encoder and head on the RATIC training split. Volumes are processed with a sliding window of $P \times P \times P$ patches ($P{=}128$): we encode all valid patches, apply global average pooling over patch embeddings to obtain a fixed-length volume descriptor, and pass it through a lightweight MLP that outputs one logit per evaluated binary or graded task (bowel, liver, kidney, and spleen; see Figure~\ref{fig:classification_mean} and Table~\ref{tab:ratic_binary_compact} and Table~\ref{tab:ratic_per_grade}). Training minimizes the sum of sigmoid cross-entropy losses over the active heads on each sample.

\subsubsection{Detection}

We report pancreas, kidney, and liver tumor \emph{detection} on AbdomenAtlas~3.0~\cite{bassi2025radgpt} (dataset details in Appendix~\ref{sec:appendixdatasets}) using \emph{only} the corresponding segmentation-trained models: we do not train a separate detector. Study-level predictions are obtained from the \emph{segmentation} decoder by aggregating voxel-wise tumor-class \emph{logits} (e.g., thresholding the maximum tumor-class probability or the volume of connected high-confidence voxels). Sensitivity is stratified by lesion long-axis diameter ($\le$2~cm vs.\ $>$2~cm); specificity is computed at the study level. Results are shown in Figure~\ref{fig:detection_plot} and Table~\ref{tab:detection_results}; formal definitions appear in Appendix~\ref{sec:appendixevaluationmetrics_classification}.

\subsection{Evaluation metrics}
\label{sec:appendixevaluationmetrics}

\subsubsection{Segmentation}
\label{sec:appendixevaluationmetrics_segmentation}

For segmentation tasks, we use the Dice similarity coefficient (DSC) and the 95th-percentile Hausdorff distance (HD95) to quantitatively assess model performance. The DSC measures the spatial overlap between the predicted segmentation \(P\) and the ground truth \(G\), and is defined as:

\begin{equation}
    \operatorname{DSC}(P,G) = \frac{2|P \cap G|}{|P| + |G|},
\end{equation}
where \(|P|\) and \(|G|\) denote the number of voxels in the predicted and ground-truth masks, respectively. 

To evaluate boundary accuracy, we report HD95, a robust variant of the Hausdorff distance that reduces sensitivity to outliers. Let $\partial P$ and $\partial G$ denote the contours of the predicted and ground-truth masks. The following definition takes the maximum of the two directed 95th-percentile contour distances:

\begin{equation}
    \operatorname{HD95}(P, G) = \max \left\{ 
\operatorname{Percentile}_{95}\big(d(p, G)\big),\ 
\operatorname{Percentile}_{95}\!\left(\{d(g,\partial P):g\in\partial G\}\right)
\right\},
\end{equation}
where $d(p,\partial G)=\min_{g\in\partial G}\|p-g\|_2$ and $d(g,\partial P)=\min_{p\in\partial P}\|g-p\|_2$, using physical coordinates in millimeters. Higher DSC and lower HD95 indicate better segmentation performance.

\subsubsection{Classification and detection}
\label{sec:appendixevaluationmetrics_classification}

For classification and detection tasks, we evaluate model performance using sensitivity, specificity, F1-score, and the area under the receiver operating characteristic curve (ROC-AUC). Let \(TP\), \(TN\), \(FP\), and \(FN\) denote the number of true-positive, true-negative, false-positive, and false-negative predictions, respectively. Sensitivity (also called recall or the true-positive rate) and specificity (the true-negative rate) are defined as:

\begin{equation}
    \text{Sensitivity} = \frac{TP}{TP + FN},
\end{equation}

\begin{equation}
    \text{Specificity} = \frac{TN}{TN + FP}.
\end{equation}

To account for class imbalance, we report the F1-score, which is the harmonic mean of precision and sensitivity. Given precision \( = \frac{TP}{TP + FP} \), the F1-score is defined as:

\begin{equation}
    \text{F1-score} = \frac{2TP}{2TP + FP + FN}.
\end{equation}

The ROC-AUC value ranges from 0 to 1, where an AUC of 0.5 indicates performance no better than random chance, and an AUC of 1.0 represents a perfect classifier. Higher sensitivity, specificity, F1-score, and ROC-AUC correspond to superior classification results.

\clearpage

\section{Ablation studies}
\label{sec:appendixablationstudies}

\subsection{Effect of the key components}

\begin{table}[h]
\centering
\caption{Ablation study on the effect of key components on the AMOS-CT \cite{ji2022amos} validation set. B, MoE, and T denote the baseline, A$^{2}$MoE, and OQMSP modules, respectively. Evaluation metrics include mDSC ($\uparrow$, \%) and mHD95 ($\downarrow$, mm). Organ abbreviations: Spl (spleen), RK/LK (right/left kidney), GB (gallbladder), Eso (esophagus), Liv (liver), Sto (stomach), Aor (aorta), IVC (inferior vena cava), Pan (pancreas), RAG/LAG (right/left adrenal gland), Duo (duodenum), Bla (bladder), and P/U (prostate/uterus). The best results are \textbf{bolded}.}
\label{tab:organ_ablation_shortcuts}
\scriptsize
\setlength{\tabcolsep}{3pt}
\resizebox{\textwidth}{!}{%
\begin{tabular}{l|ccc|ccccccccccccccc|cc}
\toprule
& \multicolumn{3}{c|}{\textbf{Config}} & \multicolumn{15}{c|}{\textbf{Organ Classes (DSC \%)}} & \multicolumn{2}{c}{\textbf{Overall}} \\ \cmidrule(l){2-4} \cmidrule(l){5-19} \cmidrule(l){20-21}
\textbf{\#} & \textbf{B} & \textbf{MoE} & \textbf{T} &
\rotatebox{90}{Spl} & \rotatebox{90}{RK} & \rotatebox{90}{LK} & \rotatebox{90}{GB} & \rotatebox{90}{Eso} & \rotatebox{90}{Liv} & \rotatebox{90}{Sto} & \rotatebox{90}{Aor} & \rotatebox{90}{IVC} & \rotatebox{90}{Pan} & \rotatebox{90}{RAG} & \rotatebox{90}{LAG} & \rotatebox{90}{Duo} & \rotatebox{90}{Bla} & \rotatebox{90}{P/U} &
\textbf{mDSC} & \textbf{mHD95} \\ \midrule
1 & \cmark  & \xmark & \xmark & 95.0 & 95.5 & 95.4 & 75.5 & 80.7 & 96.2 & 88.2 & 93.5 & 88.9 & 82.7 & 75.1 & 73.6 & 75.6 & 86.0 & 80.1 & 85.5 & 17.3 \\
2 & \cmark & \cmark & \xmark & 95.8 & 95.9 & 94.9 & 82.7 & 83.6 & 96.8 & 91.5 & 94.1 & 90.0 & 85.6 & 75.1 & 75.5 & 79.7 & 89.3 & 81.6 & 87.5 & 7.6 \\
\rowcolor{blue!10} 3 & \cmark & \cmark & \cmark & \textbf{96.7} & \textbf{96.1} & \textbf{95.6} & \textbf{84.8} & \textbf{84.9} & \textbf{97.7} & \textbf{92.2} & \textbf{94.3} & \textbf{90.6} & \textbf{87.3} & \textbf{77.0} & \textbf{77.6} & \textbf{82.9} & \textbf{89.5} & \textbf{83.6} & \textbf{88.7} & \textbf{6.2} \\
\bottomrule
\end{tabular}
}
\end{table}

\begin{table}[h]
\centering
\caption{Ablation study on the effect of key components on the PANORAMA \cite{alves2026artificial} test sets (NIH, MSD, and proprietary). B, MoE, and T denote the baseline, A$^{2}$MoE, and OQMSP modules, respectively. Evaluation metrics include mDSC ($\uparrow$, \%) and mHD95 ($\downarrow$, mm). The best results are \textbf{bolded}.}
\label{tab:panorama_experts_comparison}
\scriptsize
\setlength{\tabcolsep}{3pt}
\begin{tabular}{l|ccc|c|cc|cc|cc}
\toprule
& \multicolumn{3}{c|}{\textbf{Config}} & \textbf{NIH} & \multicolumn{2}{c|}{\textbf{MSD Pancreas}} & \multicolumn{2}{c|}{\textbf{Proprietary}} & \multicolumn{2}{c}{\textbf{Overall}} \\ \cmidrule(l){2-4} \cmidrule(l){5-5} \cmidrule(l){6-7} \cmidrule(l){8-9} \cmidrule(l){10-11}
\textbf{\#} & \textbf{B} & \textbf{MoE} & \textbf{T} & \textbf{Pancreas} & \textbf{PDAC} & \textbf{Pancreas} & \textbf{PDAC} & \textbf{Pancreas} & \textbf{mDSC} & \textbf{mHD95} \\ \midrule
1 & \cmark & \xmark & \xmark & 92.7 & 47.9 & 88.1 & 54.0 & 81.9 & 76.0 & 8.6 \\
2 & \cmark & \cmark & \xmark & 92.0 & 53.6 & 87.7 & 65.8 & 81.4 & 78.2 & 7.1 \\
\rowcolor{blue!10} 3 & \cmark & \cmark & \cmark & \textbf{93.1} & \textbf{55.2} & \textbf{89.1} & \textbf{67.2} & \textbf{82.2} & \textbf{79.1} & \textbf{5.8} \\
\bottomrule
\end{tabular}
\end{table}

\begin{table}[h]
\centering
\caption{Ablation study on the effect of key components on the AMOS-MRI \cite{ji2022amos} validation set. B, MoE, and T denote the baseline, A$^{2}$MoE, and OQMSP modules, respectively. Evaluation metrics include mDSC ($\uparrow$, \%) and mHD95 ($\downarrow$, mm). Organ abbreviations: Spl (spleen), RK/LK (right/left kidney), GB (gallbladder), Eso (esophagus), Liv (liver), Sto (stomach), Aor (aorta), IVC (inferior vena cava), Pan (pancreas), RAG/LAG (right/left adrenal gland), and Duo (duodenum). Evaluation covers 13 foreground classes; urinary bladder and
prostate/uterus are excluded because their annotations are unavailable
in the test set. The best results are \textbf{bolded}.}
\label{tab:organ_ablation_shortcuts_modified}
\scriptsize
\setlength{\tabcolsep}{3pt}
\resizebox{\textwidth}{!}{%
\begin{tabular}{l|ccc|cccccccccccccc|cc}
\toprule
& \multicolumn{3}{c|}{\textbf{Config}} & \multicolumn{13}{c|}{\textbf{Organ Classes (DSC \%)}} & \multicolumn{2}{c}{\textbf{Overall}} \\ \cmidrule(l){2-4} \cmidrule(l){5-17} \cmidrule(l){18-19}
\textbf{\#} & \textbf{B} & \textbf{MoE} & \textbf{T} &
\rotatebox{90}{Spl} & \rotatebox{90}{RK} & \rotatebox{90}{LK} & \rotatebox{90}{GB} & \rotatebox{90}{Eso} & \rotatebox{90}{Liv} & \rotatebox{90}{Sto} & \rotatebox{90}{Aor} & \rotatebox{90}{IVC} & \rotatebox{90}{Pan} & \rotatebox{90}{RAG} & \rotatebox{90}{LAG} & \rotatebox{90}{Duo} &
\textbf{mDSC} & \textbf{mHD95} \\ \midrule
1 & \cmark & \xmark & \xmark & 86.4 & 89.8 & 89.6 & 62.5 & 70.5 & 95.7 & 82.4 & 87.0 & 81.0 & 83.5 & 59.8 & 64.2 & 64.4 & 76.7 & 9.3 \\
2 & \cmark & \cmark & \xmark & 91.0 & 90.6 & 90.7 & 67.3 & 72.3 & 96.5 & 81.4 & 87.5 & 83.4 & 80.6 & 57.4 & 59.5 & 62.8 & 78.5 & 12.2 \\
\rowcolor{blue!10} 3 & \cmark & \cmark & \cmark & \textbf{91.5} & \textbf{90.9} & \textbf{91.2} & \textbf{78.9} & \textbf{75.0} & \textbf{97.4} & \textbf{86.8} & \textbf{87.6} & \textbf{84.4} & \textbf{86.5} & \textbf{64.5} & \textbf{67.0} & \textbf{68.1} & \textbf{82.3} & \textbf{8.9} \\
\bottomrule
\end{tabular}
}
\end{table}

\begin{table}[h]
\centering
\caption{Ablation study on the effect of key components on the GEPAR3D \cite{szczepanski2025gepar3d} test set. B, MoE, and T denote the baseline, A$^{2}$MoE, and OQMSP modules, respectively. Evaluation metrics include mDSC ($\uparrow$, \%) and mHD95 ($\downarrow$, mm). FDI notation is used for individual teeth across quadrants (Q1--Q4). The best results are \textbf{bolded}.}
\label{tab:dental_segmentation_empty}
\scriptsize
\setlength{\tabcolsep}{2.5pt}
\resizebox{\columnwidth}{!}{%
\begin{tabular}{l|ccc|cccccccc|cccccccc|cccccccc|cccccccc|cc}
\toprule
& \multicolumn{3}{c|}{\textbf{Config}} & \multicolumn{8}{c|}{\textbf{Upper Right (Q1)}} & \multicolumn{8}{c|}{\textbf{Upper Left (Q2)}} & \multicolumn{8}{c|}{\textbf{Lower Left (Q3)}} & \multicolumn{8}{c|}{\textbf{Lower Right (Q4)}} & \multicolumn{2}{c}{\textbf{Overall}} \\ \cmidrule(l){2-4} \cmidrule(l){5-12} \cmidrule(l){13-20} \cmidrule(l){21-28} \cmidrule(l){29-36} \cmidrule(l){37-38}
\textbf{\#} & \textbf{B} & \textbf{MoE} & \textbf{T} &
\rotatebox{90}{18} & \rotatebox{90}{17} & \rotatebox{90}{16} & \rotatebox{90}{15} & \rotatebox{90}{14} & \rotatebox{90}{13} & \rotatebox{90}{12} & \rotatebox{90}{11} &
\rotatebox{90}{21} & \rotatebox{90}{22} & \rotatebox{90}{23} & \rotatebox{90}{24} & \rotatebox{90}{25} & \rotatebox{90}{26} & \rotatebox{90}{27} & \rotatebox{90}{28} &
\rotatebox{90}{38} & \rotatebox{90}{37} & \rotatebox{90}{36} & \rotatebox{90}{35} & \rotatebox{90}{34} & \rotatebox{90}{33} & \rotatebox{90}{32} & \rotatebox{90}{31} &
\rotatebox{90}{41} & \rotatebox{90}{42} & \rotatebox{90}{43} & \rotatebox{90}{44} & \rotatebox{90}{45} & \rotatebox{90}{46} & \rotatebox{90}{47} & \rotatebox{90}{48} &
\textbf{mDSC} & \textbf{mHD95} \\ \midrule
1 & \cmark & \xmark & \xmark & 68.0 & 87.9 & 90.2 & 85.3 & 85.2 & 91.7 & \textbf{91.0} & \textbf{93.7} & 93.1 & 90.9 & 89.2 & 86.1 & 90.0 & 93.0 & 85.3 & 70.6 & 66.9 & 78.3 & 84.5 & 89.4 & 89.4 & 85.4 & 87.2 & \textbf{86.0} & 80.3 & \textbf{89.1} & \textbf{92.6} & 89.1 & 93.3 & 84.8 & 81.4 & 56.6 & 85.3 & 10.6 \\
2 & \cmark & \cmark & \xmark & 67.9 & 87.6 & \textbf{90.8} & 88.0 & 85.6 & 91.4 & 89.3 & 93.0 & \textbf{93.3} & \textbf{91.7} & \textbf{89.4} & 84.0 & \textbf{92.0} & 92.4 & \textbf{91.4} & 74.0 & \textbf{68.1} & \textbf{88.2} & \textbf{88.6} & \textbf{91.4} & 88.6 & 84.7 & 85.4 & 80.7 & 83.5 & 88.9 & 91.4 & 88.6 & 92.2 & 88.7 & 83.2 & 59.8 & 86.2 & 8.3 \\
\rowcolor{blue!10} 3 & \cmark & \cmark & \cmark & \textbf{70.4} & \textbf{92.2} & 89.1 & \textbf{88.3} & \textbf{91.0} & \textbf{93.5} & 90.4 & 93.6 & 91.5 & \textbf{91.7} & 88.7 & \textbf{87.4} & 91.3 & \textbf{94.2} & 90.7 & \textbf{76.7} & 64.3 & 87.1 & 87.5 & 89.8 & \textbf{91.1} & \textbf{86.5} & \textbf{89.5} & \textbf{86.0} & \textbf{84.5} & 88.5 & 87.6 & \textbf{91.2} & \textbf{93.8} & \textbf{90.0} & \textbf{88.3} & \textbf{64.3} & \textbf{87.7} & \textbf{8.1} \\
\bottomrule
\end{tabular}
}
\end{table}

\begin{table}[h]
\centering
\scriptsize
\caption{Ablation study on the effect of key components on the MRISegmenter \cite{zhuang2025mrisegmenter} test set. B, MoE, and T denote the baseline, A$^{2}$MoE, and OQMSP modules, respectively. We evaluate segmentation performance across all 62 anatomical classes. Evaluation metrics include mDSC ($\uparrow$, \%) and mHD95 ($\downarrow$, mm). The best results are \textbf{bolded}.}
\label{tab:mri_ablation_split_62_equal}
\scriptsize
\setlength{\tabcolsep}{1.5pt}

\resizebox{\columnwidth}{!}{%
\begin{tabular}{l|ccc|ccccccccccccccccccccccccccccccc}
\toprule
& \multicolumn{3}{c|}{\textbf{Config}} & \multicolumn{31}{c}{\textbf{Anatomical Classes (DSC \%) $\uparrow$}} \\ \cmidrule(l){2-4} \cmidrule(l){5-35}
\textbf{\#} & \textbf{B} & \textbf{MoE} & \textbf{T} &
\rotatebox{90}{1 Spleen} & \rotatebox{90}{2 Kid R} & \rotatebox{90}{3 Kid L} & \rotatebox{90}{4 Gallb} & \rotatebox{90}{5 Liver} & \rotatebox{90}{6 Esoph} & \rotatebox{90}{7 Stom} & \rotatebox{90}{8 Aorta} & \rotatebox{90}{9 IVC} & \rotatebox{90}{10 P\&S Vn} & \rotatebox{90}{11 Pancr} & \rotatebox{90}{12 Adr R} & \rotatebox{90}{13 Adr L} & \rotatebox{90}{14 Lung R} & \rotatebox{90}{15 Lung L} & \rotatebox{90}{16 Sm Bow} & \rotatebox{90}{17 Duod} & \rotatebox{90}{18 Colon} & \rotatebox{90}{19 Ilia Art L} & \rotatebox{90}{20 Ilia Art R} & \rotatebox{90}{21 Ilia Vn L} & \rotatebox{90}{22 Ilia Vn R} & \rotatebox{90}{23 GluMax L} & \rotatebox{90}{24 GluMax R} & \rotatebox{90}{25 GluMed L} & \rotatebox{90}{26 GluMed R} & \rotatebox{90}{27 Auto L} & \rotatebox{90}{28 Auto R} & \rotatebox{90}{29 Ilio L} & \rotatebox{90}{30 Ilio R} & \rotatebox{90}{31 Hip L} \\ \midrule
1 & \cmark & \xmark & \xmark  & 90.7 & 96.4 & 86.6 & 73.6 & 98.4 & 84.7 & 92.6 & 95.2 & 91.9 & 61.7 & 87.1 & 60.2 & 58.0 & 97.1 & 96.1 & 91.0 & 82.6 & 89.8 & 75.9 & 80.2 & 67.1 & 59.7 & 63.5 & 60.4 & 84.6 & 78.9 & 97.3 & 97.1 & 95.3 & 95.8 & 80.5 \\
2 & \cmark & \cmark & \xmark & 91.5 & 96.3 & 91.5 & 78.6 & 98.5 & 85.4 & 93.6 & \textbf{95.5} & 93.1 & \textbf{64.0} & 88.1 & 61.0 & 59.2 & 97.5 & 96.6 & 92.2 & 84.2 & 91.6 & 76.8 & 77.1 & 66.5 & 58.2 & 70.0 & 59.1 & 86.8 & 79.2 & \textbf{97.5} & \textbf{97.4} & 95.5 & 96.4 & 83.4 \\
\rowcolor{blue!10} 3 & \cmark & \cmark & \cmark & \textbf{92.8} & \textbf{96.7} & \textbf{93.0} & \textbf{83.1} & \textbf{98.6} & \textbf{85.6} & \textbf{94.4} & 95.4 & \textbf{93.2} & 63.7 & \textbf{89.3} & \textbf{61.3} & \textbf{60.1} & \textbf{97.6} & \textbf{96.7} & \textbf{93.9} & \textbf{84.7} & \textbf{93.7} & \textbf{81.6} & \textbf{83.9} & \textbf{78.4} & \textbf{78.1} & \textbf{79.4} & \textbf{78.5} & \textbf{90.7} & \textbf{87.8} & \textbf{97.5} & \textbf{97.4} & \textbf{95.8} & \textbf{96.5} & \textbf{86.6} \\
\bottomrule
\end{tabular}
}

\vspace{1.5em}

\resizebox{\columnwidth}{!}{%
\begin{tabular}{l|ccc|ccccccccccccccccccccccccccccccc|cc}
\toprule
& \multicolumn{3}{c|}{\textbf{Config}} & \multicolumn{31}{c|}{\textbf{Anatomical Classes (DSC \%)}} & \multicolumn{2}{c}{\textbf{Overall}} \\ \cmidrule(l){2-4} \cmidrule(l){5-35} \cmidrule(l){36-37}
\textbf{\#} & \textbf{B} & \textbf{MoE} & \textbf{T} &
\rotatebox{90}{32 Hip R} & \rotatebox{90}{33 Sacr} & \rotatebox{90}{34 Rib L4} & \rotatebox{90}{35 Rib L5} & \rotatebox{90}{36 Rib L6} & \rotatebox{90}{37 Rib L7} & \rotatebox{90}{38 Rib L8} & \rotatebox{90}{39 Rib L9} & \rotatebox{90}{40 Rib L10} & \rotatebox{90}{41 Rib L11} & \rotatebox{90}{42 Rib L12} & \rotatebox{90}{43 Rib R4} & \rotatebox{90}{44 Rib R5} & \rotatebox{90}{45 Rib R6} & \rotatebox{90}{46 Rib R7} & \rotatebox{90}{47 Rib R8} & \rotatebox{90}{48 Rib R9} & \rotatebox{90}{49 Rib R10} & \rotatebox{90}{50 Rib R11} & \rotatebox{90}{51 Rib R12} & \rotatebox{90}{52 L5} & \rotatebox{90}{53 L4} & \rotatebox{90}{54 L3} & \rotatebox{90}{55 L2} & \rotatebox{90}{56 L1} & \rotatebox{90}{57 T12} & \rotatebox{90}{58 T11} & \rotatebox{90}{59 T10} & \rotatebox{90}{60 T9} & \rotatebox{90}{61 T8} & \rotatebox{90}{62 T7} & \textbf{mDSC} & \textbf{mHD95} \\ \midrule
1 & \cmark & \xmark & \xmark & 69.6 & 49.7 & 15.9 & 38.8 & 48.3 & 53.8 & 57.1 & 60.5 & 69.0 & 70.0 & 66.7 & 18.0 & 36.0 & 50.0 & 60.6 & 64.6 & 67.0 & 69.4 & 68.1 & 64.7 & 66.4 & 85.6 & 87.0 & 87.1 & 87.3 & 88.6 & 88.2 & 84.9 & 78.7 & 69.4 & 48.9 & 74.5 & 6.8 \\
2 & \cmark & \cmark & \xmark & 67.5 & 52.1 & 18.0 & 44.1 & 53.7 & 60.2 & \textbf{64.2} & \textbf{69.5} & \textbf{73.1} & \textbf{74.0} & \textbf{69.7} & 19.4 & 47.1 & 56.7 & 65.2 & 67.0 & 68.1 & \textbf{72.0} & 70.6 & \textbf{67.5} & 76.1 & 87.7 & 88.4 & 88.1 & 87.7 & 88.8 & \textbf{89.2} & 87.7 & 80.5 & 70.2 & 46.1 & 77.6 & 5.9 \\
\rowcolor{blue!10} 3 & \cmark & \cmark & \cmark & \textbf{80.6} & \textbf{70.3} & \textbf{19.1} & \textbf{49.3} & \textbf{58.5} & \textbf{61.7} & 63.8 & 67.0 & 71.0 & \textbf{74.0} & 67.9 & \textbf{21.5} & \textbf{50.5} & \textbf{60.2} & \textbf{68.4} & \textbf{70.3} & \textbf{71.5} & 71.3 & \textbf{71.8} & 67.4 & \textbf{79.7} & \textbf{89.8} & \textbf{89.9} & \textbf{89.6} & \textbf{88.8} & \textbf{88.9} & 88.3 & \textbf{87.9} & \textbf{80.9} & \textbf{77.9} & \textbf{58.9} & \textbf{79.8} & \textbf{5.5} \\
\bottomrule
\end{tabular}
}
\end{table}

\begin{table}[h]
\centering
\caption{Ablation study on the effect of key components on the MVSeg \cite{carnahan2021deepmitral} validation set. B, MoE, and T denote the baseline, A$^{2}$MoE, and OQMSP modules, respectively. Evaluation metrics include mDSC ($\uparrow$, \%) and mHD95 ($\downarrow$, mm). Organ abbreviations: PL (posterior leaflet) and AL (anterior leaflet). The best results are \textbf{bolded}.}
\label{tab:organ_ablation_reset_config}
\scriptsize
\setlength{\tabcolsep}{8pt}
\begin{tabular}{l|ccc|cc|cc}
\toprule
& \multicolumn{3}{c|}{\textbf{Config}} & \multicolumn{2}{c|}{\textbf{Organ Classes (DSC \%)}} & \multicolumn{2}{c}{\textbf{Overall}} \\ \cmidrule(l){2-4} \cmidrule(l){5-6} \cmidrule(l){7-8}
\textbf{\#} & \textbf{B} & \textbf{MoE} & \textbf{T} & \rotatebox{90}{PL} & \rotatebox{90}{AL} & \textbf{mDSC} & \textbf{mHD95}\\ \midrule
1 & \cmark & \xmark & \xmark & 83.0 & 84.7 & 83.8 & 5.8 \\
2 & \cmark & \cmark & \xmark & 84.2 & 85.6 & 84.9 & 4.3\\
\rowcolor{blue!10} 3 & \cmark & \cmark & \cmark & \textbf{84.5} & \textbf{86.0} & \textbf{85.4} & \textbf{4.0} \\
\bottomrule
\end{tabular}
\end{table}

\clearpage

\subsection{Effect of the number of Experts per stage}

\begin{table}[h]
\centering
\caption{Ablation study on the effect of the number of experts per encoder stage on the AMOS-CT \cite{ji2022amos} validation set. Evaluation metrics include mDSC ($\uparrow$, \%) and mHD95 ($\downarrow$, mm). Organ abbreviations: Spl (spleen), RK/LK (right/left kidney), GB (gallbladder), Eso (esophagus), Liv (liver), Sto (stomach), Aor (aorta), IVC (inferior vena cava), Pan (pancreas), RAG/LAG (right/left adrenal gland), Duo (duodenum), Bla (bladder), and P/U (prostate in males or uterus in females). The best results are \textbf{bolded}.}
\label{tab:organ_ablation_shortcuts}
\scriptsize
\setlength{\tabcolsep}{3pt}
\resizebox{\textwidth}{!}{%
\begin{tabular}{l|c|ccccccccccccccc|cc}
\toprule
& & \multicolumn{15}{c|}{\textbf{Organ Classes (DSC \%)}} & \multicolumn{2}{c}{\textbf{Overall}} \\ \cmidrule(l){2-2} \cmidrule(l){3-17} \cmidrule(l){18-19}
\textbf{\#} & \textbf{Experts} &
\rotatebox{90}{Spl} & \rotatebox{90}{RK} & \rotatebox{90}{LK} & \rotatebox{90}{GB} & \rotatebox{90}{Eso} & \rotatebox{90}{Liv} & \rotatebox{90}{Sto} & \rotatebox{90}{Aor} & \rotatebox{90}{IVC} & \rotatebox{90}{Pan} & \rotatebox{90}{RAG} & \rotatebox{90}{LAG} & \rotatebox{90}{Duo} & \rotatebox{90}{Bla} & \rotatebox{90}{P/U} &
\textbf{mDSC} & \textbf{mHD95} \\ \midrule
1 & [4, 4, 4, 4] & 95.6 & 94.4 & \textbf{95.3} & \textbf{83.9} & 83.5 & \textbf{96.9} & 91.0 & 94.1 & 89.9 & \textbf{85.9} & 74.2 & 75.7 & 79.6 & 89.2 & \textbf{82.9} & \textbf{87.5} & 8.4 \\
2 & [8, 8, 8, 8] & \textbf{95.8} & 95.6 & 93.2 & 80.7 & \textbf{83.6} & 96.6 & \textbf{91.8} & \textbf{94.2} & \textbf{90.0} & 85.8 & \textbf{75.1} & \textbf{76.2} & \textbf{79.9} & 89.1 & 82.2 & 87.3 & 8.7 \\
3 & [4, 8, 16, 32] & \textbf{95.8} & \textbf{95.9} & 94.9 & 82.7 & \textbf{83.6} & 96.8 & 91.5 & 94.1 & \textbf{90.0} & 85.6 & \textbf{75.1} & 75.5 & 79.7 & \textbf{89.3} & 81.6 & \textbf{87.5} & \textbf{7.6} \\
\rowcolor{blue!10} 4 & [16, 16, 16, 16] & 95.7 & 95.4 & 92.9 & 80.5 & 83.1 & 95.3 & 91.3 & 94.1 & 89.8 & 84.7 & 74.9 & 76.0 & 79.7 & 88.3 & 81.9 & 86.9 & 9.2 \\
\bottomrule
\end{tabular}
}
\end{table}

\begin{table}[h]
\centering
\caption{Ablation study on the effect of the number of experts per encoder stage on the PANORAMA \cite{alves2026artificial} test sets (NIH, MSD, and proprietary). Evaluation metrics include mDSC ($\uparrow$, \%) and mHD95 ($\downarrow$, mm). The best results are \textbf{bolded}.}
\label{tab:panorama_experts_comparison}
\scriptsize
\setlength{\tabcolsep}{5pt}
\begin{tabular}{l|c|c|cc|cc|cc}
\toprule
& & \textbf{NIH} & \multicolumn{2}{c|}{\textbf{MSD Pancreas}} & \multicolumn{2}{c|}{\textbf{Proprietary}} & \multicolumn{2}{c}{\textbf{Overall}} \\ \cmidrule(lr){3-3} \cmidrule(lr){4-5} \cmidrule(lr){6-7} \cmidrule(lr){8-9}
\textbf{\#} & \textbf{Experts} & \textbf{Pancreas} & \textbf{PDAC} & \textbf{Pancreas} & \textbf{PDAC} & \textbf{Pancreas} & \textbf{mDSC} & \textbf{mHD95} \\ \midrule
1 & [4, 4, 4, 4] & \textbf{92.4} & 50.7 & \textbf{88.9} & 63.7 & \textbf{81.8} & 77.9 & 7.6 \\
2 & [8, 8, 8, 8] & 92.0 & 50.4 & 85.4 & 63.3 & 79.5 & 76.1 & 8.5 \\
3 & [4, 8, 16, 32] & 92.3 & 51.4 & 88.1 & 60.5 & 81.8 & 77.4 & 8.2 \\
\rowcolor{blue!10} 4 & [16, 16, 16, 16] & 92.0 & \textbf{53.6} & 87.7 & \textbf{65.8} & 81.4 & \textbf{78.2} & \textbf{7.1} \\
\bottomrule
\end{tabular}
\end{table}

\begin{table}[h]
\centering
\caption{Ablation study on the effect of the number of experts per encoder stage on the AMOS-MRI validation set. Evaluation covers 13 foreground classes; urinary bladder and
prostate/uterus are excluded because their annotations are unavailable
in the test set. Evaluation metrics include mDSC ($\uparrow$, \%) and mHD95 ($\downarrow$, mm). The best results are \textbf{bolded}.}
\label{tab:organ_ablation_shortcuts}
\scriptsize
\setlength{\tabcolsep}{4pt}
\begin{tabular}{l|c|ccccccccccccc|cc}
\toprule
& & \multicolumn{13}{c|}{\textbf{Organ Classes (DSC \%)}} & \multicolumn{2}{c}{\textbf{Overall}} \\ 
\cmidrule(l){3-15} \cmidrule(l){16-17}
\textbf{\#} & \textbf{Experts} &
\rotatebox{90}{Spl} & \rotatebox{90}{RK} & \rotatebox{90}{LK} & \rotatebox{90}{GB} & \rotatebox{90}{Eso} & \rotatebox{90}{Liv} & \rotatebox{90}{Sto} & \rotatebox{90}{Aor} & \rotatebox{90}{IVC} & \rotatebox{90}{Pan} & \rotatebox{90}{RAG} & \rotatebox{90}{LAG} & \rotatebox{90}{Duo} &
\textbf{mDSC} & \textbf{mHD95} \\ \midrule
1 & [4, 4, 4, 4]     & 90.5 & 90.3 & 90.5 & \textbf{68.4} & \textbf{72.3} & \textbf{96.7} & 80.9 & 87.0 & 83.1 & 80.3 & \textbf{58.5} & 55.8 & 60.2 & 78.0 & 13.4 \\
2 & [8, 8, 8, 8]     & 90.6 & \textbf{90.7} & 90.5 & 67.3 & 70.8 & 96.3 & \textbf{81.5} & 86.5 & 82.7 & 79.2 & \textbf{58.5} & 56.5 & 61.9 & 77.9 & 16.9 \\
3 & [4, 8, 16, 32]   & 90.4 & \textbf{90.7} & 90.5 & 63.3 & 70.9 & 96.4 & 79.6 & 87.2 & \textbf{83.5} & \textbf{81.0} & 57.7 & 53.6 & 61.6 & 77.4 & \textbf{12.2} \\
\rowcolor{blue!10} 4 & [16, 16, 16, 16] & \textbf{91.0} & 90.6 & \textbf{90.7} & 67.3 & \textbf{72.3} & 96.5 & 81.4 & \textbf{87.5} & 83.4 & 80.6 & 57.4 & \textbf{59.5} & \textbf{62.8} & \textbf{78.5} & \textbf{12.2} \\
\bottomrule
\end{tabular}
\end{table}

\begin{table}[h]
\centering
\caption{Ablation study on the effect of the number of experts per encoder stage on the GEPAR3D \cite{szczepanski2025gepar3d} test set. FDI notation is used for individual teeth across quadrants (Q1--Q4). Evaluation metrics include mDSC ($\uparrow$, \%) and mHD95 ($\downarrow$, mm). The best results are \textbf{bolded}.}
\label{tab:dental_segmentation_empty}
\scriptsize
\setlength{\tabcolsep}{2.5pt}
\resizebox{\columnwidth}{!}{%
\begin{tabular}{l|c|cccccccc|cccccccc|cccccccc|cccccccc|cc}
\toprule
& & \multicolumn{8}{c|}{\textbf{Upper Right (Q1)}} & \multicolumn{8}{c|}{\textbf{Upper Left (Q2)}} & \multicolumn{8}{c|}{\textbf{Lower Left (Q3)}} & \multicolumn{8}{c|}{\textbf{Lower Right (Q4)}} & \multicolumn{2}{c}{\textbf{Overall}} \\ \cmidrule(l){3-10} \cmidrule(l){11-18} \cmidrule(l){19-26} \cmidrule(l){27-34} \cmidrule(l){35-36}
\textbf{\#} & \textbf{Experts} &
\rotatebox{90}{18} & \rotatebox{90}{17} & \rotatebox{90}{16} & \rotatebox{90}{15} & \rotatebox{90}{14} & \rotatebox{90}{13} & \rotatebox{90}{12} & \rotatebox{90}{11} &
\rotatebox{90}{21} & \rotatebox{90}{22} & \rotatebox{90}{23} & \rotatebox{90}{24} & \rotatebox{90}{25} & \rotatebox{90}{26} & \rotatebox{90}{27} & \rotatebox{90}{28} &
\rotatebox{90}{38} & \rotatebox{90}{37} & \rotatebox{90}{36} & \rotatebox{90}{35} & \rotatebox{90}{34} & \rotatebox{90}{33} & \rotatebox{90}{32} & \rotatebox{90}{31} &
\rotatebox{90}{41} & \rotatebox{90}{42} & \rotatebox{90}{43} & \rotatebox{90}{44} & \rotatebox{90}{45} & \rotatebox{90}{46} & \rotatebox{90}{47} & \rotatebox{90}{48} &
\textbf{mDSC} & \textbf{mHD95} \\ \midrule
1 & [4, 4, 4, 4] & \textbf{68.1} & 89.6 & 88.8 & 86.6 & 84.7 & 91.3 & 90.5 & \textbf{93.8} & 91.4 & 88.5 & 88.7 & 85.5 & \textbf{92.0} & \textbf{94.4} & 89.4 & 73.1 & 67.6 & 86.0 & 85.5 & 91.0 & \textbf{91.3} & 86.1 & \textbf{87.1} & 81.7 & 80.8 & \textbf{89.4} & 91.5 & 88.3 & 91.3 & 85.4 & \textbf{83.9} & 58.0 & 85.9 & \textbf{7.7} \\
2 & [8, 8, 8, 8] & 60.6 & 88.1 & 90.7 & 87.2 & \textbf{88.1} & 88.8 & 89.9 & 90.3 & 89.0 & 91.4 & 88.5 & 83.7 & 90.4 & 92.7 & 88.2 & 65.0 & 63.4 & 82.1 & 84.4 & 90.5 & 91.0 & \textbf{86.1} & 86.7 & \textbf{86.1} & \textbf{84.3} & 86.3 & 90.8 & 88.8 & 92.3 & 87.7 & 82.1 & 59.1 & 85.0 & 9.4 \\
3 & [4, 8, 16, 32] & 67.4 & \textbf{90.8} & 86.9 & 86.6 & 85.5 & \textbf{91.7} & \textbf{91.0} & 93.7 & 93.1 & 90.9 & 89.2 & \textbf{86.1} & 90.0 & 93.0 & \textbf{92.2} & 73.0 & \textbf{71.0} & 86.5 & 85.9 & 89.8 & 89.0 & 85.1 & 86.4 & 83.1 & 81.0 & 88.8 & \textbf{92.6} & \textbf{89.4} & 91.4 & \textbf{88.9} & 83.0 & \textbf{60.9} & 86.1 & 7.9 \\
\rowcolor{blue!10} 4 & [16, 16, 16, 16] & 67.9 & 87.6 & \textbf{90.8} & \textbf{88.0} & 85.6 & 91.4 & 89.3 & 93.0 & \textbf{93.3} & \textbf{91.7} & \textbf{89.4} & 84.0 & \textbf{92.0} & 92.4 & 91.4 & \textbf{74.0} & 68.1 & \textbf{88.2} & \textbf{88.6} & \textbf{91.4} & 88.6 & 84.7 & 85.4 & 80.7 & 83.5 & 88.9 & 91.4 & 88.6 & \textbf{92.2} & 88.7 & 83.2 & 59.8 & \textbf{86.2} & 8.3 \\
\bottomrule
\end{tabular}
}
\end{table}

\begin{table}[h]
\centering
\scriptsize
\caption{Ablation study on the effect of the number of experts per encoder stage on the MRISegmenter \cite{zhuang2025mrisegmenter} test set. Evaluation metrics include mDSC ($\uparrow$, \%) and mHD95 ($\downarrow$, mm) across all 62 anatomical classes. Labels are sorted strictly by ID. The best results are \textbf{bolded}.}
\label{tab:mrisegmentator_full_experts}
\scriptsize
\setlength{\tabcolsep}{1.5pt}

\resizebox{\columnwidth}{!}{%
\begin{tabular}{l|ccccccccccccccccccccccccccccccc}
\toprule
\textbf{Experts} &
\rotatebox{90}{1 Spleen} & \rotatebox{90}{2 Kid R} & \rotatebox{90}{3 Kid L} & \rotatebox{90}{4 Gallb} & \rotatebox{90}{5 Liver} & \rotatebox{90}{6 Esoph} & \rotatebox{90}{7 Stom} & \rotatebox{90}{8 Aorta} & \rotatebox{90}{9 IVC} & \rotatebox{90}{10 P\&S Vn} & \rotatebox{90}{11 Pancr} & \rotatebox{90}{12 Adr R} & \rotatebox{90}{13 Adr L} & \rotatebox{90}{14 Lung R} & \rotatebox{90}{15 Lung L} & \rotatebox{90}{16 Sm Bow} & \rotatebox{90}{17 Duod} & \rotatebox{90}{18 Colon} & \rotatebox{90}{19 Ilia Art L} & \rotatebox{90}{20 Ilia Art R} & \rotatebox{90}{21 Ilia Vn L} & \rotatebox{90}{22 Ilia Vn R} & \rotatebox{90}{23 GluMax L} & \rotatebox{90}{24 GluMax R} & \rotatebox{90}{25 GluMed L} & \rotatebox{90}{26 GluMed R} & \rotatebox{90}{27 Auto L} & \rotatebox{90}{28 Auto R} & \rotatebox{90}{29 Ilio L} & \rotatebox{90}{30 Ilio R} & \rotatebox{90}{31 Hip L} \\ \midrule
\relax [4, 4, 4, 4] & 91.1 & \textbf{96.5} & 89.1 & 79.0 & \textbf{98.5} & 85.3 & 93.5 & \textbf{95.5} & \textbf{93.1} & \textbf{65.5} & 87.5 & 60.6 & 59.6 & \textbf{97.5} & 96.5 & \textbf{92.5} & 84.0 & 91.7 & \textbf{77.4} & \textbf{79.1} & 65.5 & 58.0 & 63.8 & \textbf{71.6} & 85.2 & 79.9 & \textbf{97.6} & \textbf{97.4} & \textbf{95.6} & \textbf{96.4} & 80.1 \\
\relax [8, 8, 8, 8] & \textbf{93.4} & 96.4 & 89.4 & 78.0 & \textbf{98.5} & 85.3 & 93.4 & 95.4 & 92.9 & 64.0 & \textbf{88.2} & 60.8 & \textbf{59.7} & 97.4 & 96.4 & 92.4 & \textbf{84.4} & \textbf{91.8} & 76.6 & 77.1 & 65.5 & 56.6 & 72.3 & 56.9 & 84.2 & \textbf{80.7} & 97.4 & \textbf{97.4} & \textbf{95.6} & 96.3 & 80.9 \\
\relax [4, 8, 16, 32] & 91.5 & 96.3 & \textbf{91.5} & 78.6 & \textbf{98.5} & \textbf{85.4} & 93.6 & \textbf{95.5} & \textbf{93.1} & 64.0 & 88.1 & \textbf{61.0} & 59.2 & \textbf{97.5} & \textbf{96.6} & 92.2 & 84.2 & 91.6 & 76.8 & 77.1 & \textbf{66.5} & \textbf{58.2} & 70.0 & 59.1 & \textbf{86.8} & 79.2 & 97.5 & \textbf{97.4} & 95.5 & \textbf{96.4} & \textbf{83.4} \\
\rowcolor{blue!10} \relax [16, 16, 16, 16] & 91.4 & \textbf{96.5} & 88.7 & \textbf{80.0} & \textbf{98.5} & 85.3 & \textbf{93.7} & \textbf{95.5} & 93.0 & 64.5 & 87.9 & 60.9 & 58.0 & \textbf{97.5} & 96.5 & 92.2 & 83.7 & 91.7 & 76.7 & 77.2 & 60.8 & 56.1 & \textbf{72.7} & 64.5 & 84.9 & 80.3 & 97.5 & \textbf{97.4} & \textbf{95.6} & \textbf{96.4} & 83.0 \\
\bottomrule
\end{tabular}
}

\vspace{1.5em} 

\resizebox{\columnwidth}{!}{%
\begin{tabular}{l|ccccccccccccccccccccccccccccccc|cc}
\toprule
\textbf{Experts} &
\rotatebox{90}{32 Hip R} & \rotatebox{90}{33 Sacr} & \rotatebox{90}{34 Rib L4} & \rotatebox{90}{35 Rib L5} & \rotatebox{90}{36 Rib L6} & \rotatebox{90}{37 Rib L7} & \rotatebox{90}{38 Rib L8} & \rotatebox{90}{39 Rib L9} & \rotatebox{90}{40 Rib L10} & \rotatebox{90}{41 Rib L11} & \rotatebox{90}{42 Rib L12} & \rotatebox{90}{43 Rib R4} & \rotatebox{90}{44 Rib R5} & \rotatebox{90}{45 Rib R6} & \rotatebox{90}{46 Rib R7} & \rotatebox{90}{47 Rib R8} & \rotatebox{90}{48 Rib R9} & \rotatebox{90}{49 Rib R10} & \rotatebox{90}{50 Rib R11} & \rotatebox{90}{51 Rib R12} & \rotatebox{90}{52 L5} & \rotatebox{90}{53 L4} & \rotatebox{90}{54 L3} & \rotatebox{90}{55 L2} & \rotatebox{90}{56 L1} & \rotatebox{90}{57 T12} & \rotatebox{90}{58 T11} & \rotatebox{90}{59 T10} & \rotatebox{90}{60 T9} & \rotatebox{90}{61 T8} & \rotatebox{90}{62 T7} & \textbf{mDSC} & \textbf{mHD95} \\ \midrule
\relax [4, 4, 4, 4] & 70.7 & 54.4 & 15.4 & \textbf{45.7} & \textbf{56.2} & \textbf{62.3} & \textbf{65.6} & 67.8 & 72.3 & \textbf{75.5} & 69.5 & \textbf{22.0} & 42.4 & 57.6 & 65.3 & \textbf{68.7} & 70.6 & 70.0 & \textbf{71.1} & 67.4 & 73.7 & 87.6 & 88.1 & 88.0 & 87.7 & 88.8 & 88.7 & 87.9 & 80.0 & 69.1 & 43.7 & \textbf{77.7} & 5.8\\
\relax [8, 8, 8, 8] & 67.2 & \textbf{57.5} & 13.7 & 43.0 & 55.7 & 61.9 & 64.2 & 68.7 & 72.0 & 72.8 & 69.2 & 19.1 & 44.9 & 56.6 & 65.4 & 68.2 & \textbf{72.0} & \textbf{73.2} & 70.5 & \textbf{67.7} & 74.9 & 86.7 & 87.1 & 87.3 & 87.8 & 88.5 & \textbf{89.2} & \textbf{88.4} & 79.6 & 73.0 & 45.2 & 77.5 & \textbf{5.6} \\
\relax[4, 8, 16, 32] & 67.5 & 52.1 & \textbf{18.0} & 44.1 & 53.7 & 60.2 & 64.2 & \textbf{69.5} & \textbf{73.1} & 74.0 & \textbf{69.7} & 19.4 & \textbf{47.1} & 56.7 & 65.2 & 67.0 & 68.1 & 72.0 & 70.6 & 67.5 & \textbf{76.1} & \textbf{87.7} & 88.4 & 88.1 & 87.7 & 88.8 & \textbf{89.2} & 87.7 & \textbf{80.5} & 70.2 & \textbf{46.1} & 77.6 & 5.9 \\
\rowcolor{blue!10}  \relax [16, 16, 16, 16] & \textbf{70.8} & 55.2 & 13.7 & 45.1 & 53.2 & 59.9 & 65.0 & 69.0 & 72.3 & 73.3 & 69.0 & 21.0 & 42.9 & \textbf{57.7} & \textbf{65.5} & 68.3 & 68.8 & 71.3 & \textbf{71.1} & 66.2 & 72.7 & 87.4 & \textbf{88.6} & \textbf{89.1} & \textbf{89.2} & \textbf{89.8} & 88.1 & 87.2 & 79.8 & \textbf{73.2} & 45.0 & 77.4 & 5.7\\
\bottomrule
\end{tabular}
}
\end{table}

\begin{table}[h]
\centering
\caption{Ablation study on the effect of the number of experts per encoder stage on the MVSeg \cite{carnahan2021deepmitral} test set. Evaluation metrics include mDSC ($\uparrow$, \%) and mHD95 ($\downarrow$, mm). Organ abbreviations: PL (posterior leaflet) and AL (anterior leaflet). The best results are \textbf{bolded}.}
\label{tab:organ_ablation_reset_config}
\scriptsize
\setlength{\tabcolsep}{8pt}
\begin{tabular}{l|c|cc|cc}
\toprule
& & \multicolumn{2}{c|}{\textbf{Organ Classes (DSC \%) $\uparrow$}} & \multicolumn{2}{c}{\textbf{Overall}} \\ \cmidrule(l){2-2} \cmidrule(l){3-4} \cmidrule(l){5-6}
\textbf{\#} & \textbf{Experts} & \rotatebox{90}{PL} & \rotatebox{90}{AL} & \textbf{mDSC} & \textbf{mHD95} \\ \midrule
1 & [4, 4, 4, 4] & 84.1 & 85.5 & 84.8 & 4.8 \\
2 & [8, 8, 8, 8] & \textbf{84.2} & \textbf{85.7} & \textbf{85.0} & \textbf{4.2}\\
3 & [4, 8, 16, 32] & 83.4 & 85.5 & 84.5 & 4.4 \\
\rowcolor{blue!10} 4 & [16, 16, 16, 16] & \textbf{84.2} & 85.6 & 84.9 & 4.3\\
\bottomrule
\end{tabular}
\end{table}

\clearpage

\section{Quantitative results}

\subsection{Segmentation}

\begin{table}[h]
\centering
\caption{Quantitative segmentation performance on the AMOS-CT \cite{ji2022amos} validation set. Evaluation metrics include mDSC (\%) and mHD95 (mm). Organ abbreviations: Spl (spleen), RK/LK (right/left kidney), GB (gallbladder), Eso (esophagus), Liv (liver), Sto (stomach), Aor (aorta), IVC (inferior vena cava), Pan (pancreas), RAG/LAG (right/left adrenal gland), Duo (duodenum), Bla (bladder), and P/U. The best results are \textbf{bolded}. $^{\dagger}$ indicates pre-trained models.}
\label{tab:organ_ablation_shortcuts}
\scriptsize
\setlength{\tabcolsep}{5pt}
\resizebox{\textwidth}{!}{%
\begin{tabular}{l|ccccccccccccccc|cc}
\toprule
& \multicolumn{15}{c|}{\textbf{Organ Classes (DSC \%) $\uparrow$}} & \multicolumn{2}{c}{\textbf{Overall}} \\ \cmidrule(l){2-16} \cmidrule(l){17-18}
\textbf{Method} &
\rotatebox{90}{Spl} & \rotatebox{90}{RK} & \rotatebox{90}{LK} & \rotatebox{90}{GB} & \rotatebox{90}{Eso} & \rotatebox{90}{Liv} & \rotatebox{90}{Sto} & \rotatebox{90}{Aor} & \rotatebox{90}{IVC} & \rotatebox{90}{Pan} & \rotatebox{90}{RAG} & \rotatebox{90}{LAG} & \rotatebox{90}{Duo} & \rotatebox{90}{Bla} & \rotatebox{90}{P/U} &
\textbf{mDSC} & \textbf{mHD95} \\ \midrule
VoCo$^{\dagger}$ \cite{wu2024voco,wu2025large} & 96.2 & 94.3 & 95.4 & 81.6 & 83.2 & 96.0 & 91.1 & 93.9 & 89.7 & 85.5 & 76.0 & 75.7 & 79.8 & 88.0 & 79.4 & 87.1 & 16.4 \\
uC 3DU-Net \cite{huang2024upping} & 94.8 & 93.1 & 94.7 & 70.1 & 79.6 & 89.2 & 86.7 & 93.0 & 86.9 & 77.8 & 73.1 & 73.3 & 72.6 & 84.5 & 76.6 & 83.1 & 16.0 \\
SuPreM$^{\dagger}$ \cite{liwell} & 96.3 & 95.7 & 95.8 & 81.2 & 83.5 & 97.3 & 90.9 & 93.6 & 89.4 & 85.5 & 74.4 & 75.2 & 79.6 & 88.4 & 80.0 & 87.1 & 7.7 \\
SegMamba \cite{xing2024segmamba} & 95.0 & 95.5 & 95.4 & 75.5 & 80.7 & 96.2 & 88.2 & 93.5 & 88.9 & 82.7 & 75.1 & 73.6 & 75.6 & 86.0 & 80.1 & 85.5 & 17.3 \\
Mamba-HoME \cite{plotka2026mamba} & 96.0 & 95.0 & 94.4 & 81.7 & 82.0 & 94.7 & 90.2 & 93.8 & 88.9 & 84.0 & 74.4 & 74.1 & 78.4 & 83.8 & 83.2 & 86.3 & 18.7  \\ \midrule
Merlin$^{\dagger}$ \cite{blankemeier2026merlin} & 94.9 & 95.3 & 95.1 & 79.5 & 80.0 & 96.7 & 90.4 & 93.4 & 88.4 & 84.0 & 74.1 & 71.6 & 77.2 & 84.9 & 79.2 & 85.6 & 29.8 \\
fVLM$^{\dagger}$ \cite{shuilarge} & 95.4 & 94.8 & 95.4 & 75.2 & 81.4 & 94.7 & 88.0 & 93.1 & 88.0 & 83.1 & 74.6 & 73.9 & 76.5 & 84.2 & 80.4 & 85.2 & 16.4 \\
BTB3D$^{\dagger}$ \cite{hamamci2025better} & 96.0 & 95.2 & 93.8 & 80.2 & 81.8 & 97.1 & 90.0 & 93.2 & 87.2 & 85.0 & 73.6 & 73.4 & 78.8 & 87.3 & 79.2 & 86.1 & 18.3 \\\midrule
\rowcolor{blue!10} \textbf{A$^{2}$MoE}$^{\dagger}$ & \textbf{96.7} & \textbf{96.1} & \textbf{95.6} & \textbf{84.8} & \textbf{84.9} & \textbf{97.7} & \textbf{92.2} & \textbf{94.3} & \textbf{90.6} & \textbf{87.3} & \textbf{77.0} & \textbf{77.6} & \textbf{82.9} & \textbf{89.5} & \textbf{83.6} & \textbf{88.7} & \textbf{6.2} \\
\bottomrule
\end{tabular}
}
\end{table}

\begin{table}[h]
\centering
\caption{Quantitative segmentation and detection performance on the PANORAMA \cite{alves2026artificial} test sets (NIH, MSD, and proprietary). Evaluation metrics include mDSC (\%) and mHD95 (mm). The best results are \textbf{bolded}. $^{\dagger}$~indicates pre-trained models.}
\label{tab:panorama_segmentation_refined}
\scriptsize
\setlength{\tabcolsep}{5pt}
\begin{tabular}{l|c|cc|cc|cc}
\toprule
& \textbf{NIH} & \multicolumn{2}{c|}{\textbf{MSD Pancreas}} & \multicolumn{2}{c|}{\textbf{Proprietary}} & \multicolumn{2}{c}{\textbf{Overall}} \\ \cmidrule(lr){2-2} \cmidrule(lr){3-4} \cmidrule(lr){5-6} \cmidrule(lr){7-8}
\textbf{Method} & \textbf{Pancreas} & \textbf{PDAC} & \textbf{Pancreas} & \textbf{PDAC} & \textbf{Pancreas} & \textbf{mDSC} & \textbf{mHD95} \\ \midrule
VoCo$^{\dagger}$ \cite{wu2024voco,wu2025large} & 90.1 & 38.7 & 86.5 & 43.3 & 80.3 & 71.6 & 15.9 \\
SegMamba \cite{xing2024segmamba} & 92.7 & 47.9 & 88.1 & 54.0 & 81.9 & 76.0 & 8.6 \\
uC 3DU-Net \cite{huang2024upping} & 92.6 & 49.0 & 88.9 & 56.8 & 80.2 & 76.6 & \textbf{4.7} \\
SuPreM$^{\dagger}$ \cite{liwell} & 92.3 & 46.0 & 88.6 & 61.4 & 82.0 & 76.6 & \textbf{4.7} \\
Mamba-HoME \cite{plotka2026mamba} & 92.9 & 51.2 & 88.4 & 60.8 & 81.7 & 77.5 & 4.8 \\ \midrule
Merlin$^{\dagger}$ \cite{blankemeier2026merlin} & 92.5 & 51.0 & 88.1 & 62.7 & 81.2 & 77.5 & 9.4\\
fVLM$^{\dagger}$ \cite{shuilarge} & 91.5 & 44.1 & 88.0 & 58.4 & 81.0 & 76.8 & 9.8\\
BTB3D$^{\dagger}$ \cite{hamamci2025better} & 92.0 & 52.1 & 88.4 & 61.5 & 81.4 & 77.3 & 9.2\\ \midrule
\rowcolor{blue!10} \textbf{A$^{2}$MoE}$^{\dagger}$ & \textbf{93.1} & \textbf{55.2} & \textbf{89.1} & \textbf{67.2} & \textbf{82.2} & \textbf{79.1} & 5.8\\
\bottomrule
\end{tabular}
\end{table}

\begin{table}[h]
\centering
\caption{Quantitative segmentation performance on the AMOS-MRI \cite{ji2022amos} validation set. Evaluation metrics include mDSC ($\uparrow$, \%) and mHD95 ($\downarrow$, mm). Organ abbreviations: Spl (spleen), RK/LK (right/left kidney), GB (gallbladder), Eso (esophagus), Liv (liver), Sto (stomach), Aor (aorta), IVC (inferior vena cava), Pan (pancreas), RAG/LAG (right/left adrenal gland), and Duo (duodenum). Evaluation covers 13 foreground classes; urinary bladder and
prostate/uterus are excluded because their annotations are unavailable
in the test set. The best results are \textbf{bolded}. $^{\dagger}$ indicates pre-trained models.}
\label{tab:organ_ablation_shortcuts_modified}
\scriptsize
\setlength{\tabcolsep}{5pt}
\resizebox{\textwidth}{!}{%
\begin{tabular}{l|ccccccccccccc|cc}
\toprule
& \multicolumn{13}{c|}{\textbf{Organ Classes (DSC \%) $\uparrow$}} & \multicolumn{2}{c}{\textbf{Overall}} \\ \cmidrule(l){2-14} \cmidrule(l){15-16}
\textbf{Method} &
\rotatebox{90}{Spl} & \rotatebox{90}{RK} & \rotatebox{90}{LK} & \rotatebox{90}{GB} & \rotatebox{90}{Eso} & \rotatebox{90}{Liv} & \rotatebox{90}{Sto} & \rotatebox{90}{Aor} & \rotatebox{90}{IVC} & \rotatebox{90}{Pan} & \rotatebox{90}{RAG} & \rotatebox{90}{LAG} & \rotatebox{90}{Duo} &
\textbf{mDSC} & \textbf{mHD95} \\ \midrule
VoCo$^{\dagger}$ \cite{wu2024voco,wu2025large} & 91.2 & 90.5 & 91.1 & 71.2 & 74.9 & 97.2 & 84.7 & 87.1 & 83.9 & 82.9 & 62.0 & 59.5 & 65.0 & 80.1 & 10.5 \\
uC 3DU-Net \cite{huang2024upping} & 88.7 & 89.7 & 88.7 & 52.0 & 70.2 & 95.7 & 76.6 & 85.4 & 81.2 & 76.4 & 56.3 & 52.2 & 53.8 & 74.4 & 16.9 \\
SuPreM$^{\dagger}$ \cite{liwell} & 89.9 & 89.8 & 89.2 & 67.7 & 67.3 & 95.6 & 79.5 & 85.1 & 79.4 & 79.8 & 55.2 & 54.1 & 58.0 & 75.8 & 19.5 \\
SegMamba \cite{xing2024segmamba} & 86.4 & 89.8 & 89.6 & 62.5 & 70.5 & 95.7 & 82.4 & 87.0 & 81.0 & 83.5 & 59.8 & 64.2 & 64.4 & 76.7 & 9.3 \\
Mamba-HoME \cite{plotka2026mamba} & 87.4 & 88.9 & 88.3 & 69.8 & 71.7 & 95.9 & 82.7 & 85.3 & 82.3 & 83.4 & 59.0 & 64.2 & 63.5 & 77.7 & 9.9  \\ \midrule
Merlin$^{\dagger}$ \cite{blankemeier2026merlin} & 90.2 & 89.5 & 90.1 & 70.2 & 73.9 & 96.2 & 83.7 & 86.1 & 82.9 & 81.9 & 61.0 & 58.5 & 64.0 & 79.1 & 10.4 \\
fVLM$^{\dagger}$ \cite{shuilarge} & 89.2 & 88.5 & 89.1 & 69.2 & 72.9 & 95.2 & 82.7 & 85.1 & 81.9 & 80.9 & 60.0 & 57.5 & 63.0 & 78.1 & 10.9 \\
BTB3D$^{\dagger}$ \cite{hamamci2025better} & 89.0 & 88.3 & 88.9 & 69.0 & 72.7 & 95.0 & 82.5 & 84.9 & 81.7 & 80.7 & 59.8 & 57.3 & 62.8 & 77.8 & 11.1 \\ \midrule
\rowcolor{blue!10} A$^{2}$MoE & \textbf{91.5} & \textbf{90.9} & \textbf{91.2} & \textbf{78.9} & \textbf{75.0} & \textbf{97.4} & \textbf{86.8} & \textbf{87.6} & \textbf{84.4} & \textbf{86.5} & \textbf{64.5} & \textbf{67.0} & \textbf{68.1} & \textbf{82.3} & \textbf{8.9} \\
\bottomrule
\end{tabular}
}
\end{table}

\begin{table}[h]
\centering
\caption{Quantitative evaluation of tooth segmentation on the GEPAR3D \cite{szczepanski2025gepar3d} test set. FDI notation is used for individual teeth across quadrants (Q1--Q4). Evaluation metrics include mDSC ($\uparrow$, \%) and mHD95 ($\downarrow$, mm). The best results are \textbf{bolded}. $^{\dagger}$ indicates pre-trained models.}
\label{tab:dental_segmentation_empty}
\scriptsize
\setlength{\tabcolsep}{2.5pt}
\resizebox{\columnwidth}{!}{%
\begin{tabular}{l|cccccccc|cccccccc|cccccccc|cccccccc|cc}
\toprule
& \multicolumn{8}{c|}{\textbf{Upper Right (Q1)}} & \multicolumn{8}{c|}{\textbf{Upper Left (Q2)}} & \multicolumn{8}{c|}{\textbf{Lower Left (Q3)}} & \multicolumn{8}{c|}{\textbf{Lower Right (Q4)}} & \multicolumn{2}{c}{\textbf{Overall}} \\ \cmidrule(l){2-9} \cmidrule(l){10-17} \cmidrule(l){18-25} \cmidrule(l){26-33} \cmidrule(l){34-35}
\textbf{Method} &
\rotatebox{90}{18} & \rotatebox{90}{17} & \rotatebox{90}{16} & \rotatebox{90}{15} & \rotatebox{90}{14} & \rotatebox{90}{13} & \rotatebox{90}{12} & \rotatebox{90}{11} &
\rotatebox{90}{21} & \rotatebox{90}{22} & \rotatebox{90}{23} & \rotatebox{90}{24} & \rotatebox{90}{25} & \rotatebox{90}{26} & \rotatebox{90}{27} & \rotatebox{90}{28} &
\rotatebox{90}{38} & \rotatebox{90}{37} & \rotatebox{90}{36} & \rotatebox{90}{35} & \rotatebox{90}{34} & \rotatebox{90}{33} & \rotatebox{90}{32} & \rotatebox{90}{31} &
\rotatebox{90}{41} & \rotatebox{90}{42} & \rotatebox{90}{43} & \rotatebox{90}{44} & \rotatebox{90}{45} & \rotatebox{90}{46} & \rotatebox{90}{47} & \rotatebox{90}{48} &
\textbf{mDSC} & \textbf{mHD95} \\ \midrule
VoCo$^{\dagger}$ \cite{wu2024voco,wu2025large} & 63.9 & 89.6 & 88.4 & 86.8 & 88.6 & 92.7 & 89.4 & 92.3 & 90.0 & 86.7 & 86.3 & 86.3 & 89.8 & 91.8 & 89.2 & 65.4 & 64.3 & 86.6 & 85.9 & 87.4 & 85.5 & 78.5 & 81.2 & 74.7 & 77.8 & 85.8 & 86.4 & 84.5 & 89.1 & 84.6 & 83.5 & 56.2 & 83.5 & 6.8 \\
uC 3DU-Net \cite{huang2024upping} & 67.7 & 89.5 & 87.7 & 84.6 & 83.0 & 87.3 & 88.5 & 92.3 & 90.7 & 87.9 & 87.7 & 86.5 & 88.8 & 91.0 & 88.2 & 66.4 & 64.9 & 79.0 & 82.8 & 86.0 & 85.9 & 83.8 & 87.5 & 82.8 & 82.6 & 89.5 & 89.0 & 84.5 & 87.5 & 87.1 & 82.4 & 53.6 & 83.8 & 12.5 \\
SuPreM$^{\dagger}$ \cite{liwell} & 62.7 & 85.9 & 87.0 & 82.1 & 75.1 & 81.9 & 86.8 & 84.6 & 87.0 & 86.1 & 86.1 & 83.0 & 88.4 & 89.5 & 85.3 & 66.8 & 61.4 & 79.5 & 83.7 & 83.2 & 76.7 & 81.0 & 81.5 & 76.3 & 75.8 & 81.6 & 88.0 & 78.6 & 83.8 & 85.8 & 79.2 & 58.1 & 80.6 & 11.0 \\
SegMamba \cite{xing2024segmamba} & 68.0 & 87.9 & 90.2 & 85.3 & 85.2 & 91.7 & 91.0 & 93.7 & \textbf{93.1} & 90.9 & 89.2 & 86.1 & 90.0 & 93.0 & 85.3 & 70.6 & 66.9 & 78.3 & 84.5 & 89.4 & 89.4 & 85.4 & 87.2 & 86.0 & 80.3 & 89.1 & 92.6 & 89.1 & 93.3 & 84.8 & 81.4 & 56.6 & 85.3 & 10.6 \\
Mamba-HoME \cite{plotka2026mamba} & 69.7 & 90.6 & \textbf{91.2} & 86.5 & 84.2 & \textbf{93.9} & 91.1 & \textbf{93.9} & 91.4 & \textbf{92.0} & 89.9 & 86.8 & 90.9 & 94.4 & 90.1 & 69.6 & 66.8 & 84.0 & 83.5 & 88.9 & 87.4 & 85.8 & 85.9 & 81.7 & 84.4 & \textbf{89.4} & 92.3 & 89.8 & 93.5 & 85.4 & 84.2 & 60.0 & 86.1 & 6.5 \\ \midrule
Merlin$^{\dagger}$ \cite{blankemeier2026merlin} & 68.2 & 88.8 & 89.5 & 86.0 & 82.9 & 91.8 & 90.9 & 93.2 & 89.1 & 87.2 & \textbf{90.7} & 86.6 & \textbf{91.6} & 94.1 & \textbf{90.8} & \textbf{76.7} & \textbf{69.6} & \textbf{89.9} & 87.5 & \textbf{90.9} & 89.0 & \textbf{86.7} & 85.2 & 84.3 & 80.7 & 87.5 & \textbf{92.9} & 89.7 & 92.0 & 89.3 & 86.9 & 61.9 & 86.5 & \textbf{2.7} \\
fVLM$^{\dagger}$\cite{shuilarge} & \textbf{71.2} & 88.2 & 87.0 & 86.5 & 85.7 & 88.1 & 87.6 & 92.6 & 90.4 & 86.0 & 85.3 & 85.1 & 90.4 & 93.8 & 89.2 & \textbf{76.7} & 68.1 & 84.1 & \textbf{88.1} & 88.0 & 85.6 & 81.5 & 83.1 & 76.2 & 76.6 & 82.8 & 89.0 & 86.8 & 91.8 & 83.5 & 81.1 & 59.5 & 84.1 & 10.1 \\
BTB3D$^{\dagger}$ \cite{hamamci2025better} & 71.0 & 88.0 & 86.8 & 86.3 & 85.5 & 87.9 & 87.4 & 92.4 & 90.2 & 85.8 & 85.1 & 84.9 & 90.2 & 93.6 & 89.0 & 76.5 & 67.9 & 83.9 & 87.9 & 87.8 & 85.4 & 81.3 & 82.9 & 76.0 & 76.4 & 82.5 & 88.7 & 86.5 & 91.5 & 83.2 & 80.8 & 59.2 & 84.0 & 9.8\\ \midrule
\rowcolor{blue!10} \textbf{A$^{2}$MoE}$^{\dagger}$ & 70.4 & \textbf{92.2} & 89.1 & \textbf{88.3} & \textbf{91.0} & 93.5 & \textbf{90.4} & 93.6 & 91.5 & 91.7 & 88.7 & \textbf{87.4} & 91.3 & \textbf{94.2} & 90.7 & \textbf{76.7} & 64.3 & 87.1 & 87.5 & 89.8 & \textbf{91.1} & 86.5 & \textbf{89.5} & \textbf{86.0} & \textbf{84.5} & 88.5 & 87.6 & \textbf{91.2} & \textbf{93.8} & \textbf{90.0} & \textbf{88.3} & \textbf{64.3} & \textbf{87.7} & 8.1 \\
\bottomrule
\end{tabular}
}
\end{table}

\begin{table}[h]
\centering
\scriptsize
\caption{Quantitative evaluation of MRISegmenter \cite{zhuang2025mrisegmenter} across all 62 anatomical classes. Labels are sorted strictly by ID. Evaluation metrics include mDSC ($\uparrow$, \%) and mHD95 ($\downarrow$, mm). The best results are \textbf{bolded}. $^{\dagger}$ indicates pre-trained models.}
\label{tab:mri_segmentation_full}
\scriptsize
\setlength{\tabcolsep}{1.5pt}

\resizebox{\columnwidth}{!}{%
\begin{tabular}{l|ccccccccccccccccccccccccccccccc}
\toprule
\textbf{Method} &
\rotatebox{90}{1 Spleen} & \rotatebox{90}{2 Kid R} & \rotatebox{90}{3 Kid L} & \rotatebox{90}{4 Gallb} & \rotatebox{90}{5 Liver} & \rotatebox{90}{6 Esoph} & \rotatebox{90}{7 Stom} & \rotatebox{90}{8 Aorta} & \rotatebox{90}{9 IVC} & \rotatebox{90}{10 P\&S Vn} & \rotatebox{90}{11 Pancr} & \rotatebox{90}{12 Adr R} & \rotatebox{90}{13 Adr L} & \rotatebox{90}{14 Lung R} & \rotatebox{90}{15 Lung L} & \rotatebox{90}{16 Sm Bow} & \rotatebox{90}{17 Duod} & \rotatebox{90}{18 Colon} & \rotatebox{90}{19 Ilia Art L} & \rotatebox{90}{20 Ilia Art R} & \rotatebox{90}{21 Ilia Vn L} & \rotatebox{90}{22 Ilia Vn R} & \rotatebox{90}{23 GluMax L} & \rotatebox{90}{24 GluMax R} & \rotatebox{90}{25 GluMed L} & \rotatebox{90}{26 GluMed R} & \rotatebox{90}{27 Auto L} & \rotatebox{90}{28 Auto R} & \rotatebox{90}{29 Ilio L} & \rotatebox{90}{30 Ilio R} & \rotatebox{90}{31 Hip L} \\ \midrule
VoCo \cite{wu2024voco} & 91.0 & 96.3 & 88.9 & 74.4 & 98.4 & 84.2 & 93.0 & 95.2 & 92.1 & 62.0 & 86.2 & 59.7 & 55.1 & 97.2 & 96.0 & 90.9 & 81.9 & 89.6 & 74.9 & 75.0 & 59.3 & 50.8 & 63.6 & 57.1 & 83.6 & 76.9 & 97.2 & 97.1 & 95.2 & 95.8 & 73.7 \\
SuPreM \cite{liwell} & 90.2 & 96.1 & 85.7 & 70.9 & 98.2 & 83.1 & 91.1 & 95.1 & 91.5 & 61.6 & 85.4 & 59.1 & 54.5 & 97.1 & 95.7 & 88.6 & 80.3 & 86.8 & 69.7 & 72.3 & 55.2 & 49.4 & 50.0 & 43.2 & 83.1 & 75.4 & 97.0 & 96.8 & 94.6 & 95.6 & 72.4 \\
uC 3DU-Net \cite{huang2024upping} & 91.0 & 96.2 & 87.3 & \textbf{79.2} & \textbf{98.5} & \textbf{85.4} & 92.9 & 95.3 & 92.8 & \textbf{67.0} & 87.9 & 59.5 & 57.1 & \textbf{97.5} & \textbf{96.6} & 92.0 & 82.2 & 91.3 & 76.1 & 79.5 & 66.0 & 57.5 & 66.8 & 58.5 & 84.5 & 79.0 & \textbf{97.5} & 97.3 & \textbf{95.6} & \textbf{96.4} & \textbf{84.3} \\
SegMamba \cite{xing2024segmamba} & 90.7 & \textbf{96.4} & 86.6 & 73.6 & 98.4 & 84.7 & 92.6 & 95.2 & 91.9 & 61.7 & 87.1 & 60.2 & 58.0 & 97.1 & 96.1 & 91.0 & 82.6 & 89.8 & 75.9 & \textbf{80.2} & 67.1 & 59.7 & 63.5 & 60.4 & 84.6 & 78.9 & 97.3 & 97.1 & 95.3 & 95.8 & 80.5 \\
Mamba-HoME \cite{plotka2026mamba} & 91.3 & 96.3 & 90.3 & 76.9 & \textbf{98.5} & 84.7 & 93.4 & 95.1 & 92.3 & 65.4 & 88.2 & 60.2 & 58.1 & 97.4 & 96.5 & 92.5 & 83.2 & 91.7 & 77.3 & 79.8 & 74.3 & 72.6 & 74.2 & 72.8 & 87.2 & 84.0 & 97.3 & 97.2 & 95.4 & 95.9 & 82.6 \\
\midrule
Merlin \cite{blankemeier2026merlin} & \textbf{91.5} & 96.3 & \textbf{91.7} & 78.1 & \textbf{98.5} & 84.7 & \textbf{93.7} & 95.1 & 92.5 & 66.8 & \textbf{88.6} & 60.3 & 58.3 & \textbf{97.5} & \textbf{96.6} & \textbf{93.0} & 83.4 & \textbf{92.4} & \textbf{77.8} & 79.7 & \textbf{77.0} & \textbf{77.4} & \textbf{78.1} & \textbf{77.3} & \textbf{88.1} & \textbf{85.9} & 97.3 & 97.2 & 95.5 & 96.0 & 83.4 \\
fVLM \cite{shuilarge} & 88.3 & 93.3 & 86.6 & 74.7 & 93.9 & 80.6 & 90.2 & 91.9 & 89.3 & 59.5 & 84.3 & 56.4 & 53.9 & 94.0 & 93.3 & 89.1 & 79.9 & 88.0 & 74.5 & 77.3 & 70.8 & 73.7 & 69.7 & 69.9 & 84.5 & 82.2 & 94.0 & 94.0 & 92.2 & 93.0 & 78.6 \\
BTB3D \cite{hamamci2025better} & 91.2 & 96.2 & 89.5 & 77.6 & 98.3 & 83.5 & 93.1 & 94.8 & 92.2 & 62.4 & 87.2 & 59.3 & 56.8 & 97.1 & 96.2 & 92.0 & 82.8 & 90.9 & 77.4 & \textbf{80.2} & 73.7 & 76.6 & 72.6 & 72.8 & 87.4 & 85.1 & 97.1 & 97.1 & 95.1 & 95.9 & 81.5 \\ \midrule
\rowcolor{blue!10} \textbf{A$^{2}$MoE} & \textbf{91.5} & 96.3 & 91.5 & 78.6 & \textbf{98.5} & \textbf{85.4} & 93.6 & \textbf{95.5} & \textbf{93.1} & 64.0 & 88.1 & \textbf{61.0} & \textbf{59.2} & \textbf{97.5} & \textbf{96.6} & 92.2 & \textbf{84.2} & 91.6 & 76.8 & 77.1 & 66.5 & 58.2 & 70.0 & 59.1 & 86.8 & 79.2 & \textbf{97.5} & \textbf{97.4} & 95.5 & \textbf{96.4} & 83.4 \\
\bottomrule
\end{tabular}
}

\vspace{1.5em}

\resizebox{\columnwidth}{!}{%
\begin{tabular}{l|ccccccccccccccccccccccccccccccc|cc}
\toprule
\textbf{Method} &
\rotatebox{90}{32 Hip R} & \rotatebox{90}{33 Sacr} & \rotatebox{90}{34 Rib L4} & \rotatebox{90}{35 Rib L5} & \rotatebox{90}{36 Rib L6} & \rotatebox{90}{37 Rib L7} & \rotatebox{90}{38 Rib L8} & \rotatebox{90}{39 Rib L9} & \rotatebox{90}{40 Rib L10} & \rotatebox{90}{41 Rib L11} & \rotatebox{90}{42 Rib L12} & \rotatebox{90}{43 Rib R4} & \rotatebox{90}{44 Rib R5} & \rotatebox{90}{45 Rib R6} & \rotatebox{90}{46 Rib R7} & \rotatebox{90}{47 Rib R8} & \rotatebox{90}{48 Rib R9} & \rotatebox{90}{49 Rib R10} & \rotatebox{90}{50 Rib R11} & \rotatebox{90}{51 Rib R12} & \rotatebox{90}{52 L5} & \rotatebox{90}{53 L4} & \rotatebox{90}{54 L3} & \rotatebox{90}{55 L2} & \rotatebox{90}{56 L1} & \rotatebox{90}{57 T12} & \rotatebox{90}{58 T11} & \rotatebox{90}{59 T10} & \rotatebox{90}{60 T9} & \rotatebox{90}{61 T8} & \rotatebox{90}{62 T7} & \textbf{mDSC} & \textbf{mHD95} \\ \midrule
VoCo \cite{wu2024voco} & 67.2 & 48.8 & 15.3 & 36.8 & 50.8 & 55.4 & 56.1 & 58.8 & 65.6 & 68.7 & 63.2 & 13.9 & 34.3 & 47.3 & 57.1 & 59.7 & 61.4 & 62.2 & 65.9 & 62.6 & 66.1 & 85.1 & 86.4 & 86.1 & 85.3 & 84.9 & 84.1 & 79.6 & 71.5 & 55.0 & 38.5 & 73.0 & 7.7 \\
SuPreM \cite{liwell} & 61.2 & 44.6 & 12.0 & 34.7 & 47.7 & 49.2 & 51.7 & 54.1 & 63.3 & 67.5 & 62.5 & 14.4 & 32.9 & 46.6 & 52.1 & 56.3 & 59.8 & 62.4 & 62.2 & 58.9 & 60.4 & 80.4 & 83.9 & 86.5 & 85.4 & 83.2 & 78.6 & 73.6 & 65.4 & 50.7 & 39.5 & 70.8 & 8.4 \\
uC 3DU-Net \cite{huang2024upping} & 70.5 & 49.9 & 0.0 & 39.3 & 57.9 & 63.3 & 63.8 & 67.6 & 71.3 & 72.6 & 68.6 & 13.6 & 38.7 & 57.8 & 65.4 & 67.0 & 67.9 & 69.7 & 70.4 & 65.6 & 72.9 & 87.2 & 86.9 & 86.7 & 85.9 & 87.7 & 85.8 & 85.2 & 77.4 & 67.6 & 41.7 & 76.7 & \textbf{5.9} \\
SegMamba \cite{xing2024segmamba} & 69.6 & 49.7 & 15.9 & 38.8 & 48.3 & 53.8 & 57.1 & 60.5 & 69.0 & 70.0 & 66.7 & 18.0 & 36.0 & 50.0 & 60.6 & 64.6 & 67.0 & 69.4 & 68.1 & 64.7 & 66.4 & 85.6 & 87.0 & 87.1 & 87.3 & 88.6 & 88.2 & 84.9 & 78.7 & 69.4 & 48.9 & 74.5 & 6.8 \\
Mamba-HoME \cite{plotka2026mamba} & 75.7 & 60.3 & 4.2 & 41.3 & 52.1 & 58.7 & 62.0 & 64.7 & 70.0 & 70.9 & 66.1 & 19.3 & 43.2 & 55.7 & 63.7 & 65.8 & 67.3 & 68.3 & 67.9 & 66.2 & 74.7 & 87.1 & 87.7 & 87.1 & 87.3 & 87.4 & 87.2 & 85.6 & 76.8 & 68.2 & 48.1 & 75.2 & 6.5 \\
\midrule
Merlin \cite{blankemeier2026merlin} & \textbf{78.0} & \textbf{64.4} & 0.0 & 42.3 & 53.6 & 60.6 & 64.0 & 66.4 & 70.4 & 71.2 & 65.9 & 19.9 & 46.0 & 58.0 & 65.0 & 66.2 & 67.4 & 67.9 & 67.8 & 66.8 & 77.8 & 87.6 & 88.0 & 87.1 & 87.3 & 86.9 & 86.8 & 85.9 & 76.1 & 67.8 & 48.0 & 75.8 & 6.0 \\
fVLM \cite{shuilarge} & 70.7 & 53.0 & 0.0 & 42.3 & 54.1 & 55.9 & 58.5 & 60.9 & 67.0 & 67.7 & 60.4 & 13.7 & 36.0 & 53.7 & 58.1 & 61.2 & 66.7 & 67.1 & 66.3 & 60.1 & 72.3 & 85.4 & 84.8 & 84.7 & 83.8 & 82.8 & 83.9 & 81.0 & 72.5 & 68.8 & 45.9 & 72.0 & 8.1 \\
BTB3D \cite{hamamci2025better} & 73.6 & 55.9 & 0.0 & 45.2 & 57.0 & 58.8 & 61.4 & 63.8 & 69.9 & 70.6 & 63.3 & 16.6 & 38.9 & 56.6 & 61.0 & 64.1 & 69.6 & 70.0 & 69.2 & 63.0 & 75.2 & 88.3 & 87.7 & 87.6 & 86.7 & 85.7 & 86.8 & 83.9 & 75.4 & 71.7 & 48.8 & 76.9 & 6.2 \\ \midrule
\rowcolor{blue!10} \textbf{A$^{2}$MoE} & 75.0 & 60.0 & \textbf{22.0} & \textbf{48.0} & \textbf{58.0} & \textbf{64.0} & \textbf{68.0} & \textbf{72.5} & \textbf{75.5} & \textbf{76.0} & \textbf{72.0} & \textbf{24.0} & \textbf{51.0} & \textbf{60.0} & \textbf{68.0} & \textbf{70.0} & \textbf{71.0} & \textbf{75.0} & \textbf{73.5} & \textbf{70.0} & \textbf{80.0} & \textbf{90.0} & \textbf{90.5} & \textbf{90.5} & \textbf{90.0} & \textbf{91.0} & \textbf{91.5} & \textbf{90.0} & \textbf{85.0} & \textbf{78.0} & \textbf{55.0} & \textbf{79.8} & \textbf{5.9} \\
\bottomrule
\end{tabular}
}
\end{table}

\begin{table}[h]
\centering
\caption{Quantitative segmentation performance on the MVSeg2023 \cite{carnahan2021deepmitral} test set. Evaluation metrics include mDSC ($\uparrow$, \%) and mHD95 ($\downarrow$, mm). Organ abbreviations: PL (posterior leaflet) and AL (anterior leaflet). The best results are \textbf{bolded}. $^{\dagger}$ indicates pre-trained models.}
\label{tab:organ_ablation_reset}
\scriptsize
\setlength{\tabcolsep}{10pt}
\begin{tabular}{l|cc|cc}
\toprule
& \multicolumn{2}{c|}{\textbf{Organ Classes (DSC \%) $\uparrow$}} & \multicolumn{2}{c}{\textbf{Overall}} \\ \cmidrule(l){2-3} \cmidrule(l){4-5}
\textbf{Method} & \rotatebox{90}{PL} & \rotatebox{90}{AL} & \textbf{mDSC} & \textbf{mHD95} \\ \midrule
VoCo$^{\dagger}$ \cite{wu2024voco,wu2025large} & 83.3 & 85.3 & 84.3 & 5.1 \\
uC 3DU-Net \cite{huang2024upping} & 82.9 & 85.0 & 83.9 & 4.7 \\
SuPreM$^{\dagger}$ \cite{liwell} & 83.2 & 85.4 & 84.3 & 4.9 \\
SegMamba \cite{xing2024segmamba} & 83.0 & 84.7 & 83.8 & 5.8 \\
Mamba-HoME \cite{plotka2026mamba} & 83.6 & 85.3 & 84.5 & 4.8 \\ \midrule
Merlin$^{\dagger}$ \cite{blankemeier2026merlin} & 83.3 & 84.9 & 84.1 & \textbf{3.9} \\
fVLM$^{\dagger}$ \cite{shuilarge} & 79.0 & 81.3 & 80.2 & 5.0 \\
BTB3D$^{\dagger}$ \cite{hamamci2025better} & 83.5 & 85.7 & 84.6 & 4.0\\ \midrule
\rowcolor{blue!10} \textbf{A$^{2}$MoE}$^{\dagger}$ & \textbf{84.5} & \textbf{86.0} & \textbf{85.4} & 4.0 \\ 
\bottomrule
\end{tabular}
\end{table}

\clearpage

\subsection{Classification}

\begin{table}[h]
\centering
\scriptsize
\caption{Binary trauma classification on RATIC (0 vs.\ injury): ROC-AUC values multiplied by 100 and rounded to one decimal place. The best results are \textbf{bolded}.}
\label{tab:ratic_binary_compact}
\setlength{\tabcolsep}{6pt}
\begin{tabular}{lccccc}
\toprule
Method & Bowel & Liver & Kidney & Spleen & Mean \\
\midrule
VoCo~\cite{wu2024voco}       & 64.2 & 67.1 & 88.1 & 86.2 & 76.4 \\
SuPreM~\cite{liwell}   & 72.6 & 76.2 & 90.4 & \textbf{92.5} & 82.9 \\
BTB3D~\cite{hamamci2025better}     & 70.1 & 74.8 & 89.4 & 87.2 & 80.4 \\
Merlin~\cite{blankemeier2026merlin}   & 79.8 & 81.2 & \textbf{94.2} & 90.8 & 86.5 \\
\rowcolor{blue!10} A$^2$MoE & \textbf{82.4} & \textbf{84.6} & 93.8 & 92.1 & \textbf{88.2} \\
\bottomrule
\end{tabular}
\end{table}

\begin{table}[h]
\centering
\scriptsize
\caption{Per-grade abdominal trauma classification on RATIC \cite{rudie2024rsna}. AUC: area under the receiver operating characteristic curve; Sens: sensitivity; Spec: specificity. All metrics are reported as percentages. The best result for each metric in each row is \textbf{bolded}.}
\label{tab:ratic_per_grade}
\resizebox{\textwidth}{!}{%
\begin{tabular}{ll ccc ccc ccc ccc ccc}
\toprule
& & \multicolumn{3}{c}{VoCo \cite{wu2024voco}} & \multicolumn{3}{c}{SuPreM \cite{liwell}} & \multicolumn{3}{c}{BTB3D \cite{hamamci2025better}} & \multicolumn{3}{c}{Merlin \cite{blankemeier2026merlin}} & \multicolumn{3}{c}{A$^2$MoE} \\
\cmidrule(lr){3-5}\cmidrule(lr){6-8}\cmidrule(lr){9-11}\cmidrule(lr){12-14}\cmidrule(lr){15-17}
Organ & Grading & AUC & Sens. & Spec. & AUC & Sens. & Spec. & AUC & Sens. & Spec. & AUC & Sens. & Spec. & AUC & Sens. & Spec. \\
\midrule
Bowel & 0 vs.\ 1
 & 64.2 & 61.9 & 63.1
 & 72.6 & 71.4 & 68.4
 & 70.1 & 66.7 & 65.8
 & 79.8 & \textbf{81.0} & 74.2
 & \textbf{82.4} & 76.2 & \textbf{79.5} \\
\midrule
\multirow{3}{*}{Liver}
 & 0 vs.\ 1+2
 & 67.1 & 62.8 & 64.2
 & 76.2 & 69.1 & 72.8
 & 74.8 & 67.0 & 71.5
 & 81.2 & 74.5 & 77.9
 & \textbf{84.6} & \textbf{78.7} & \textbf{80.8} \\
 & 0 vs.\ 1
 & 63.8 & 60.5 & 64.2
 & 72.4 & 65.8 & 72.8
 & 70.9 & 63.2 & 71.5
 & \textbf{78.1} & 69.7 & \textbf{77.9}
 & 77.3 & \textbf{72.4} & 74.2 \\
 & 0 vs.\ 2
 & 76.8 & 72.2 & 64.2
 & 85.4 & 83.3 & 72.8
 & 83.2 & 77.8 & 71.5
 & 89.1 & \textbf{88.9} & 77.9
 & \textbf{91.8} & 83.3 & \textbf{80.8} \\
\midrule
\multirow{3}{*}{Kidney}
 & 0 vs.\ 1+2
 & 88.1 & 76.4 & 82.3
 & 90.4 & 80.0 & \textbf{89.2}
 & 89.4 & 78.2 & 83.5
 & \textbf{94.2} & 83.6 & 87.1
 & 93.8 & \textbf{85.5} & 88.6 \\
 & 0 vs.\ 1
 & 85.4 & 68.4 & \textbf{88.6}
 & 87.9 & 73.5 & 83.5
 & 86.7 & 73.5 & 82.4
 & 90.8 & \textbf{82.4} & 86.3
 & \textbf{92.1} & 79.4 & 87.2 \\
 & 0 vs.\ 2
 & 91.8 & 81.0 & 84.6
 & 94.2 & 85.7 & 87.3
 & 93.2 & 85.7 & 85.8
 & 95.7 & \textbf{95.2} & 88.5
 & \textbf{96.8} & 90.5 & \textbf{89.8} \\
\midrule
\multirow{3}{*}{Spleen}
 & 0 vs.\ 1+2
 & 86.2 & 71.6 & 79.8
 & \textbf{92.5} & 75.2 & 82.4
 & 87.2 & 72.5 & \textbf{87.5}
 & 90.8 & 79.8 & 85.8
 & 92.1 & \textbf{81.7} & 87.2 \\
 & 0 vs.\ 1
 & 83.1 & 66.7 & 78.5
 & 85.2 & 69.8 & 81.2
 & 84.2 & 68.3 & 79.8
 & 88.7 & 74.6 & 84.8
 & \textbf{90.4} & \textbf{76.2} & \textbf{86.2} \\
 & 0 vs.\ 2
 & 89.8 & 78.3 & 81.2
 & 92.1 & 80.4 & 83.5
 & 91.2 & 80.4 & \textbf{89.0}
 & 93.8 & \textbf{84.8} & 86.2
 & \textbf{94.8} & 82.6 & 87.9 \\
\bottomrule
\end{tabular}%
}
\end{table}

\clearpage

\subsection{Detection}

\begin{table}[ht]
\centering
\scriptsize
\caption{Detection performance on the AbdomenAtlas 3.0 test set. We compare sensitivity (Sens.) for tumor sizes $\le$2 cm and $>$2 cm, along with specificity (Spec.) in \%. The best results are \textbf{bolded}.}
\label{tab:detection_results}
\setlength{\tabcolsep}{3.5pt} 
\begin{tabular}{l ccc c ccc c ccc}
\toprule
 & \multicolumn{3}{c}{Pancreatic Tumor (\%)} & & \multicolumn{3}{c}{Kidney Tumor (\%)} & & \multicolumn{3}{c}{Liver Tumor (\%)} \\
\cmidrule(lr){2-4} \cmidrule(lr){6-8} \cmidrule(lr){10-12}
Model & \makecell{Sen.\\($\le$2 cm)} & \makecell{Sens.\\($>$2 cm)} & Spec. & & \makecell{Sens.\\($\le$2 cm)} & \makecell{Sens.\\($>$2 cm)} & Spec. & & \makecell{Sens.\\($\le$2 cm)} & \makecell{Sens.\\($>$2 cm)} & Spec. \\
\midrule
CT-CHAT & \textbf{66.7} & 51.9 & 61.2 & & 31.1 & 32.8 & 74.2 & & 5.7 & 3.2 & 94.7 \\
CT2Rep & 0.0 & 0.0 & 92.5 & & 36.5 & 39.3 & 70.4 & & 35.8 & 49.2 & 70.4 \\
M3D & 0.0 & 7.4 & 97.2 & & 8.1 & 16.4 & 84.1 & & 9.4 & 12.7 & 86.0 \\
Merlin & 33.3 & 51.9 & 71.8 & & 28.4 & 45.9 & 86.6 & & 30.2 & 41.3 & \textbf{95.9} \\
RadFM & 0.0 & 0.0 & \textbf{99.9} & & 3.7 & 6.3 & \textbf{95.6} & & 3.3 & 5.7 & 93.9 \\
RadGPT & \textbf{66.7} & 81.5 & 93.2 & & 54.8 & 93.3 & 51.8 & & 39.6 & \textbf{96.8} & 64.4 \\
\rowcolor{blue!10} \textbf{A$^{2}$MoE (ours)} & \textbf{66.7} & \textbf{83.3} & 94.1 & & \textbf{58.2} & \textbf{94.1} & 55.4 & & \textbf{41.2} & 95.2 & 85.4 \\
\bottomrule
\end{tabular}
\end{table}

\begin{figure}[h]
    \centering
    \includegraphics[width=0.45\linewidth]{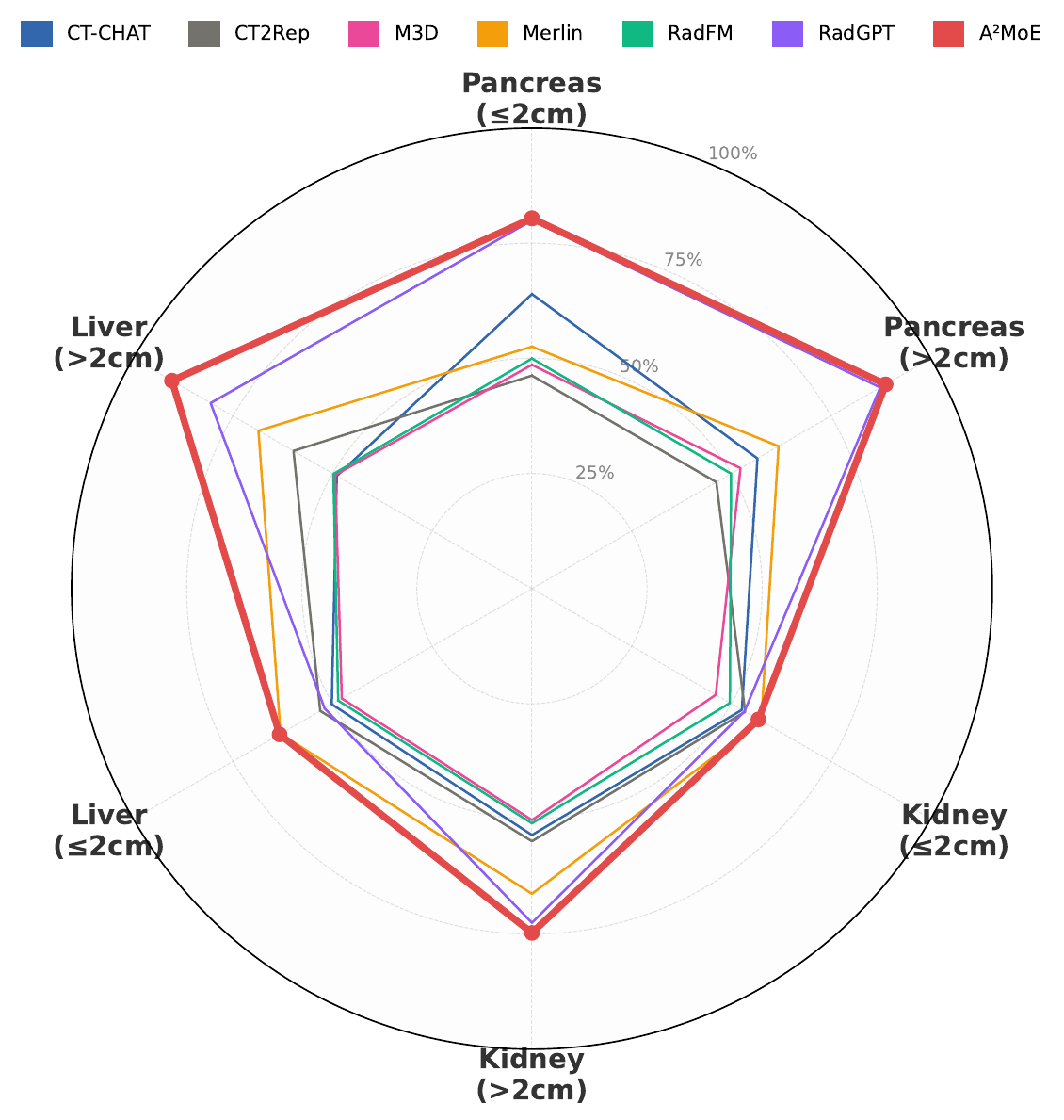}
    \caption{\textbf{Detection performance on AbdomenAtlas 3.0} \cite{bassi2025radgpt}. We compare balanced sensitivity and specificity for tumor sizes $\le$2 cm vs. $>$2 cm across pancreatic, kidney, and liver tumors.}
    \label{fig:detection_plot}
\end{figure}

\clearpage

\section{Qualitative results}
\label{appendix:qualitative_results}

\begin{figure}[h]
    \centering
    \includegraphics[width=1\linewidth]{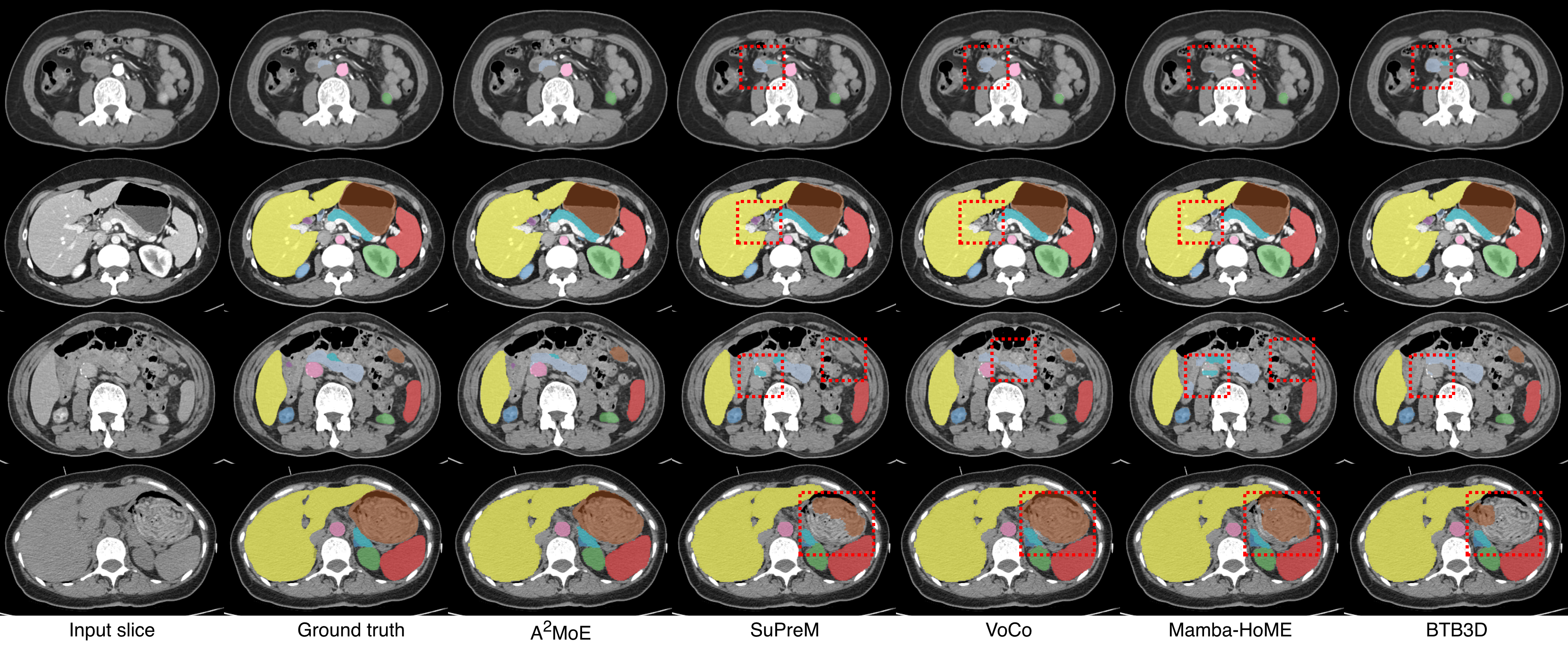}
    \caption{\textbf{Qualitative results on the AMOS-CT validation set.} We present the input slice, reference segmentation, and predictions from A$^{2}$MoE and the four best-performing comparison methods. Key differences are highlighted with red bounding boxes.}
    \label{fig:amos_ct_visualization_results}
\end{figure}

\begin{figure}[h]
    \centering
    \includegraphics[width=1\linewidth]{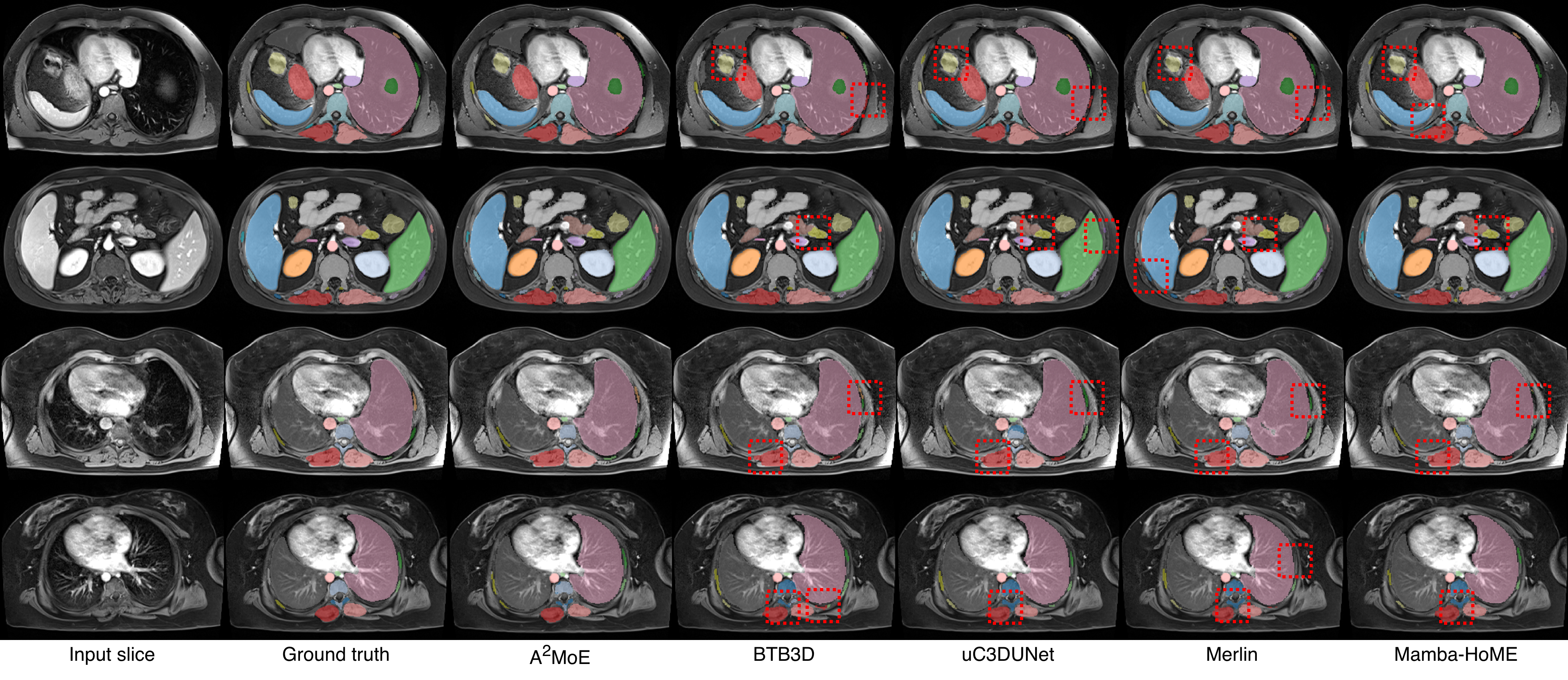}
    \caption{\textbf{Qualitative results on the MRISegmenter test set.} We present the input slice, reference segmentation, and predictions from A$^{2}$MoE and the four best-performing comparison methods. Key differences are highlighted with red bounding boxes.}
    \label{fig:mrisegmenter_figure}
\end{figure}

\begin{figure}
    \centering
    \includegraphics[width=1\linewidth]{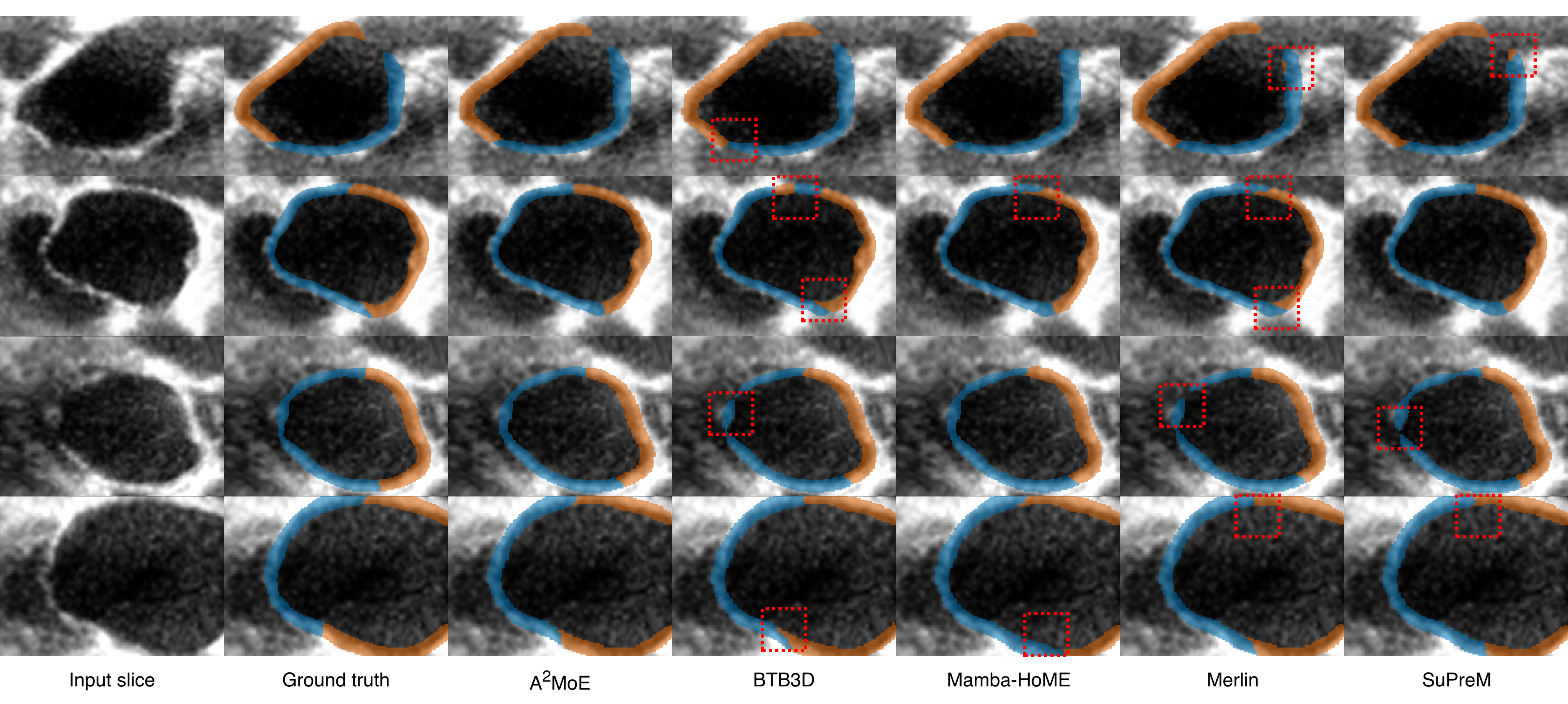}
    \caption{\textbf{Qualitative results on the MVSeg test set.} We present the input slice, reference segmentation, and predictions from A$^{2}$MoE and the four best-performing comparison methods. Key differences are highlighted with red bounding boxes.}
    \label{fig:mvseg_us_visualization_results}
\end{figure}

\begin{figure}
    \centering
    \includegraphics[width=1\linewidth]{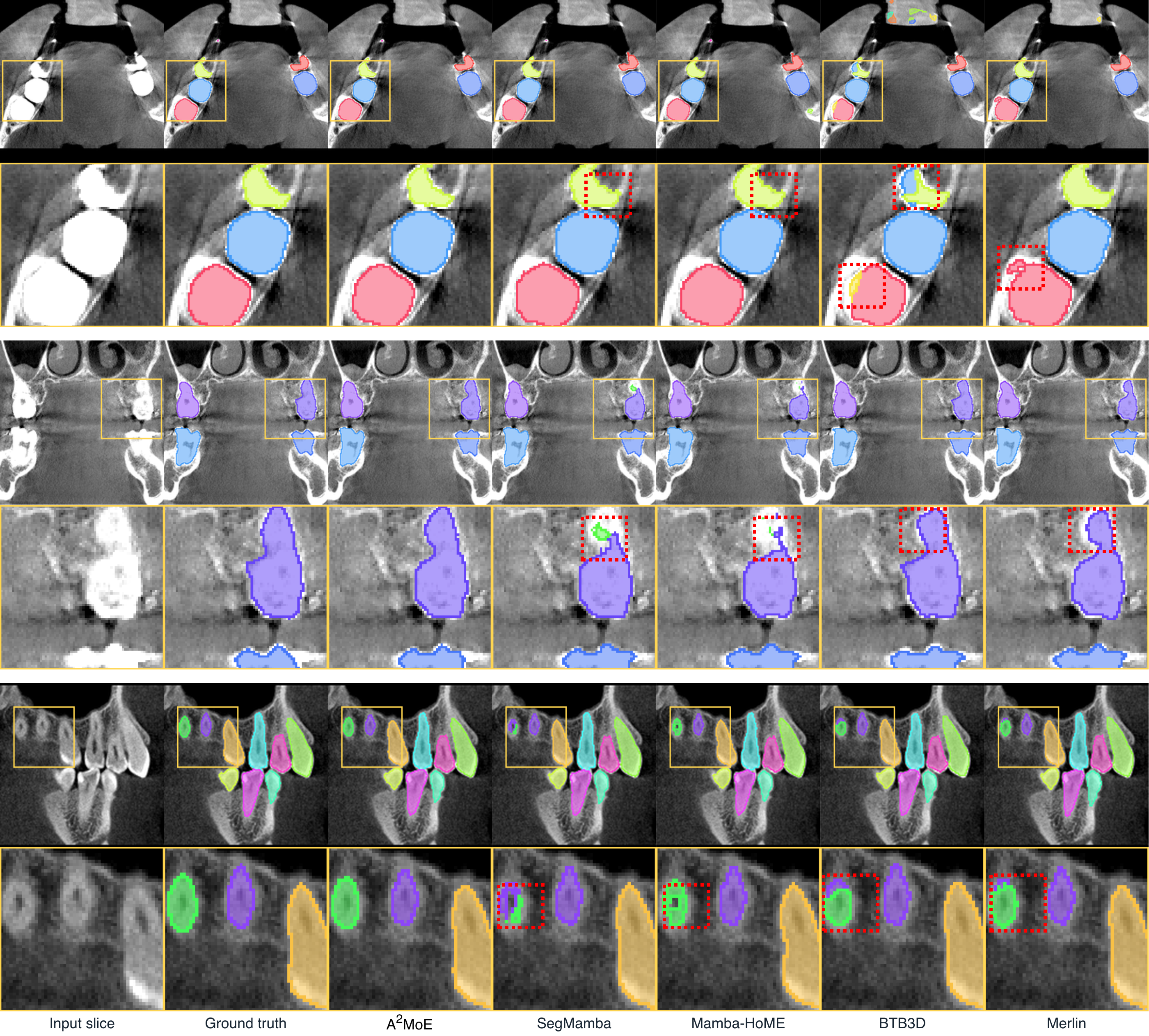}
    \caption{\textbf{Qualitative results on the GEPAR3D test set.} We present the 2D input slice, ground truth, A$^{2}$MoE, and the four next best-performing methods. For each example, the first row shows the original slice, while the second row shows a magnified view of the region. Key differences are highlighted with red bounding boxes.}
    \label{fig:gepar3d_figure}
\end{figure}

\begin{figure}
    \centering
    \includegraphics[width=1\linewidth]{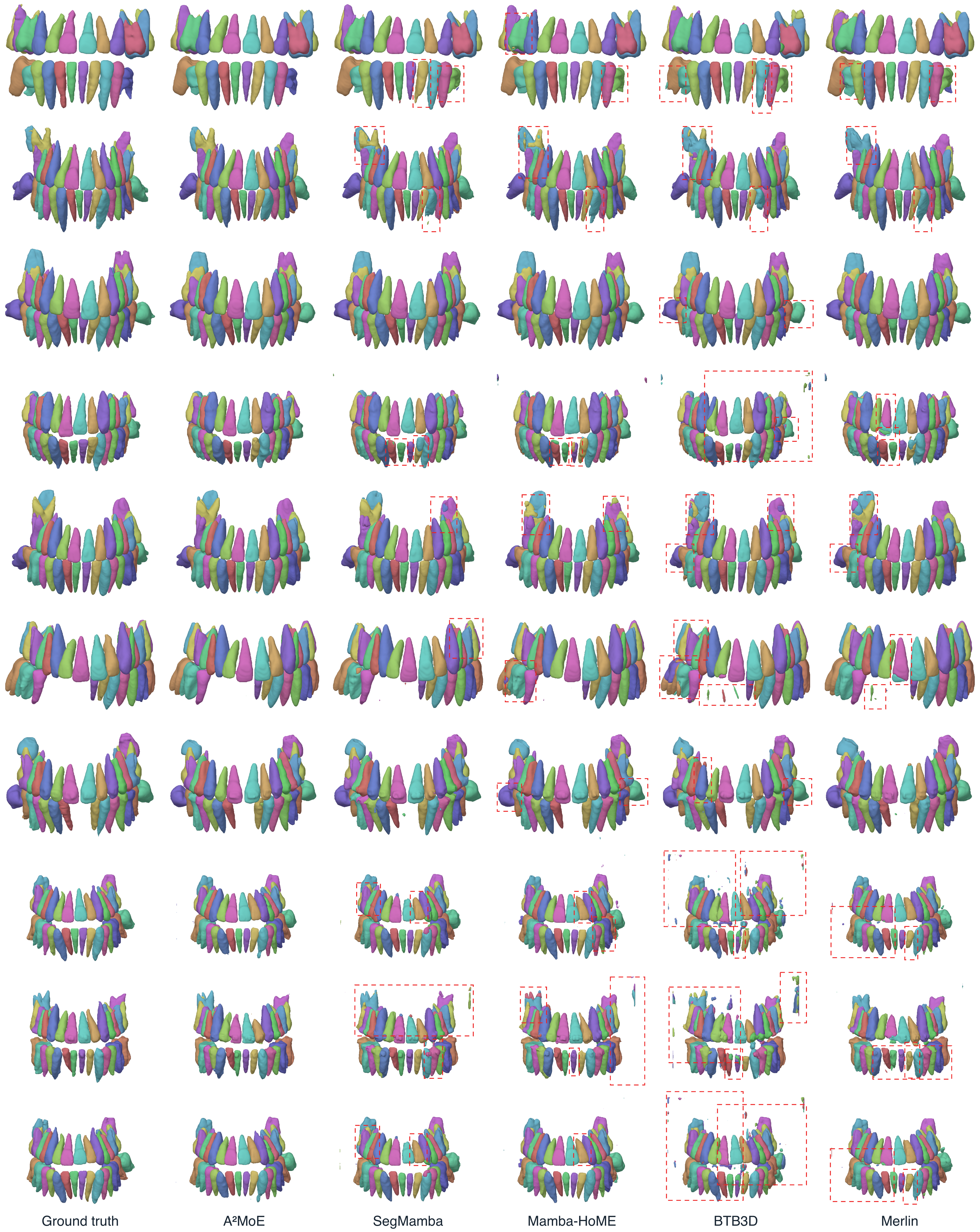}
    \caption{\textbf{Qualitative results on the GEPAR3D test set.} We present the 3D reference segmentation, and predictions from A$^{2}$MoE and the four best-performing comparison methods. Key differences are highlighted with red bounding boxes.}
    \label{fig:3d_gepar3d_figure}
\end{figure}

\clearpage

\section{Statistical significance}
\label{appendix:statistical_significance}

\begin{figure}[h]
    \centering
    \includegraphics[width=1\linewidth]{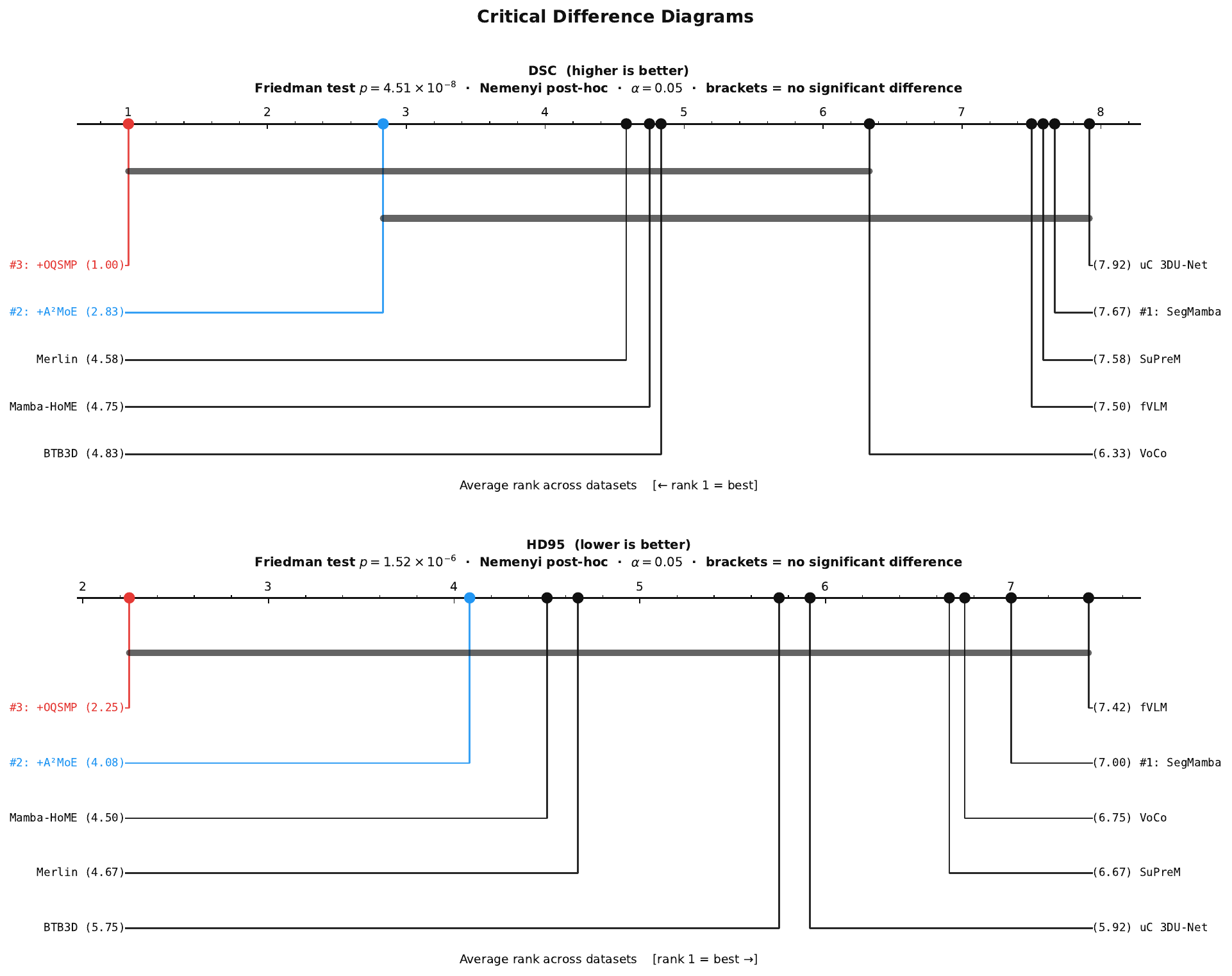}
    \caption{\textbf{Critical difference (CD) diagrams comparing ten segmentation methods
    across six benchmark datasets for DSC (top; higher is better) and
    HD95 (bottom; lower is better).}
    Average ranks are computed independently per dataset and then averaged
    across all six benchmarks; rank~1 denotes the best-performing method.
    Horizontal brackets connect groups of methods for which no statistically
    significant difference was detected (Friedman test followed by Nemenyi
    post-hoc test, $\alpha{=}0.05$).
    Our proposed method A$^2$MoE is evaluated in two configurations:
    \textbf{\#2~{+A$^2$MoE}} applied on top of SegMamba, and
    \textbf{\#3~{+OQMSP}}, which further extends it with organ-query-guided
    semi-supervised pre-training; \#1~SegMamba serves as the baseline.
    The full model (\#3~+OQMSP) achieves the best average rank on both
    metrics (DSC:~1.00, HD95:~2.25) and is statistically
    distinguishable from the majority of competing methods
    ($p = 4.51\cdot 10^{-8}$ and $p = 1.88\cdot 10^{-6}$,
    respectively).}
    \label{fig:cd_combined_statistical_tests}
\end{figure}

\clearpage

\section{Generalizability}
\label{sec:appendixgeneralizability}
The strength of a medical foundation model lies not in its performance on a single dataset, but in its ability to transcend the specific conditions of its training \cite{bassi2024improving}. A central design goal of A$^{2}$MoE is that the inductive biases it
encodes -- conditional computation routed by anatomical composition and
mask-free organ--text alignment via shared queries -- should not be
specific to any single modality, organ system, or label set.  We provide
empirical evidence for this claim along four orthogonal axes:
(i)~\emph{cross-modality transfer} from CT pre-training to MRI and 3D
ultrasound; (ii)~\emph{cross-anatomy transfer} from
abdominal and pelvic supervision to dental cone-beam CT;
(iii)~\emph{out-of-distribution robustness} on multi-center,
multi-vendor, and multi-protocol cohorts that were never seen during
pre-training; and (iv)~\emph{multi-task transfer} to downstream tasks
that the pre-training objective does not directly optimize (organ injury classification on RATIC and tumor detection on
AbdomenAtlas~3.0).

\begin{enumerate}
    \item \textbf{Cross-modality transfer (CT~$\rightarrow$~MRI/US).}
Although A$^{2}$MoE is pre-trained exclusively on multi-phase
CT, the encoder transfers without architectural modification to volumetric MRI
(AMOS-MRI~\cite{ji2022amos}, MRISegmenter~\cite{zhuang2025mrisegmenter})
and 3D transesophageal ultrasound
(MVSeg2023~\cite{carnahan2021deepmitral}). On AMOS-MRI e observe a gain of 5.6 DSC percentage points over the SegMamba backbone and $+2.2$ over
VoCo (Table~\ref{tab:segmentationbenchmark}), and on MRISegmenter the
full A$^{2}$MoE attains the best mean DSC across all 62 anatomical
classes (Table~\ref{tab:mri_segmentation_full}).  These improvements are
notable because (a)~the histogram router was supervised only on
TotalSegmentator CT labels and never observed MRI tissue contrast, and
(b)~OQMSP queries were aligned to English CT-report snippets rather
than MRI findings.  The fact that the same queries still localize the
correct organs on MRI/US supports the interpretation that A$^{2}$MoE
learns a \emph{modality-agnostic} anatomical prior rather than a
CT-specific texture model.
    \item \textbf{Cross-anatomy transfer.} GEPAR3D~\cite{szczepanski2025gepar3d} evaluates fine-grained
multi-class dental segmentation in cone-beam CT, an anatomy that is
\emph{absent} from the pre-training cohort and the histogram class
list (which targets abdominal and pelvic structures).
A$^{2}$MoE nevertheless improves over the SegMamba baseline by
$+0.8$ DSC and matches or exceeds modality-matched 3D pre-training
baselines, indicating that the encoder retains capacity for transfer
to anatomies outside the routing prior.
    \item \textbf{Out-of-distribution robustness.} The downstream cohorts used in this study cover heterogeneous
acquisition conditions: AMOS spans multiple scanners and slice
thicknesses; PANORAMA aggregates seven contributing centers from two
continents; MRISegmenter spans four imaging phases per patient;
RATIC~\cite{rudie2024rsna} consists of 4{,}274 trauma studies from 23
institutions across 14 countries; and AbdomenAtlas~3.0 aggregates 17
previously released public CT datasets.  None of these datasets
contributes to A$^{2}$MoE pre-training. The consistent improvements
reported in Tables~\ref{tab:segmentationbenchmark},
\ref{tab:ratic_binary_compact}, \ref{tab:ratic_per_grade}, and
\ref{tab:detection_results} therefore reflect generalization across
sites, scanners, and protocols, not in-distribution memorization.
    \item \textbf{Multi-task transfer.}
Although the pre-training loss couples segmentation, presence
prediction, and contrastive alignment, A$^{2}$MoE also generalizes to
downstream \emph{classification} (binary trauma $0$ versus $\{1,2\}$ and per-grade
$0$ versus $1$, $0$ versus $2$ on RATIC; Tables~\ref{tab:ratic_binary_compact}--\ref{tab:ratic_per_grade}) and
\emph{detection} (per-organ tumor detection split by lesion size on
AbdomenAtlas~3.0; Table~\ref{tab:detection_results}). On RATIC, the same
backbone improves binary mean ROC-AUC from 86.5 (Merlin) to 88.2
without organ-specific fine-tuning of the encoder, and on
AbdomenAtlas~3.0 it dominates sensitivity for clinically critical
$\leq2$~cm lesions across  pancreatic, kidney, and liver tumors. Together,
these results support the claim that the anatomical prior introduced
by A$^{2}$MoE is broadly useful, rather than a segmentation-specific
inductive bias.

\end{enumerate}

\clearpage

\section{Limitations and future work}

\label{sec:appendixlimitations}

\begin{enumerate}
    \item \textbf{Anatomical scope.}
The histogram and the per-stage class-to-expert affinity matrix
$\mathbf{A}^{(s)}$ are defined over $C-1=14$ abdominal and pelvic
classes, with $C=15$ including background. The pre-training labels themselves are
\emph{not} raw automatic segmentations: TotalSegmentator outputs and
LLM-decomposed report snippets are subsequently revised by
experienced radiologists, as detailed in
Appendix~\ref{sec:appendixdatapreprocessing}. The remaining limitation
is therefore one of \emph{coverage} rather than label quality: anatomies
outside the 14-class taxonomy are routed using the residual visual
context only and do not benefit from the structural prior.
Whole-body, head--neck, musculoskeletal, and pediatric imaging are
likely to require either a richer class taxonomy or hierarchical
routing priors.
    \item \textbf{Modality and language scope.}
Pre-training is performed on adult multi-phase CT paired with
English-language reports. Transfer to MRI, ultrasound, and dental CBCT
is supported by our results, but A$^{2}$MoE has not been pre-trained
with non-English reports, T2/diffusion-weighted
MRI, X-ray, PET, or echocardiographic videos.
The mask-free OQMSP architecture is modality-agnostic by construction,
so extending pre-training to additional modalities is a natural
direction but is not demonstrated here.
    \item \textbf{Backbone and decoder choice.}
We instantiate A$^{2}$MoE on top of a SegMamba-style 3D
encoder~\cite{xing2024segmamba} with a UNETR-style
decoder~\cite{hatamizadeh2022unetr}. The histogram router and OQMSP
are designed to be backbone-agnostic and the gating equation makes no
assumption about the operator inside each expert (we use bottleneck
$1{\times}1{\times}1$ convolutions). Verifying that the same gains
hold for purely convolutional, transformer-only, or state-space-only
backbones is left to future work.
    \item \textbf{Compute and reproducibility.}
Pre-training A$^{2}$MoE requires non-trivial compute and access to one
proprietary CT cohort in addition to the public Merlin~\cite{blankemeier2026merlin}, and CancerVerse~\cite{li2026cancerverse} releases. We mitigate the latter by
(i) reporting the public-only ablations needed for reproducibility,
(ii) providing all hyperparameters and pre-processing details in
Appendix~\ref{sec:experimentalsetup}, and (iii) committing to release
checkpoints, pre-processed metadata, and training code on acceptance.
Reproducing the full proprietary-data results is, however, contingent
on data-use agreements that are outside the authors' control.
    \item \textbf{Clinical readiness.}
A$^{2}$MoE is a representation-learning method evaluated on
retrospective benchmarks. It has not been validated prospectively, has
not undergone reader studies, and is not approved as a medical device.
Its results should not be used to guide clinical decisions.

\end{enumerate}

\textbf{Future directions.} Our findings open several concrete directions:
(i)~scaling the histogram taxonomy beyond 14 classes via hierarchical
or learned anatomical ontologies, with the goal of native whole-body
support;
(ii)~hybrid token/patch routing that recovers fine-grained
specialization when crops contain many organs in similar proportions;
(iii)~longitudinal modeling that pairs the per-organ embeddings of
OQMSP across time points to support follow-up and treatment-response
reasoning;
(iv)~extension to non-CT pre-training corpora (MRI, ultrasound,
PET/CT) to test whether the same anatomy-activated prior emerges natively
in those modalities;
(v)~replacement of the SegMamba backbone with alternative 3D encoders
to confirm backbone independence;
and (vi)~prospective and human-in-the-loop validation on triage and
reporting workflows, paired with calibration and uncertainty
estimation, to bridge the gap between retrospective benchmark
performance and clinical utility.

\clearpage

\section{Impact statement}
\label{sec:appendiximpactstatement}

\label{sec:appendiximpactstatement}

\begin{enumerate}
    \item \textbf{Contribution.} A$^{2}$MoE contributes a 3D medical vision--language pre-training
framework that combines anatomy-activated mixture-of-experts routing with
mask-free organ-query alignment and multi-task supervision. The
intended scientific advance is a more structured, scalable use of
imaging--report pairs for learning representations that support
segmentation, classification, and detection, with explicit modeling
of organ-specific evidence rather than purely global image--text
matching. The approach is mask-free at inference, transfers across
modalities (CT, MRI, ultrasound) and anatomies, and reaches
state-of-the-art performance on external benchmarks, providing a
reusable building block for future medical foundation models.
    \item \textbf{Clinical impact.}
If validated and integrated thoughtfully, the representations learned
by A$^{2}$MoE could support \emph{assistive} radiology workflows:
faster contouring and quality control for segmentation-heavy planning,
triage and worklist prioritization when paired with calibrated
uncertainty, retrieval of similar prior studies, teaching tools that
link images to report language, and research pipelines that reduce
manual annotation burden. Because the model exposes an organ-indexed
soft-localization map per organ via OQMSP cross-attention, downstream
clinical tools can show \emph{which} regions support a given finding,
although the
interpretability and clinical usefulness of these maps require dedicated validation \cite{chrabaszcz2025aggregated}. These benefits
are contingent on rigorous local evaluation, regulatory pathways where
applicable, and continued human responsibility for diagnosis and
treatment decisions.
    \item \textbf{Negative impact.} A$^{2}$MoE inherits limitations common to medical machine-learning systems and can produce confident errors, particularly under distribution shift, in patients with atypical anatomy, or in pediatric or pregnant populations. It may encode and amplify biases present
in who is scanned, how findings are documented, and which sites and
scanners dominate the training data, and it has not been audited for
fairness across demographic, scanner, or protocol strata. Without
careful governance, misuse could include over-reliance on automation,
privacy risks from mishandling of patient data, deployment to
populations underrepresented in training, or use as a stand-alone
diagnostic tool. We therefore emphasize that this work is a
methodological contribution, not a clinical product, and that any
real-world deployment must be preceded by prospective validation,
robustness and fairness auditing, clinician oversight, transparent
limitation reporting, and full compliance with ethics review,
de-identification standards, and data-use agreements.
\end{enumerate}

\clearpage

\section{Notation}

Throughout the paper, we denote the batch size by~$B$ and the number of
segmentation classes, including background, by~$C$ ($C=15$ in pre-training), with $d_e$ denoting the dimension
of the shared vision--text embedding space.  An input volume
$\mathbf{X}\in\mathbb{R}^{B\times 1\times D\times H\times W}$ is paired with a
voxel-wise label map $L\in\{0,\dots,C{-}1\}^{B\times D\times H\times W}$, and
the pyramid encoder produces MoE-refined features
$\mathbf{X}_s\in\mathbb{R}^{B\times C_s\times D_s\times H_s\times W_s}$ at four
stages $s\in\{1,\dots,4\}$, with $K_s$ experts per stage. Let $N_s$ denote the number of projector key/value tokens after any strided convolution.
At stage~$s$ the A$^{2}$MoE router combines a global context $\mathbf{v}_s$, a
crop-level routing representation $\widetilde{\mathbf h}=\mathcal T(\mathbf h)$, where each raw histogram $\mathbf h\in\Delta^{C-1}$ is replaced at inference
by the stage-specific composition estimate $\hat{\mathbf h}_s$ before normalization, and a log-prior built from a
learnable affinity matrix $\mathbf{A}^{(s)}\!\in\!\mathbb{R}^{C\times K_s}$,
producing expert weights $\mathbf W_s$ from the joint route $\Psi_{\rm gate}^{(s)}$, direct route $\Psi_{\rm anat}^{(s)}$, and structural prior; the stage-specific coefficient
$\beta_s>0$ controls the prior's strength and $\epsilon$ stabilizes its
logarithm.  On top of the encoder, the OQMSP uses a single set of
organ queries $\mathbf{Q}\!\in\!\mathbb{R}^{(C-1)\times d_e}$ shared across
stages, produces per-stage, per-organ embeddings $\hat{\mathbf{E}}_s$, and
fuses them with softmax-normalized weights $\boldsymbol{\alpha}$ into
$\mathbf{Z}_{\text{vis}}\!\in\!\mathbb{R}^{B\times(C-1)\times d_e}$, which is
aligned via a learnable log-temperature $\log\tau$ with text embeddings
$\mathbf{z}_{\text{txt}}\!\in\!\mathbb{S}^{d_e-1}$ drawn from a frozen
PubMedBERT through a projection head $\phi_{\text{txt}}$. The final objective weights are segmentation, anatomy-conditioned routing, organ--text alignment, cross-stage consistency, and query-grounding losses 
$\lambda_{\rm seg},\lambda_{\rm MoE},\lambda_{\rm align},\lambda_{\rm stage},\lambda_{\rm ground}$.

\end{document}